\pdfoutput=1

\documentclass[11pt]{article}

\usepackage[final]{acl}

\usepackage{times}
\usepackage{latexsym}
\usepackage{times}
\usepackage{latexsym}
\usepackage{booktabs}
\usepackage{svg}
\usepackage{amsmath}
\usepackage{graphicx}
\usepackage[normalem]{ulem}
\useunder{\uline}{\ul}{}
\usepackage{float}
\usepackage{multirow}
\usepackage{bbold}
\usepackage{hhline}
\usepackage{cleveref}
\usepackage{adjustbox}
\usepackage{todonotes}
\usepackage{listings}
\usepackage{lipsum}

\usepackage{booktabs}

\usepackage[T1]{fontenc}

\usepackage[utf8]{inputenc}

\usepackage{microtype}

\usepackage{inconsolata}

\usepackage{svg}

\usepackage{xcolor}

\usepackage{colortbl}

\usepackage{scalerel}

\usepackage{amssymb}
\usepackage{pifont}
\newcommand{\cmark}{\ding{51}}%

\usepackage{xspace}

\usepackage{listings}
\definecolor{MidnightBlue}{HTML}{191970}
\newcommand{\ourdata}{FEDI\xspace}

\newcolumntype{g}{>{\columncolor{Gray}}r}

\definecolor{positive_color}{HTML}{1E8449}
\definecolor{negative_color}{HTML}{fa8072}

\title{Learning from Implicit User Feedback, Emotions and Demographic Information in Task-Oriented and  Document-Grounded Dialogues}

\author{Dominic Petrak 
     \and
     Thy Thy Tran
     \and 
     Iryna Gurevych%
    \\
    Ubiquitous Knowledge Processing Lab (UKP Lab), \\
    Department of Computer Science and Hessian Center for AI (hessian.AI), \\
    Technical University of Darmstadt \\ 
    \url{www.ukp.tu-darmstadt.de}
  \\}

\begin{document}
\maketitle


\begin{abstract}

Implicit user feedback, user emotions and demographic information have shown to be promising sources for improving the accuracy and user engagement of responses generated by dialogue systems. However, the influence of such information on task completion and factual consistency, which are important criteria for task-oriented and document-grounded dialogues, is not yet known. To address this, we introduce \ourdata, the first English task-oriented and document-grounded dialogue dataset annotated with this information. Our experiments with Flan-T5, GPT-2 and Llama 2 show a particularly positive impact on task completion and factual consistency. Participants in our human evaluation reported that the responses generated by the feedback-trained models were more informative (Flan-T5 and GPT-2), relevant and factual consistent (Llama 2).\footnote{\label{github} Code and data are available in \href{https://github.com/UKPLab/FEDI}{GitHub}.}
\end{abstract}

\begin{table*}[htb]
  \centering
  \resizebox*{\linewidth}{!}{
\begin{tabular}{lcccccrrrr}
\multicolumn{1}{c}{\textbf{Dataset}} & \textbf{Source} & \textbf{Type} & \textbf{\begin{tabular}[c]{@{}c@{}}Demographic\\Information\end{tabular}} & \textbf{\begin{tabular}[c]{@{}c@{}}User\\Emotions\end{tabular}} & \textbf{\begin{tabular}[c]{@{}c@{}}Implicit User\\Feedback\end{tabular}} & \multicolumn{1}{c}{\textbf{\#Dialogues}} & \multicolumn{1}{c}{\textbf{\begin{tabular}[c]{@{}c@{}}Avg. Num.\\ of Turns\end{tabular}}} & \multicolumn{1}{c}{\textbf{\begin{tabular}[c]{@{}c@{}}Avg. Utt.\\ Length\end{tabular}}} & \multicolumn{1}{c}{\textbf{\begin{tabular}[c]{@{}c@{}}Lexical\\ Diversity\end{tabular}}} \\ \hhline{==========}
\multicolumn{1}{l|}{\begin{tabular}[c]{@{}l@{}}EmoWOZ\\ \cite{feng-etal-2022-emowoz}\end{tabular}} & \multicolumn{1}{c|}{\multirow{9}{*}{Crowdsourced}} & \multicolumn{1}{c|}{\begin{tabular}[c]{@{}c@{}}Task-Oriented\end{tabular}} & \multicolumn{1}{c|}{} & \multicolumn{1}{c|}{\cmark} & \multicolumn{1}{c|}{} & 12k & 9.5 & 8.2 & 55.7 \\ \cline{1-1} \cline{3-10} 

\multicolumn{1}{l|}{\begin{tabular}[c]{@{}l@{}}FITS\\ \cite{xu-etal-2023-learning}\end{tabular}} & \multicolumn{1}{c|}{} & \multicolumn{1}{c|}{\begin{tabular}[c]{@{}c@{}}Document-\\ Grounded\end{tabular}} & \multicolumn{1}{c|}{} & \multicolumn{1}{c|}{} & \multicolumn{1}{c|}{\cmark} & 22k & 7.1 & 15.0 & 52.8 \\ \cline{1-1} \cline{3-10} 

\multicolumn{1}{l|}{\begin{tabular}[c]{@{}l@{}}SaferDialogues\\ \cite{ung-etal-2022-saferdialogues}\end{tabular}} & \multicolumn{1}{c|}{} & \multicolumn{1}{c|}{\multirow{2}{*}{Open-Domain}} & \multicolumn{1}{c|}{} & \multicolumn{1}{c|}{} & \multicolumn{1}{c|}{\cmark} & 8k & 2.5 & 14.8 & 53.3 \\ \cline{1-1} \cline{4-10} 
\multicolumn{1}{l|}{\begin{tabular}[c]{@{}l@{}}EmotionLines\\ \cite{hsu-etal-2018-emotionlines}\end{tabular}} & \multicolumn{1}{c|}{} & \multicolumn{1}{c|}{} & \multicolumn{1}{c|}{} & \multicolumn{1}{c|}{\cmark} & \multicolumn{1}{c|}{} & 1k & 7.3 & 7.8 & 68.5 \\ \cline{1-1} \cline{2-10} 
\multicolumn{1}{l|}{\begin{tabular}[c]{@{}l@{}}SODA\\ \cite{kim-etal-2023-soda}\end{tabular}} & \multicolumn{1}{c|}{\multirow{2}{*}{\begin{tabular}[c]{@{}c@{}}LLM-\\ Generated\end{tabular}}} & \multicolumn{1}{c|}{\multirow{2}{*}{Open-Domain}} & \multicolumn{1}{c|}{} & \multicolumn{1}{c|}{\cmark} & \multicolumn{1}{c|}{} & 1.5M & 7.6 & 16.1 & 68.0 \\ \cline{1-1} \cline{4-10} 
\multicolumn{1}{l|}{\begin{tabular}[c]{@{}l@{}}PersonaChatGen\\ \cite{lee-etal-2022-personachatgen}\end{tabular}} & \multicolumn{1}{c|}{} & \multicolumn{1}{c|}{} & \multicolumn{1}{c|}{\cmark} & \multicolumn{1}{c|}{} & \multicolumn{1}{c|}{} & 1.6k & 16.0 & 9.5 & 56.7 \\ \hhline{==========}
\multicolumn{1}{l|}{\textbf{\ourdata}} & \multicolumn{1}{c|}{\textbf{\begin{tabular}[c]{@{}c@{}}LLM-\\ Generated\end{tabular}}} & \multicolumn{1}{c|}{\textbf{\begin{tabular}[c]{@{}c@{}}Task-Oriented\\ Document-\\ Grounded\end{tabular}}} & \multicolumn{1}{c|}{\textbf{\cmark}} & \multicolumn{1}{c|}{\textbf{\cmark}} & \multicolumn{1}{c|}{\textbf{\cmark}} & \textbf{8.8k} & \textbf{7.6} & \textbf{16.8} & \textbf{62.1}
\end{tabular}
}

  \caption{Comparison of \ourdata to other datasets that provide related annotations. \ourdata is comparable to other synthetic datasets generated by large language models (LLMs) in terms of avg. turn and utterance length. It also has a higher lexical diversity than many of the crowdsourced datasets\protect\footnotemark.}
  \label{tab:comparison_lexical_diversity}
\end{table*}

\section{Introduction}
\begin{figure}[t!]
\centering
  \includegraphics[width=\linewidth]{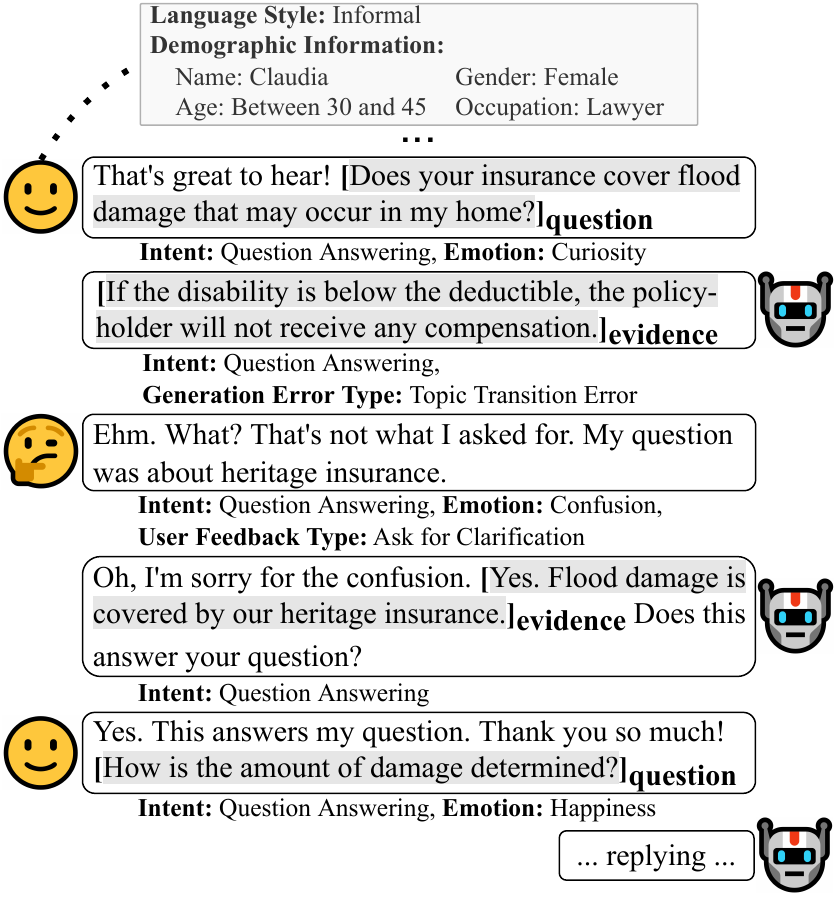}
  \caption{A feedback dialogue from \ourdata, annotated with user emotions and implicit user feedback (generation error and user feedback types).}
  \label{fig:example_dialog}
\end{figure}
Implicit user feedback~\cite{xu-etal-2023-learning, veron2020evaluate, hancock-etal-2019-learning}, such as clarification questions, user emotions~\cite{hwang-etal-2023-aligning, rashkin-etal-2019-towards, hsu-etal-2018-emotionlines} and demographic information~\cite{lee-etal-2022-personachatgen, zhang-etal-2018-personalizing}, such as age or language style, are promising sources for improving the accuracy and user engagement of responses generated by dialogue systems. For example, in the second utterance of Figure~\ref{fig:example_dialog}, the system generates a response unrelated to the user's question, which affects her emotional state. She asks the system for clarification, getting a more satisfying response. This makes her happy and she continues the conversation. However, we do not know to what extent the generated response contributes to achieving the user's goal and reflects the underlying knowledge source. This is commonly referred to as task completion and factual consistency. Both are important criteria for task-oriented and document-grounded dialogue systems~\cite{nekvinda-dusek-2021-shades, honovich-etal-2021-q2, budzianowski-etal-2018-multiwoz}, but the impact of implicit user feedback, emotions and demographic information on them is an open research question. 

To address this gap, we introduce the \ourdata dataset. Following recent research that includes information-seeking in task-oriented dialogues~\cite{taranukhin2024air, braunschweiler-etal-2023-evaluating, feng-2021-dialdoc, campos-etal-2020-doqa}, e.g., for handling multi-domain scenarios, \ourdata provides annotations for required knowledge documents and is the first English task-oriented and document-grounded dialogue dataset annotated with implicit user \textbf{\underline{F}}eedback, \textbf{\underline{E}}motions and \textbf{\underline{D}}emographic \textbf{\underline{I}}nformation. \ourdata allows us to investigate the impact of this information on task completion and the factual consistency of responses generated by dialogue systems, for which we use  Flan-T5~\cite{flan_t5}, GPT-2~\cite{gpt2} and Llama 2~\cite{touvron2023llama2} in this work. 
\footnotetext{We used the Python package \href{https://github.com/kristopherkyle/lexical_diversity}{lexical-diversity} v0.1.1 for calculation (last accessed 04 January 2024), which implements the approach proposed by \newcite{mccarthy2010mtld}.}
We use GPT-3.5\footnote{We used GPT-3.5-Turbo (\href{https://platform.openai.com/docs/models/gpt-3-5}{OpenAI GPT-3.5 Model Page}, last accessed on 02 January 2024). The model is based on \newcite{instruct_gpt}. The data was generated between March and June 2023.} to generate and annotate the training and validation data for FEDI. We recruit humans to assess its quality and to collect a separate set of test dialogues. 
In summary, we provide these contributions:

\begin{enumerate}
    \item New experimental insights on the impact of learning from implicit user feedback, user emotions and demographic information, including task completion and factual consistency, and how humans perceive the responses generated by the resulting models.
    \item \ourdata, the first task-oriented and document-grounded dialogue dataset for learning from implicit user feedback, emotions and demographic information. It is comparable to other related datasets in terms of size, lexical diversity and dialogue length (see Table~\ref{tab:comparison_lexical_diversity}).
    \item A framework for generating and annotating task-oriented and document-grounded feedback-annotated dialogue data. Our analysis provides insights into the quality of the generated dialogues annotations.
\end{enumerate}


\section{Related Work}\label{sec:realted_work}
Learning from user emotions and demographic information can improve generation accuracy and user engagement in dialogue systems~\cite{feng-etal-2022-emowoz, hsu-etal-2018-emotionlines, siddique, zhang-etal-2018-personalizing}. The same applies to implicit user feedback, which usually requires user interaction for data collection and continual learning~\cite{xu-etal-2023-learning, ung-etal-2022-saferdialogues, wang-etal-2019-incremental, hancock-etal-2019-learning}. Table~\ref{tab:comparison_lexical_diversity} provides a concise overview of the related datasets. None of them contains annotations for all three signals, and most of them were collected in resource-intensive crowdsourcing efforts. This is particularly complex in the context of user feedback~\cite{xu-etal-2023-learning, ung-etal-2022-saferdialogues} and no guarantee for quality~\cite{parmar-etal-2023-dont, yang-etal-2023-refgpt, thorn-jakobsen-etal-2022-sensitivity, prabhakaran-etal-2021-releasing}. Although LLMs are heavily dependent on detailed instructions and still tend to generate biased, hallucinated, or harmful data~\cite{yang-etal-2023-refgpt, Ji_2023, zhang2023sirens, malaviya2023expertqa}, recent works suggest these models, especially GPT-3.5, as a more efficient approach to generate dialogue data~\cite{stricker-paroubek-2024-chitchat, kim-etal-2023-soda, li-etal-2023-normdial, lee-etal-2022-personachatgen}.

To investigate the impact of implicit user feedback, emotions and demographic information in task-oriented and document-grounded dialogues, we create \ourdata by combining the best of both worlds. We use GPT-3.5 to generate training and validation data and recruit human annotators for dialogue quality assessment, annotation curation, and collection of a separate set of test dialogues.

\section{\ourdata}\label{sec:feedback}
Self-service terminals are increasingly common in the service sector, including postal services, access controls (e.g., to security-critical areas or in the hospitality industry), or customer service~\cite{abbate2024104568, applications_of_service_robots, receptionist_information_kiosk}. They implement specific workflows to serve customers and support employees. \ourdata covers four use cases from these domains. For postal sevices, we include customer support for parcel shipping and topping up a prepaid SIM card. For receptionist and insurance services, we include one use case each,  i.e., access control (the reception and registration of new visitors in office buildings) and question answering (in the context of financial topics and pet, health and heritage insurance). The question answering dialogues are additionally annotated with knowledge documents. Appendix~\ref{appendix:task_descriptions} describes the tasks in more detail, including slots (information required for task completion), intents and document sources.

\paragraph{Implicit User Feedback ($GE$, $F$)}
We use the taxonomies proposed by \newcite{petrak-etal-2023-learning} to generate and annotate generation errors ($GE$) and subsequent implicit user feedback ($F$). They distinguish ten types of generation errors. Nine of which are relevant for \ourdata, such as \emph{Attribute Error}, \emph{Factually Incorrect} or \emph{Lack of Sociality}. For implicit user feedback, they distinguish five types, e.g., \emph{Ask for Clarification}, \emph{Ignore and Continue} and \emph{Repeat or Rephrase}. Definitions, further details and examples can be found in Appendix~\ref{appendix:dataset_features}.

\paragraph{Demographic Information ($DI$)}
We consider gender, age, occupation, name, and language style as demographic information in this work. Overall, we distinguish 12 different language styles, such as formal, dialect and jargon, five demographic cohorts, ranging from \emph{Boomers} (born between 1952 and 1962) to \emph{Generation Alpha} (born between 2007 and 2016), a variety of $1,155$ occupations,  and $2,000$ names. We provide more details, including data sources in Appendix~\ref{appendix:dataset_features}.


%
%

\paragraph{User Emotions ($E$)}
Inspired by \newcite{hsu-etal-2018-emotionlines}, \newcite{rashkin-etal-2019-towards} and \newcite{kim-etal-2023-soda}, we derive a taxonomy of 11 emotions that are potentially relevant for our dialogue tasks, including \emph{Neutral}, three positive emotions (\emph{Curiosity}, \emph{Surprise} and \emph{Happiness}) and seven negative emotions (\emph{Confusion}, \emph{Frustration}, \emph{Fear}, \emph{Sadness}, \emph{Disgust}, \emph{Stress}, and \emph{Anger}).

\paragraph{Problem Formulation} We define a dialogue as a set of multiple turns $T$. Each turn consists of two utterances, a user utterance $U_t$ and a system utterance $S_t$. Given the dialogue context $C=[T_0, ..., T_{t-1}]$, and additional information $K$, the task is to predict the user intent $I_t$, generate belief state $B_t$ and system utterance $S_t$:
\begin{equation}
(I_t, B_t, S_t) = \text{generate}(K, C, U_t)
\end{equation}
Depending on whether knowledge from a document $D_t$ is required to generate $S_t$ or the user emotion $E_t$, demographic information $DI$, generation error $GE_t$, or implicit user feedback  $F_t$ should be considered, $K=\{D_t, DI, E_t, GE_t, F_t\}$. $DI$ includes the gender, age range, occupation, name, and language style of the user. Belief state $B_t$ includes the slots predicted for user utterance $U_t$.

\begin{figure*}[ht]
\centering
  \includegraphics[width=1.0\linewidth]{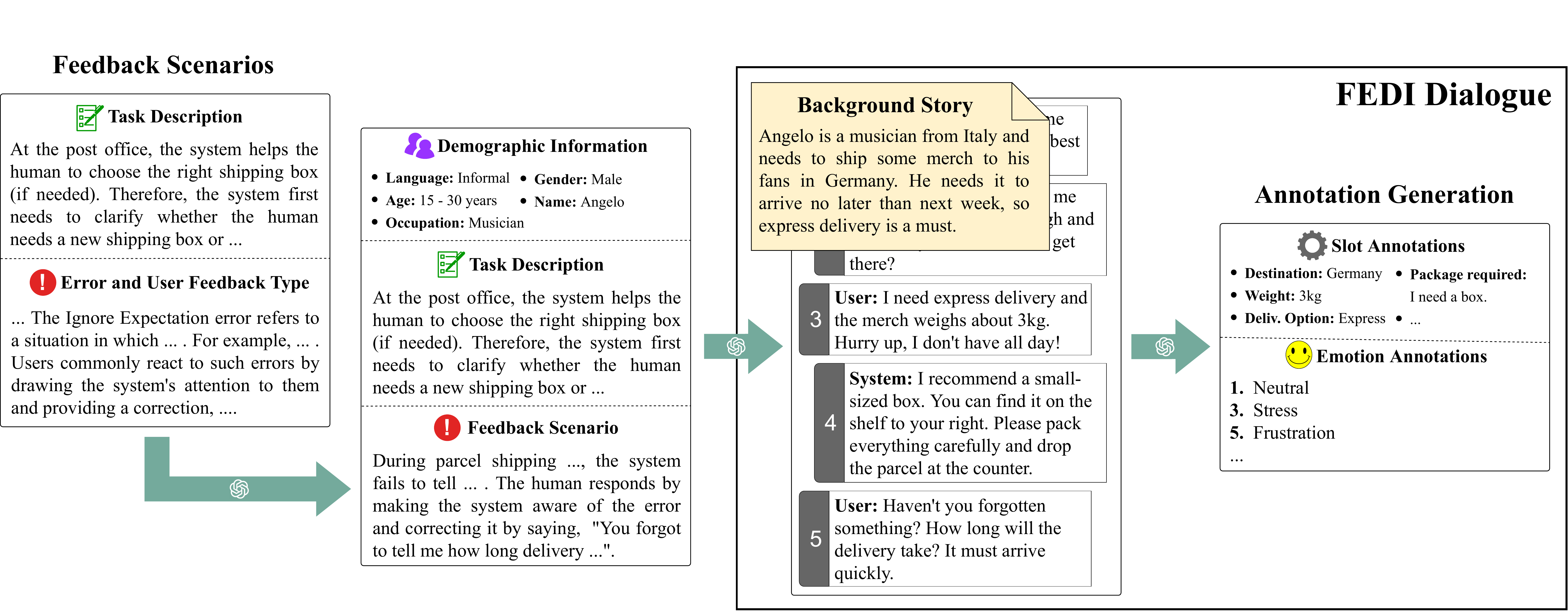}
  \caption{Overview of our framework for generating and annotating dialogues. \protect\scalerel*{\includegraphics{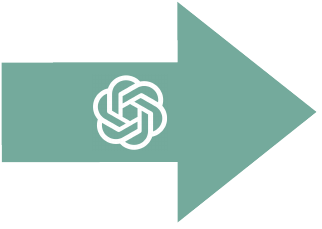}}{B} (the green arrow) symbolizes GPT-3.5. The generation of feedback dialogues requires feedback scenarios as additional source. For question answering dialogues, we include the respective documents in the task description.} 
  \label{fig:overview}
\end{figure*}

\section{Framework for Generating and Annotating Dialogues}\label{sec:dialog_generation}
Figure~\ref{fig:overview} gives an overview of our framework for generating and annotating dialogues. We distinguish feedback-free and feedback dialogues, i.e., dialogues that provide annotations for generation errors and implicit user feedback. For each step, we require GPT-3.5 to return the results in a predefined JSON scheme. If in one step the generation does not match this requirement, the whole dialogue is discarded. We provide more details, including the instructions used in this procedure, in Appendix~\ref{appendix:prompts}.

%
%

\subsection{General Approach to Dialogue Generation}\label{sub_sec:non_erroneous_dialogs}
The procedure is basically the same for feedback-free and feedback dialogues. It starts in the second box from the left in Figure~\ref{fig:overview}. We provide GPT-3.5 with randomly sampled demographic information for the user and a task description, which describes the flow of events to fulfill the task, including the role of the starting actor, i.e., user or system, and a randomly sampled list of documents in the case of question answering. Feedback dialogues require feedback scenarios as additional sources (Section~\ref{sub_sec:erroneous_dialogs}). We instruct the model to use the task description and demographic information to generate a background story to guide the conversation~\cite{lee-etal-2022-personachatgen, stricker-paroubek-2024-chitchat}, such as depicted in the center of the figure. We require the model to return the utterance-level annotations for intents (not included in the figure) and limit the dialogue to 13 turns, since we found that longer dialogues tend to deviate from the task description. We limit the length of background stories to five sentences to avoid them becoming a distraction.

%
%

\paragraph{Annotation Generation}
For slot annotations, we provide GPT-3.5 with the generated dialogue and the task description (including a list of all slots, possible values, and examples\footnote{We also tried to reduce API calls by combining dialogue and annotation generation but found that this does not produce reliable results.}). We instruct the model to only copy values from the dialogue and to return the annotations on utterance-level. For emotion annotations, we provide the model with the emotion taxonomy (instead of the task description) and instruct it to predict the emotion for each user utterance.

%
%

\subsection{Feedback Dialogues}\label{sub_sec:erroneous_dialogs}
\paragraph{Feedback Scenarios} A feedback scenario describes a generation error and the following implicit user feedback. Figure~\ref{fig:overview} shows an example in the second box from the left. For generation (first box), we provide GPT-3.5 with the task description and a list of randomly sampled generation error and implicit user feedback types. To ensure coherence, feedback scenarios must not be mutually exclusive and together form a story in the context of the task description. For each feedback dialogue, we generate three feedback scenarios that are then used as an additional source for dialogue generation\footnote{We generate all feedback scenarios for a dialogue at once, using a single API call.}.  

\paragraph{Feedback Dialogue Generation} For feedback dialogues, we instruct GPT-3.5 to consider each feedback scenario in three consecutive utterances in the generated dialogue: First, the system utterance with the generation error, e.g., \emph{Yes, I can help you send a parcel to Paris}. Then the subsequent user utterance, e.g., \emph{No, the destination is London, not Paris!}, which we consider as implicit user feedback. Finally, the following system utterance that addresses the user feedback, e.g., \emph{Apologies for the mistake. Thank you for correcting me. The destination is London, United Kingdom. Now, please provide me with the weight of the package}. We consider the generated dialogue as Version 1 and generate three additional versions of the same dialogue, each resolving one of the feedback scenarios.

\begin{figure}[ht]
\centering
  \includegraphics[width=0.9\linewidth]{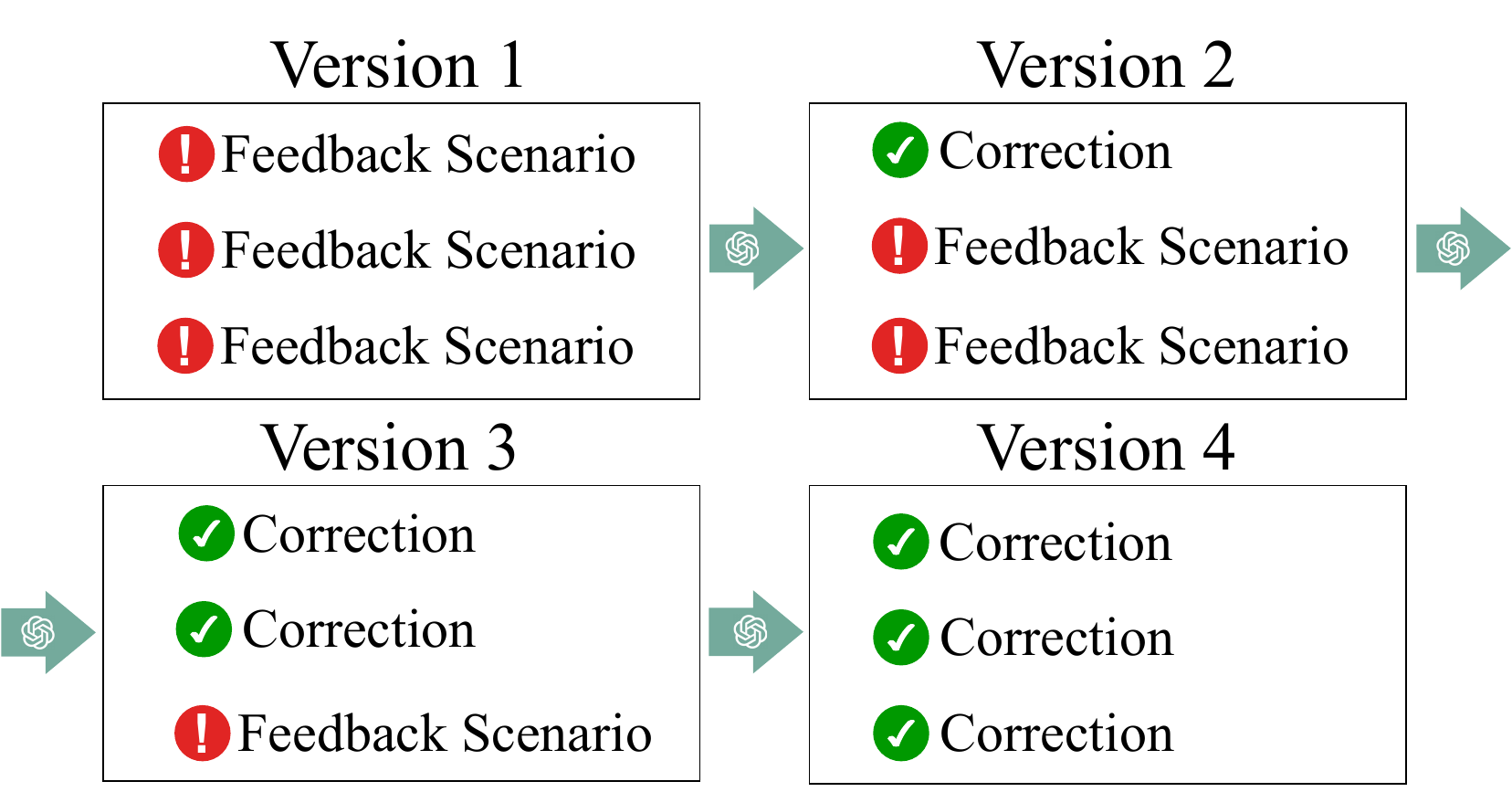}
  \caption{Feedback dialogue generation. Each version solves one of the feedback scenarios from Version 1. See Appendix~\ref{appendix:prompts} (Figure~\ref{fig:feedback_example}) for an example dialogue.} 
  \label{fig:feedback_dialogues}
\end{figure}

\paragraph{Resolving Feedback Scenarios} Figure~\ref{fig:feedback_dialogues} illustrates the idea. For each version, we first mask the affected system utterance and generate a replacement using the preceding dialogue context and task-specific information. Next, we drop the following two utterances, since they are directly related to the generation error. This way, the dialogue remains coherent and the conversation continues with the next regular user utterance\footnote{We experimented with different ideas for resolving feedback scenarios (see Appendix~\ref{appendix:prompts}), but the naive approach described here turned out to be the most reliable.}.
 We continue the process until all feedback scenarios have been resolved as in Version 4. For slot values, we only regenerate the annotations for the replaced system utterances in Version 2 to 4 and retain the other annotations from Version 1.


\section{\ourdata Analysis}\label{sec:analysis}
\ourdata consists of 8,852 dialogues, divided into 1,988 feedback-free dialogues, including 326 for testing, and 6,864 feedback dialogues (1,716 in four versions). The test dialogues were collected human-human by eight computer science students. In the following, we focus on the completeness of generated slot and intent annotations, the distribution of user emotions and the feedback scenarios represented in the dialogues. We provide additional statistical analysis in Appendix~\ref{appendix:additional_analysis}, including split sizes and the distribution of demographic information. Details on recruitment, salary, procedure, and our experiences and findings from collecting and annotating dialogue data with humans vs. LLMs can be found in Appendix~\ref{appendix:human_annotators}. 

%
%

\paragraph{Slot and Intent Annotations}\label{sub_sec:slot_intent_annotation}
Table~\ref{tab:slot_accuracies} shows the ratio of dialogues that provide all annotations for intent and slot values.
\begin{table}[htb]
  \centering
  \resizebox*{1.0\linewidth}{!}{
\begin{tabular}{lrrrrrr}
\multicolumn{1}{c}{\textbf{Task}} & \multicolumn{2}{c}{\textbf{\begin{tabular}[c]{@{}c@{}}Feedback-Free\end{tabular}}} & \multicolumn{4}{c}{\textbf{Feedback}} \\ \hline
\multicolumn{1}{c|}{\textbf{}} & \multicolumn{1}{c}{\textbf{Gen.}} & \multicolumn{1}{c|}{\textbf{Test}} & \multicolumn{1}{c}{\textbf{V1}} & \multicolumn{1}{c}{\textbf{V2}} & \multicolumn{1}{c}{\textbf{V3}} & \multicolumn{1}{c}{\textbf{V4}} \\ \hhline{=======}
\multicolumn{1}{l|}{\begin{tabular}[c]{@{}l@{}}Parcel\\ Shipping\end{tabular}} & 0.87 & \multicolumn{1}{r|}{0.51} & \multicolumn{1}{r}{0.74} & 0.72 & 0.70 & 0.70 \\ \hline
\multicolumn{1}{l|}{\begin{tabular}[c]{@{}l@{}}Top Up \\ SIM Card\end{tabular}} & 0.87 & \multicolumn{1}{r|}{0.51} & \multicolumn{1}{r}{0.74} & 0.72 & 0.71 & 0.69 \\ \hline
\multicolumn{1}{l|}{\begin{tabular}[c]{@{}l@{}}Access\\ Control\end{tabular}} & 0.86 & \multicolumn{1}{r|}{0.68} & \multicolumn{1}{r}{0.82} & 0.83 & 0.84 & 0.84 \\ \hline
\multicolumn{1}{l|}{\begin{tabular}[c]{@{}l@{}}Question\\ Answering\end{tabular}} & 0.99 & \multicolumn{1}{r|}{0.87} & \multicolumn{1}{r}{0.73} & 0.99 & 0.99 & 0.99
\end{tabular}}
  \caption{Ratio of dialogues that contain all annotations for related intent and slot values\protect\footnotemark. For feedback-free dialogues, we distinguish generated (Gen.) and test dialogues (Test). The feedback dialogues are divided into versions, i.e. Version 1 (v1) to Version 4 (V4).}
  \label{tab:slot_accuracies}
\end{table}
\footnotetext{Hallucinated slot values, i.e., slot annotations with a value that does not occur in the respective utterance, were small in number and are not considered in the results.}

We observe a difference between \emph{Gen.} and \emph{Test} in the feedback-free dialogues, as the slot values often depend on the background stories. For example, with parcel shipping, if the user already has a shipping box, details about available shipping boxes are negligible. Human annotators consider this and omit slots if they are not required~\cite{zang-etal-2020-multiwoz}. GPT-3.5 strictly follows our instructions, which include all slots as part of the task description. Question answering is less affected by this due to the more trivial annotation scheme (see Appendix~\ref{appendix:task_descriptions}). In the feedback dialogues, the generated corrections sometimes do not contain all the required slot values. This is expected, because these dialogues focus on learning how to handle errors and feedback situations. We provide more findings as part of our manual analysis in Section~\ref{sub_sec:human_curation_study}.

%
%

\begin{figure}[htb]
\centering
  \includegraphics[width=1.0\linewidth]{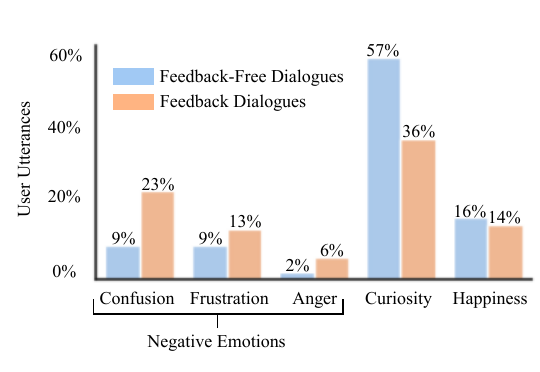}  
  \caption{Ratio of the most commonly observed user emotions in \ourdata (excluding the Neutral emotion).}
  \label{fig:emotions_grouped}
\end{figure}

\paragraph{Emotion Annotations}\label{sub_sec:nonverbal_verbal_feedback}
Figure~\ref{fig:emotions_grouped} shows the distribution of the five most common emotions observed in user utterances from both the feedback-free and feedback dialogues\footnote{We do not distinguish between generated and test dialogues here. We also leave out the neutral emotion as it is in general the most frequently observed emotion (40.5\% of all annotated emotions).}.

As expected, negative emotions are more common in feedback dialogues. \emph{Happiness} in feedback dialogues is mostly observed when the system addresses the implicit user feedback. This is similar for \emph{Curiosity}, although we also observe this emotion when the system suddenly changes the topic. While this emotion fits most dialogue context, it can also be the result of insufficient information in the emotion annotation instruction, as we only use the dialogue context as additional information and no further examples (see Appendix~\ref{appendix:prompts}).

%
%

\begin{figure}[htbp]
    \centering
  \includegraphics[width=1.0\linewidth]{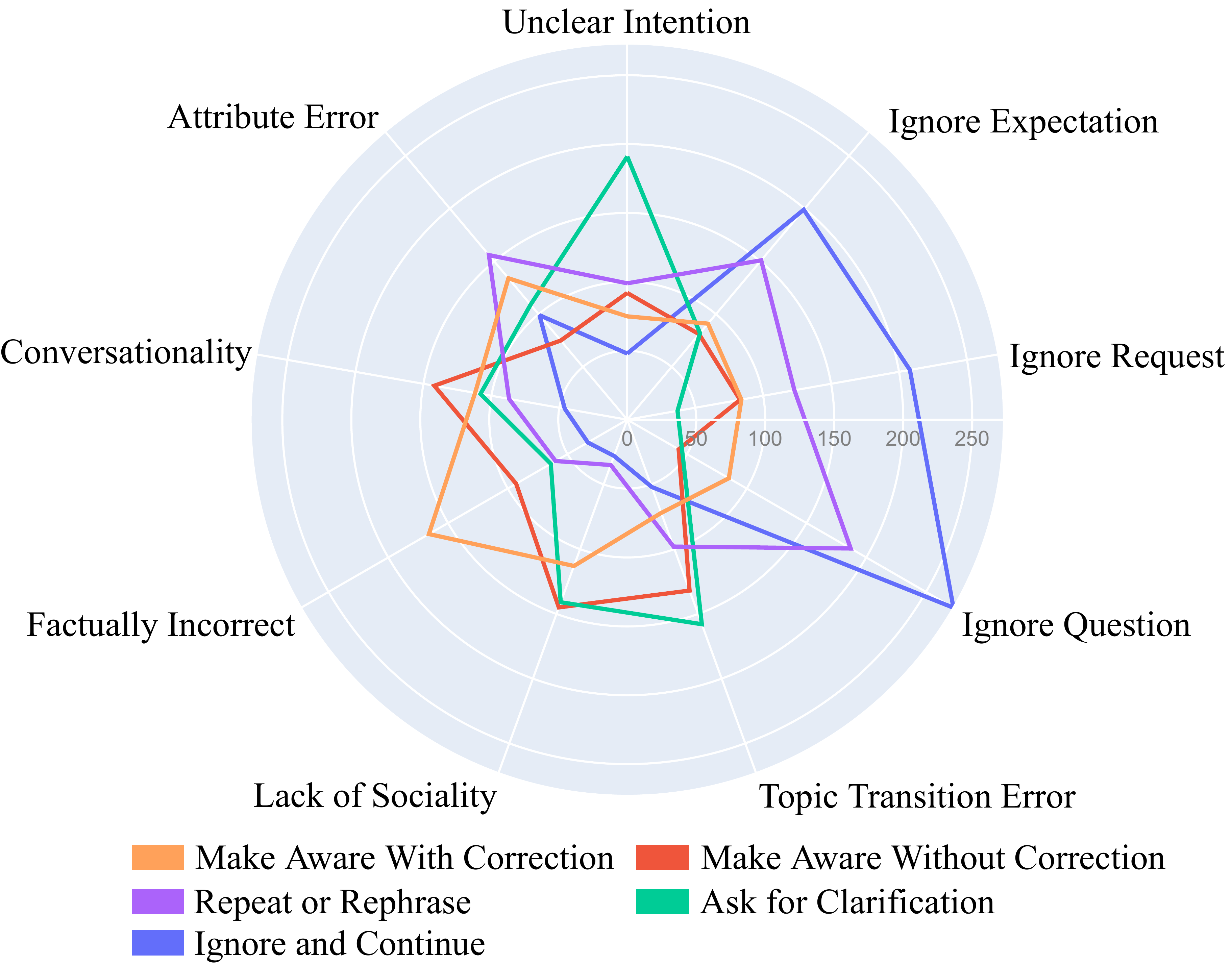}
  \caption{Distribution of user feedback types in relation to generation error types in feedback scenarios.}
  \label{fig:error_user_reaction_types_main}
\end{figure}

\paragraph{Feedback Scenarios}\label{sub_sec:verbal_feedback}
Figure~\ref{fig:error_user_reaction_types_main} shows the distribution of user feedback types in relation to generation error types represented in the feedback scenarios of the feedback dialogues.

It shows that our approach for generating feedback scenarios mostly resulted in meaningful combinations of generation error and user feedback types. For example, \emph{Factually Incorrect} is mostly addressed by \emph{Make Aware with Correction}. \emph{Unclear Intention} and \emph{Attribute Error} are frequently addressed by \emph{Ask for Clarification} and \emph{Repeat or Rephrase}. The latter one is also frequently observed in combination with \emph{Ignore Question} and \emph{Ignore Expectation} errors, although \emph{Ignore and Continue} is the most frequent user feedback to these generation error types.

\section{Quality Control for \ourdata}\label{sub_sec:human_curation_study}
We asked two participants from our test data collection to assess and curate the intent, slot and emotion annotations in 480 feedback-free dialogues and the generation error and implicit user feedback type annotations in 380 feedback dialogues (see Appendix~\ref{appendix:procedure_data_curation} for the procedure). The dialogues were randomly sampled from the train and dev splits of \ourdata. We used INCEpTION~\cite{klie-etal-2018-inception} as a platform for this study. We calculate the inter-annotator agreement (IAA) using Krippendorff's Alpha~\cite{krippendorff} with a nominal weighting function (as provided in the platform). Table~\ref{tab:human_curation_details} shows the results\footnote{Overall, 26 dialogues were reported as off-topic (13/480 feedback-free and 13/380 feedback). They are not considered in these results. The curated dialogues were not considered in our experiments but are included as a separate set in the published dataset.}.

\begin{table}[htb]
  \centering
  \resizebox*{0.95\linewidth}{!}{
\begin{tabular}{clrrr}
\multicolumn{1}{l}{} & \multicolumn{1}{c}{\textbf{Annotation Type}} & \multicolumn{1}{c}{\textbf{Missing}} 
& \multicolumn{1}{c}{\textbf{Changed}} 
& \multicolumn{1}{l}{\textbf{IAA}} \\ \hhline{=====}
\multicolumn{1}{l|}{\multirow{3}{*}{\begin{tabular}[c]{@{}l@{}}Feedback-Free\\ Dialogues\end{tabular}}} & \multicolumn{1}{l|}{Intent} & 0.06 
& 0.35 
& 0.90 \\ \cline{2-5} 
\multicolumn{1}{c|}{} & \multicolumn{1}{l|}{Slot Values} & 0.56 
& 0.19 
& 0.83 \\ \cline{2-5}
\multicolumn{1}{c|}{} & \multicolumn{1}{l|}{User Emotions} & 0.02 
& 0.81 
& 0.91 \\ \hhline{=====}
\multicolumn{1}{l|}{\multirow{2}{*}{\begin{tabular}[c]{@{}l@{}}Feedback\\ Dialogues\end{tabular}}} & \multicolumn{1}{l|}{Generation Error Type} & 0.16 
& 0.36 
& 0.97 \\ \cline{2-5} 
\multicolumn{1}{c|}{} & \multicolumn{1}{l|}{User Feedback Type} & 0.16 
& 0.34 
& 0.89
\end{tabular}}  
  \caption{The ratio of dialogues with at least one missing or changed annotation in our human curation study.}
  \label{tab:human_curation_details}
\end{table}

Overall, the ratio of dialogues with at least one missing annotation is rather low, except for slot annotations. We found that most of them are parcel shipping dialogues, which has a comparatively complex annotation scheme (see Appendix~\ref{appendix:task_descriptions}). A detailed analysis revealed that an average of 1.8 annotations were added to these dialogues. For the dialogues with at least one changed annotation, annotators reported that in many of these cases placeholders, e.g., the slot name put in brackets ([shipping\_box\_name]), were used instead of the slot values from the dialogues. We attribute this to our observation from Section~\ref{sec:analysis} (GPT-3.5 strictly follows the slot annotation scheme, even if the values are not in the dialogue). Emotions, whose perception is very subjective, are the most frequently changed annotation type (on average 2.09 times per affected dialogue), whereby the originally annotated emotion was sometimes not part of our taxonomy (hallucination). We provide further analysis of the impact of human curation on data quality in Appendix~\ref{appendix:curation_further_analysis}.

\begin{table*}[ht]
  \centering
  \resizebox*{0.9\linewidth}{!}{\begin{tabular}{llrrrrrrrrr}
    \multicolumn{1}{c}{\textbf{}} &  \multicolumn{1}{c}{\textbf{Experiment}} & \multicolumn{4}{c}{\textbf{Task Completion}} & \multicolumn{2}{c}{\textbf{Quality}} & \multicolumn{3}{c}{\textbf{Generation Accuracy}} \\ \hline
     & \multicolumn{1}{l}{} & \multicolumn{1}{c}{\textbf{Inform}} & \multicolumn{1}{c}{\textbf{Success}} & \multicolumn{1}{c}{\textbf{Intent Acc.}} & \multicolumn{1}{c|}{\textbf{Slot Acc.}} & \multicolumn{1}{c}{\textbf{Q²}} & \multicolumn{1}{c|}{\textbf{Toxicity}} & \multicolumn{1}{c}{\textbf{F1}} & \multicolumn{1}{c}{\textbf{BLEU}} & \multicolumn{1}{c|}{\textbf{BertScore}} \\ \hhline{===========}
     
    \multicolumn{1}{l|}{\multirow{4}{*}{\begin{tabular}[c]{@{}l@{}}Flan-T5 \\ Feedback-Free\end{tabular}}} & \multicolumn{1}{l|}{Flan-T5} & 86.7 & 85.9 & 54.8 & \multicolumn{1}{r|}{60.9} & 52.7 & \multicolumn{1}{r|}{0.02} & 45.0 & 20.0 & \multicolumn{1}{r|}{88.3} \\ \hhline{~|==========|}
    
    \multicolumn{1}{c|}{}  & \multicolumn{1}{l|}{\textbf{\quad+User Emotions}} & \textbf{\textcolor{negative_color}{83.9}} & \textbf{\textcolor{negative_color}{83.2}} & \textbf{\textcolor{positive_color}{61.2}} & \multicolumn{1}{r|}{\textbf{\textcolor{negative_color}{58.3}}} & \textbf{\textcolor{positive_color}{57.5}} & \multicolumn{1}{r|}{\textbf{0.02}} &\textbf{\textcolor{positive_color}{46.7}} & \textbf{\textcolor{positive_color}{21.0}} & \multicolumn{1}{r|}{\textbf{88.9}} \\ \cline{2-11}
    
    \multicolumn{1}{c|}{}  & \multicolumn{1}{l|}{\quad+Demographic Info.} & 87.0 & 86.0 & \textcolor{negative_color}{33.5} & \multicolumn{1}{r|}{\textcolor{negative_color}{29.3}} & \textcolor{positive_color}{54.5} & \multicolumn{1}{r|}{0.03} & \textcolor{negative_color}{43.2} & \textcolor{negative_color}{18.4} & \multicolumn{1}{r|}{87.7} \\ \cline{2-11}
    
    \multicolumn{1}{c|}{}  & \multicolumn{1}{l|}{\begin{tabular}[c]{@{}l@{}}\quad+User Emotions\\ \quad+Demographic Info.\end{tabular}} & \textcolor{negative_color}{85.3} & 85.1 & \textcolor{negative_color}{43.9} & \multicolumn{1}{r|}{\textcolor{negative_color}{36.7}} & \textcolor{positive_color}{56.4} & \multicolumn{1}{r|}{0.02} & 44.2 & 19.1 & \multicolumn{1}{r|}{88.1}\\ \hline
    
    \multicolumn{1}{l|}{\multirow{3}{*}{\begin{tabular}[c]{@{}l@{}}Feedback\end{tabular}}}  & \multicolumn{1}{l|}{\quad+Generation Error} & \multicolumn{1}{r}{\textcolor{positive_color}{96.8}} & \multicolumn{1}{r}{\textcolor{positive_color}{92.7}} & \multicolumn{1}{r}{\textcolor{positive_color}{72.5}} & \multicolumn{1}{r|}{\textcolor{positive_color}{76.7}} & \textcolor{positive_color}{56.9} & \multicolumn{1}{r|}{0.02} & \multicolumn{1}{r}{\textcolor{negative_color}{41.4}} & \multicolumn{1}{r}{19.8} & \multicolumn{1}{r|}{87.8} \\ \cline{2-11}
    
    \multicolumn{1}{c|}{}  & \multicolumn{1}{l|}{\quad+User Feedback} & \multicolumn{1}{r}{\textcolor{positive_color}{96.6}} & \multicolumn{1}{r}{\textcolor{positive_color}{94.1}} & \multicolumn{1}{r}{\textcolor{positive_color}{69.0}} & \multicolumn{1}{r|}{\textcolor{positive_color}{76.2}} &  \textcolor{positive_color}{56.3} & \multicolumn{1}{r|}{0.02} & \multicolumn{1}{r}{\textcolor{negative_color}{41.3}} & \multicolumn{1}{r}{19.3} & \multicolumn{1}{r|}{87.6} \\ \cline{2-11}
    
    \multicolumn{1}{c|}{}  & \multicolumn{1}{l|}{\textbf{\begin{tabular}[c]{@{}l@{}}\quad+Generation Error\\     \quad+User Feedback\end{tabular}}} & \multicolumn{1}{r}{\textbf{\textcolor{positive_color}{96.9}}} & \multicolumn{1}{r}{\textbf{\textcolor{positive_color}{95.3}}} & \multicolumn{1}{r}{\textbf{\textcolor{positive_color}{83.5}}} & \multicolumn{1}{r|}{\textbf{\textcolor{positive_color}{77.2}}} & \textbf{\textcolor{positive_color}{60.2}} & \multicolumn{1}{r|}{\textbf{0.02}} & \multicolumn{1}{r}{\textbf{44.4}} & \multicolumn{1}{r}{\textbf{\textcolor{positive_color}{22.1}}} & \multicolumn{1}{r|}{\textbf{\textcolor{negative_color}{88.2}}} \\ \hhline{===========}
    
    \multicolumn{1}{l|}{\multirow{4}{*}{\begin{tabular}[c]{@{}l@{}}GPT-2 \\ Feedback-Free\end{tabular}}} & \multicolumn{1}{l|}{GPT-2} & 88.3 & 81.6 & 78.7 & \multicolumn{1}{r|}{69.6} & 28.1 & \multicolumn{1}{r|}{0.02} & 34.9 & 10.4 & \multicolumn{1}{r|}{87.1} \\ \hhline{~|==========|}
    
    \multicolumn{1}{c|}{}  & \multicolumn{1}{l|}{\quad+User Emotions} & \textcolor{negative_color}{84.1} & \textcolor{positive_color}{83.8} & \textcolor{negative_color}{75.4} & \multicolumn{1}{r|}{\textcolor{negative_color}{67.3}} & \textcolor{negative_color}{26.7} & \multicolumn{1}{r|}{0.02} & 35.1 & 10.4 & \multicolumn{1}{r|}{87.1} \\ \cline{2-11}
    
    \multicolumn{1}{c|}{}  & \multicolumn{1}{l|}{\quad+Demographic Info.} & \textcolor{negative_color}{80.2} & \textcolor{negative_color}{80.2} & \textcolor{negative_color}{69.3} & \multicolumn{1}{r|}{\textcolor{negative_color}{57.5}} & \textcolor{negative_color}{26.3} & \multicolumn{1}{r|}{0.02} & 34.6 & 10.4 & \multicolumn{1}{r|}{87.1} \\ \cline{2-11}
    
    \multicolumn{1}{c|}{}  & \multicolumn{1}{l|}{\textbf{\begin{tabular}[c]{@{}l@{}}\quad+User Emotions\\ \quad+Demographic Info.\end{tabular}}} & \textbf{\textcolor{negative_color}{85.1}} & \textbf{\textcolor{positive_color}{84.8}} & \textbf{\textcolor{negative_color}{71.6}} & \multicolumn{1}{r|}{\textbf{\textcolor{negative_color}{66.7}}} & \textbf{\textcolor{positive_color}{29.2}} & \multicolumn{1}{r|}{\textbf{0.02}} & \textbf{\textcolor{positive_color}{36.0}} & \textbf{\textcolor{positive_color}{11.4}} & \multicolumn{1}{r|}{\textbf{87.3}} \\ \hline
    
    \multicolumn{1}{l|}{\multirow{3}{*}{\begin{tabular}[c]{@{}l@{}}Feedback\end{tabular}}} & \multicolumn{1}{l|}{\quad+Generation Error} & \multicolumn{1}{r}{\textcolor{positive_color}{92.4}} & \multicolumn{1}{r}{\textcolor{positive_color}{91.7}} & \multicolumn{1}{r}{\textcolor{positive_color}{84.3}} & \multicolumn{1}{r|}{\textcolor{positive_color}{79.3}} & \textcolor{positive_color}{30.9} & \multicolumn{1}{r|}{0.02} & \multicolumn{1}{r}{\textcolor{negative_color}{29.2}} & \multicolumn{1}{r}{\textcolor{negative_color}{8.0}} & \multicolumn{1}{r|}{86.2} \\ \cline{2-11}
    
    \multicolumn{1}{c|}{}  & \multicolumn{1}{l|}{\quad+User Feedback} & \multicolumn{1}{r}{\textcolor{positive_color}{98.9}} & \multicolumn{1}{r}{\textcolor{positive_color}{96.5}} & \multicolumn{1}{r}{\textcolor{positive_color}{83.0}} & \multicolumn{1}{r|}{\textcolor{positive_color}{80.3}} & \textcolor{positive_color}{32.3} & \multicolumn{1}{r|}{0.02} & \multicolumn{1}{r}{\textcolor{negative_color}{30.0}} & \multicolumn{1}{r}{\textcolor{negative_color}{8.3}} & \multicolumn{1}{r|}{86.3}\\ \cline{2-11}
    
    \multicolumn{1}{c|}{}  & \multicolumn{1}{l|}{\textbf{\begin{tabular}[c]{@{}l@{}}\quad+Generation Error\\     \quad+User Feedback\end{tabular}}} & \multicolumn{1}{r}{\textbf{\textcolor{positive_color}{94.7}}} & \multicolumn{1}{r}{\textbf{\textcolor{positive_color}{93.3}}} & \multicolumn{1}{r}{\textbf{\textcolor{positive_color}{88.0}}} & \multicolumn{1}{r|}{\textbf{\textcolor{positive_color}{80.8}}} & \textbf{\textcolor{positive_color}{35.5}} & \multicolumn{1}{r|}{\textbf{0.01}} & \multicolumn{1}{r}{\textbf{\textcolor{negative_color}{30.3}}} & \multicolumn{1}{r}{\textbf{\textcolor{negative_color}{9.7}}} & \multicolumn{1}{r|}{\textbf{\textcolor{negative_color}{86.4}}} \\ \hhline{===========}
    
    \multicolumn{1}{l|}{\multirow{4}{*}{\begin{tabular}[c]{@{}l@{}}Llama 2 \\ Feedback-Free\end{tabular}}} & \multicolumn{1}{l|}{Llama 2} & 85.9 & 81.2 & 37.6 & \multicolumn{1}{r|}{39.2} & 28.3 & \multicolumn{1}{r|}{0.02} & 29.3 & 7.1 & \multicolumn{1}{r|}{86.1} \\ \hhline{~|==========|}
    
    \multicolumn{1}{c|}{}  & \multicolumn{1}{l|}{\textbf{\quad+User Emotions}} & \textbf{\textcolor{positive_color}{89.3}} & \textbf{\textcolor{positive_color}{85.3}} & \textbf{\textcolor{positive_color}{40.2}} & \multicolumn{1}{r|}{\textbf{\textcolor{positive_color}{41.3}}} & \textbf{\textcolor{negative_color}{18.7}} & \multicolumn{1}{r|}{\textbf{0.01}} & \textbf{\textcolor{positive_color}{36.3}} & \textbf{\textcolor{positive_color}{14.9}} & \multicolumn{1}{r|}{\textbf{85.4}} \\ \cline{2-11}
    
    \multicolumn{1}{c|}{}  & \multicolumn{1}{l|}{\quad+Demographic Info.} & 85.6 & \textcolor{positive_color}{82.5} & 37.1 & \multicolumn{1}{r|}{40.1} & \textcolor{negative_color}{21.3} & \multicolumn{1}{r|}{0.02} & \textcolor{positive_color}{33.8} & \textcolor{negative_color}{4.5} & \multicolumn{1}{r|}{86.5} \\ \cline{2-11}
    
    \multicolumn{1}{c|}{} & \multicolumn{1}{l|}{\begin{tabular}[c]{@{}l@{}}\quad+User Emotions\\ \quad+Demographic Info.\end{tabular}} & 86.7 & \textcolor{positive_color}{87.9} & \textcolor{positive_color}{41.4} & \multicolumn{1}{r|}{39.6} & \textcolor{negative_color}{20.6} & \multicolumn{1}{r|}{0.03} & 28.8 & \textcolor{negative_color}{5.6} & \multicolumn{1}{r|}{\textcolor{negative_color}{81.3}} \\ \hline
    
    \multicolumn{1}{l|}{\multirow{3}{*}{\begin{tabular}[c]{@{}l@{}}Feedback\end{tabular}}}  & \multicolumn{1}{l|}{\quad+Generation Error} & \textcolor{positive_color}{93.1} & \textcolor{positive_color}{95.7} & \textcolor{positive_color}{54.8} & \multicolumn{1}{r|}{\textcolor{positive_color}{59.6}} & 29.1 & \multicolumn{1}{r|}{0.01} & \textcolor{negative_color}{24.1} & 7.9 & \multicolumn{1}{r|}{\textcolor{negative_color}{77.4}} \\ \cline{2-11}
    
    \multicolumn{1}{c|}{}  & \multicolumn{1}{l|}{\textbf{\quad+User Feedback}} & \textbf{\textcolor{positive_color}{94.9}} & \textbf{\textcolor{positive_color}{93.2}} & \textbf{\textcolor{positive_color}{63.5}} & \multicolumn{1}{r|}{\textbf{\textcolor{positive_color}{70.1}}} & \textbf{\textcolor{negative_color}{27.1}} & \multicolumn{1}{r|}{\textbf{0.02}} & \textbf{\textcolor{negative_color}{24.5}} & \textbf{6.9} & \multicolumn{1}{r|}{\textbf{\textcolor{negative_color}{78.8}}} \\ \cline{2-11}
    
    \multicolumn{1}{c|}{}  & \multicolumn{1}{l|}{\begin{tabular}[c]{@{}l@{}}\quad+Generation Error\\ \quad+User Feedback\end{tabular}} & \textcolor{negative_color}{82.4} & \textcolor{positive_color}{83.6} & \textcolor{positive_color}{46.3} & \multicolumn{1}{r|}{\textcolor{positive_color}{47.2}} & \textcolor{positive_color}{33.5} & \multicolumn{1}{r|}{0.03} & \textcolor{negative_color}{25.0} & \textcolor{positive_color}{9.2} & \multicolumn{1}{r|}{\textcolor{negative_color}{80.1}} \\ \hhline{===========}

    \multicolumn{1}{l|}{Llama 2} & \multicolumn{1}{l|}{In-Context} & 10.6 & 12.4 & 8.6 & \multicolumn{1}{r|}{5.6} & 13.1 & \multicolumn{1}{r|}{0.02} & 11.3 & 3.7 & \multicolumn{1}{r|}{81.4}
    \end{tabular}}

  \caption{Results of our main experiments (averaged over three runs). The best-performing models are printed in \textbf{bold}. Differences from the baselines that are greater than $\pm1.0$ are colored \textcolor{positive_color}{green} and \textcolor{negative_color}{red}.}  \label{tab:main_experiments_automatic_evaluation}
\end{table*}
\section{Experiments and Results}\label{sec:experiments}
We conduct experiments using three models of different architecture and pretraining approaches, including Flan-T5~\cite{flan_t5} (780M), GPT-2~\cite{gpt2} (780M) and Llama 2~\cite{touvron2023llama2} (7B, plain pretrained version)\footnote{The model weights for \href{https://huggingface.co/google/flan-t5-large}{Flan-T5} and \href{https://huggingface.co/gpt2-large}{GPT-2} are available in the Huggingface Model Hub (last accessed 04 January 2024). Access to the weights for Llama 2 must be requested from \href{https://ai.meta.com/llama/}{Meta AI} (last acessed 04 January 2024).}. We first finetune the pretrained models using the feedback-free dialogues (baselines) and include the gold user emotions, demographic information and documents as part of the input sequences. We use the gold annotations of these signals to avoid bias from external components, such as emotion classifiers or document retrievers. For Llama 2, we only finetune the LoRA~\cite{hu2022lora} weights. We also provide in-context results for this model. We use the best feedback-free models for experiments with the feedback dialogues, in which we include the generation error and the user feedback utterance in the input sequence. We provide additional details in the Appendix, including hyperparameters and data configuration for feedback training (\ref{appendix:hyperparameters}) and input sequences (\ref{appendix:input_sequences}). We also provide results for experiments using Llama 3~\cite{llama3} (\ref{appendix:llama3}), which was published shortly after we had completed our main experiments, and continual learning from feedback data (\ref{appendix:continual_learning}).

\subsection{Evaluation Metrics}
For task completion, we use the Inform and Success~\cite{budzianowski-etal-2018-multiwoz} metrics and additionally measure the accuracy of the predicted intent and slot values. To measure the factual consistency of the generated responses in question answering, we use $\text{Q}^2$~\cite{honovich-etal-2021-q2}. Since the generation errors in \ourdata include social aspects (see Appendix~\ref{appendix:dataset_features}), we use Perspective API to measure their toxicity, and F1-Score, BLEU(-n)~\cite{papineni-etal-2002-bleu} and BertScore~\cite{bert-score}\footnote{For Inform and Success, we use the \href{https://github.com/Tomiinek/MultiWOZ_Evaluation}{implementation} from \newcite{nekvinda-dusek-2021-shades} as a reference. For $\text{Q}^2$, we use the reference implementation which is available in \href{https://github.com/orhonovich/q-squared}{GitHub}. \href{https://perspectiveapi.com/}{Perspective API} is a free-to-use service provided by Google and Jigsaw. We measure the F1-Score based on the overlapping tokens in target and prediction. For BLEU~\cite{papineni-etal-2002-bleu} and BertScore~\cite{bert-score}, we use the implementation from the HuggingFace \href{https://github.com/huggingface/evaluate}{evaluation library} v0.4.1 and with $n=4$ for BLEU (last access to all resources on 04 January 2024).} to measure their generation accuracy.

\subsection{Results}\label{sub_sec:second_set_of_experiments}
Table~\ref{tab:main_experiments_automatic_evaluation} shows the results achieved in the test dialogues. The feedback-free experiments show that including user emotions has the most positive impact. It improves the generation accuracy and factual consistency for Flan-T5~\cite{flan_t5} and GPT-2~\cite{gpt2} (here in combination with demographic information), and the generation accuracy and task completion for Llama 2~\cite{touvron2023llama2}. The feedback experiments show improved task completion and factual consistency ($Q^2$) across all models. Regarding toxicity, we did not observe any negative impact from including generation errors, except for some outliers in Flan-T5 and Llama 2 (see Appendix~\ref{appendix:impact_toxicity}).

\paragraph{On the Influence of Generation Errors and User Feedback} We assume that the generation errors and user feedback used in training served as negative examples, helping the models to learn to generate more accurate intents and slots and responses that better reflect the knowledge documents (see Appendix~\ref{appendix:negative_example} for examples). An analysis on dialogue type level (Appendix~\ref{appendix:performance_gap}) also shows an increased generation accuracy for Flan-T5 and GPT-2, but only for question answering. We assume this is due to the knowledge document, which as part of the input sequence regulates the influence of error and feedback information~\cite{xu2023improving, xu-etal-2023-learning, ung-etal-2022-saferdialogues}. The responses generated for the other tasks still fit the context but often deviate from the target sequences. For Flan-T5 and GPT-2, this is reflected in the F1-Score, which measures word overlapping and is more affected than BLEU~\cite{papineni-etal-2002-bleu} and BertScore~\cite{bert-score}. 

\paragraph{Additional Insights Regarding Llama 2} For Llama 2, we found that the generated responses often suffer from hallucinations (especially in the feedback-free dialogues). The reduced intent and slot accuracy also suggest a tendency towards hallucination here. We did not observe this in the experiments with Llama 3~\cite{llama3} (Appendix~\ref{appendix:llama3}). We also observe that the results of the finetuned Llama 2 models are significantly higher than those of the in-context experiment (we included the task descriptions along with examples in the instruction), emphasizing the importance of finetuning for task-oriented and knowledge-grounded dialogues~\cite{zhang2024transfertodgeneralizablechinesemultidomain, zhang-etal-2023-sgp}. 
 
\paragraph{Human Evaluation}\label{sub_sec:human_evaluation}
To investigate how humans perceive the impact of feedback training, we recruited 42 participants from Prolific\footnote{\href{https://www.prolific.com/}{Prolific} is a widely used crowdsourcing platform for scientific research (last accessed 08 May 2024).}. We asked them to rate the human likeness (Hum.), relevancy (Rel.), sociality (Soc.), engagement (Eng.), and factual consistency (Fact.) of the responses generated for 300 randomly sampled test dialogues in the feedback and feedback-free experiments highlighted in Table~\ref{tab:main_experiments_automatic_evaluation} (50 test dialogues from each experiment). We used a Likert scale from one to five for each attribute (with one as the lowest value). We received 40 valid submissions (we checked them manually in detail). Thus, each dialogue was rated by at least five participants. Table~\ref{tab:human_evaluation} shows the results. We provide more details on our rating scheme, annotator background and procedure in Appendix~\ref{appendix:crowdsourcing}.  

\begin{table}[htb]
  \centering
  \resizebox*{0.9\linewidth}{!}{
\begin{tabular}{lrrrrrl}
\multicolumn{1}{c}{\textbf{Model}} & \multicolumn{1}{c}{\textbf{\begin{tabular}[c]{@{}c@{}}Hum.\end{tabular}}} & \multicolumn{1}{c}{\textbf{Rel.}} & \multicolumn{1}{c}{\textbf{Soc.}} & \multicolumn{1}{c}{\textbf{Eng.}} & \multicolumn{1}{c}{\textbf{\begin{tabular}[c]{@{}c@{}}Fact.\end{tabular}}} & \multicolumn{1}{c}{\textbf{IAA}} \\ \hhline{=======}
\multicolumn{7}{c}{\textbf{Flan-T5}} \\ \hline
\multicolumn{1}{l|}{Feedb.-Free} & 3.41 & 4.12 & 4.66 & 3.56 & 4.12 & 0.25 \textsubscript{0.15} \\ \hline
\multicolumn{1}{l|}{Feedback} & \textbf{3.27} & \textbf{3.99} & 4.56 & 3.57 & 4.02 & 0.20 \textsubscript{0.16} \\ \hhline{=======}
\multicolumn{7}{c}{\textbf{GPT-2}} \\ \hline
\multicolumn{1}{l|}{Feedb.-Free} & 3.25 & 3.97 & 4.70 & 3.60 & 3.63 & 0.18 \textsubscript{0.05} \\ \hline
\multicolumn{1}{l|}{Feedback} & \textbf{4.02} & 3.88 & \textbf{4.58} & 3.52 & 3.64 & 0.21 \textsubscript{0.11} \\ \hhline{=======}
\multicolumn{7}{c}{\textbf{Llama 2}} \\ \hline
\multicolumn{1}{l|}{Feedb.-Free} & 3.0 & 3.31 & 4.49 & 3.16 & 2.74 & 0.25 \textsubscript{0.10} \\ \hline
\multicolumn{1}{l|}{Feedback} & \textbf{3.12} & \textbf{3.87} & \textbf{4.64} & \textbf{3.54} & \textbf{3.69} & 0.23 \textsubscript{0.10}
\end{tabular}
}
  \caption{Results of our human evaluation. If statistically significant, they are printed in bold. (independent two-sample t-test, $p \leq 0.05$). We calculate IAA across all metrics (standard deviation in subscript) using Krippendorff's Alpha~\cite{krippendorff}\protect\footnotemark.}
  \label{tab:human_evaluation}
\end{table}

\footnotetext{We used \href{https://docs.scipy.org/doc/scipy/index.html\#}{SciPy} v1.13.0 for the t-test (last accessed 08 April 2024). For Krippendorff's Alpha, we used \href{https://www.k-alpha.org/}{K-Alpha Calculator}~\cite{MARZI2024102545} (interval weighting).}

For Flan-T5~\cite{flan_t5} and GPT-2~\cite{gpt2}, annotators reported that the responses generated by the feedback models are more informative (which is not captured by the scores), but do not always cover the knowledge document as well as the responses from the feedback-free models (Flan-T5). They also reported them to be more direct and contain more counter-questions (which is actually desirable). This is often perceived as unfriendly, inattentive or disruptive and reflected in the slightly lower scores for relevancy and sociality (see Appendix~\ref{appendix:crowdsourcing_add_insights} for examples). For Llama 2~\cite{touvron2023llama2}, annotators reported some responses of the feedback-free model as illogical, unrelated to the dialogue context and factually incorrect. The responses generated by the feedback model were rated much higher, especially their relevancy and factual consistency. The IAA is rather low for most measures, which we attribute to their subjectivity and the diversity of annotators.

\section{Conclusion}
We introduce \ourdata, the first English task-oriented and document-grounded dialogue dataset annotated with implicit user feedback, user emotions and demographic information. Our analysis shows the usefulness of our framework for generating feedback-annotated dialogues from various domains, and that \ourdata is comparable to other related datasets. Our experiments show that learning from implicit user feedback improves task completion and factual consistency. Humans perceive the responses generated by feedback models as more informative (Flan-T5 and GPT-2), more relevant and more factually consistent (Llama 2). However, our results also show room for improvements in future work, e.g., the varying impact of learning from errors and feedback data on the generated responses and how they are perceived by humans.


%
%

\section{Limitations}
\paragraph{Taxonomies Used}
The taxonomies used for generating implicit user feedback, user emotions and demographic information only reflect subsets of possible values. They are not exhaustive. For example, we do not consider educational background for demographic information, or other emotions than those that seemed meaningful to us in the context of this work. Our taxonomy of user emotions may differ from the original works.

\paragraph{Synthetically Generated Data}
The training and validation dialogues in \ourdata were generated using GPT-3.5. They are the result of a scripted generation procedure, and there is a probability that some data is unfaithful, hallucinated, or even harmful~\cite{kumar-etal-2023-language, zhang2023sirens, malaviya2023expertqa}. Model-specific bias could also be a factor, which we haven't investigated further. Although our analysis shows that the generated annotations are of high quality and we have invested a lot of effort in developing the instructions used, some values may be incorrect or inappropriate in the context, e.g., in the case of user emotions. This also applies to the slot and intent annotations, where analysis has shown that human annotators can react more flexibly to the dialogue background. In contrast, GPT-3.5 focuses entirely on the instruction and tends to return placeholder values in case of doubt. In addition, some of these dialogues may seem artificial and unnatural due to potentially conflicting demographic information, e.g., language style contradicting age or occupation. The same applies to the feedback scenarios represented in the feedback dialogues. Some user feedback may appear unnatural and counterintuitive and may not even relate to the underlying generation error. Although we conducted a fairly extensive human curation study in which we did not observe these issues, a more thorough review of the whole dataset would be required for a final assessment. 

To solve feedback scenarios, we experimented with different ideas to incorporate the feedback into regenerating the affected system utterance. However, this led to unnatural and inconsistent dialogues, so we decided to use the naive approach described in the paper. As a result, the regenerated system utterances may not always directly reflect the feedback. 

\paragraph{Toxicity Through Learning From Generation Errors}
In our feedback experiments, we also use generation errors for learning. Since they also include social aspects, such as disrespectful or toxic response behavior, we used Perspective API to analyze the toxicity in generated responses. Although conspicuous responses were very rare, we acknowledge that the detector may not capture all the potentially harmful content. The generated data may also contain positive stereotypes, i.e., seemingly harmless words or patterns offensive to specific demographic groups, which are not marked by the detector~\cite{cheng-etal-2023-marked}. 

\paragraph{Human Evaluation}
We conducted the human evaluation as a crowdsourcing study and recruited 42 participants so that each dialogue was evaluated seven times. Some participants submitted their assessment far below the time limit, which is why we carefully checked each individual submission. Due to deviations from our rating scheme, we had to discard two submissions, which is why 100 of the 300 dialogues considered received fewer than seven ratings. Another limitation is the study design. We only considered the quality of the generated responses and not that of the generated slot and intent values. During the study, we found that our rating scheme has limitations as well. For example, hallucinations were not considered as a separate measure. Some annotators reported them as comments to the affected dialogues. However, the number was very small and we did not notice any additional cases when checking the submissions. 


\section*{Acknowledgments}

This work has been funded by the European Union under the Horizon Europe grant № 200009-57100412 (\href{https://sermasproject.eu/}{SERMAS}).

\bibliography{anthology,custom}

\appendix

\section{Task Descriptions}\label{appendix:task_descriptions}
In the following, we provide details on the tasks included in \ourdata and their slot values. Following \cite{budzianowski-etal-2018-multiwoz}, we distinguish requestable and informable slots, since this is necessary to calculate the task completion metrics in Section~\ref{sec:experiments}.

\paragraph{Post Office Services} \ourdata includes dialogues from two basic services provided in post offices, customer support for parcel shipping and topping up a prepaid SIM card. In customer support for parcel shipping, the task is to help the user choose the right shipping box and delivery option for their needs (given the weight of the goods to be sent and the destination). Topping up a prepaid SIM card is less of an advisory service since customers usually know how much they want to recharge, their telephone number, and which telephone provider they are with. Table~\ref{tab:parcel_choice_recharge_phone} lists the slots for each task.
\begin{table}[htb]
  \centering
  \resizebox*{\linewidth}{!}{
\begin{tabular}{lllll}
\multicolumn{1}{c}{\textbf{Slot Name}} & \multicolumn{1}{c}{\textbf{Informable}} & \multicolumn{1}{c}{\textbf{Requestable}} & \multicolumn{1}{c}{\textbf{Description}} \\ \hhline{=====}
\multicolumn{4}{c}{\textbf{Parcel Shipping}} \\ \hhline{=====}
\multicolumn{1}{l|}{Destination} & \multicolumn{1}{c|}{\cmark} & \multicolumn{1}{c|}{} & \begin{tabular}[c]{@{}l@{}}The city and country of \\ destination; national or \\ international.\end{tabular} \\ \cline{1-4} 
\multicolumn{1}{l|}{Weight} & \multicolumn{1}{c|}{\cmark} & \multicolumn{1}{c|}{} & \begin{tabular}[c]{@{}l@{}}The weight of the item to be\\ shipped, lightweight (up to \\ 5kg), average (up to 20kg),\\ heavy (up to 30kg).\end{tabular} \\ \cline{1-4} 
\multicolumn{1}{l|}{Package Required} & \multicolumn{1}{c|}{\cmark} & \multicolumn{1}{c|}{} & \begin{tabular}[c]{@{}l@{}}Whether or not a new\\ shipping box is required.\end{tabular} \\ \cline{1-4} 
\multicolumn{1}{l|}{Delivery Option} & \multicolumn{1}{c|}{\cmark} & \multicolumn{1}{c|}{} & \begin{tabular}[c]{@{}l@{}}Express or standard delivery.\end{tabular} \\ \hline
\multicolumn{1}{l|}{Country of Destination} & \multicolumn{1}{c|}{\cmark} & \multicolumn{1}{c|}{} & The destination country. \\ \cline{1-4} 
\multicolumn{1}{l|}{Shipping Box Name} & \multicolumn{1}{c|}{} & \multicolumn{1}{c|}{\cmark} & \begin{tabular}[c]{@{}l@{}}Name of the best suitable\\ shipping box (small-sized,\\ medium-sized, large-sized),\\ based on the weight of the\\ item to be sent.\end{tabular} \\ \cline{1-4} 
\multicolumn{1}{l|}{Shipping Box Description} & \multicolumn{1}{c|}{} & \multicolumn{1}{c|}{\cmark} & \begin{tabular}[c]{@{}l@{}}Brief description on why \\ the suggested shipping box\\ is a good choice.\end{tabular} \\ \cline{1-4} 
\multicolumn{1}{l|}{Shipping Procedure} & \multicolumn{1}{c|}{} & \multicolumn{1}{c|}{\cmark} & \begin{tabular}[c]{@{}l@{}}Description of the shipping\\ procedure (e.g., take the box\\ to the counter...).\end{tabular} \\ \cline{1-4} 
\multicolumn{1}{l|}{Shipping Time} & \multicolumn{1}{c|}{} & \multicolumn{1}{c|}{\cmark} & \begin{tabular}[c]{@{}l@{}}Expected delivery time, one\\ to three days for national,\\ four to six days for european,\\ and 3-4 weeks for international\\ deliveries.\end{tabular} \\ \hhline{=====}
\multicolumn{4}{c}{\textbf{Top Up SIM Card}} \\ \hhline{=====}
\multicolumn{1}{l|}{Phone Number} & \multicolumn{1}{c|}{\cmark} & \multicolumn{1}{c|}{} & \begin{tabular}[c]{@{}l@{}}Table or mobile phone \\ number with country code,\\ e.g., +39 XXX XXXXXXX.\end{tabular} \\ \cline{1-4} 
\multicolumn{1}{l|}{Phone Provider} & \multicolumn{1}{c|}{\cmark} & \multicolumn{1}{c|}{} & \begin{tabular}[c]{@{}l@{}}The phone provider, e.g. \\ Vodafone, POSTE Mobile, ... .\end{tabular} \\ \cline{1-4} 
\multicolumn{1}{l|}{Import Payment} & \multicolumn{1}{c|}{\cmark} & \multicolumn{1}{c|}{} & \begin{tabular}[c]{@{}l@{}}The recharge amount, e.g.,\\ 10 euro, 20 euro, 30 euro.\end{tabular} \\ \hline
\multicolumn{1}{l|}{Outcome Operation} & \multicolumn{1}{c|}{} & \multicolumn{1}{c|}{\cmark} &\begin{tabular}[c]{@{}l@{}}If all required information \\ were provided, the system \\ asks the user to insert the \\ card for payment.\end{tabular} \\ \hhline{=====}
\multicolumn{4}{c}{\textbf{Request Ticket}} \\ \hhline{=====}
\multicolumn{1}{l|}{Type of Service} & \multicolumn{1}{c|}{\cmark} & \multicolumn{1}{c|}{} & \begin{tabular}[c]{@{}l@{}}The type of service for\\ which the user wants to \\ request support, i.e., parcel \\ shipping or topping up a \\
prepaid SIM card.\end{tabular} \\ \hline
\multicolumn{1}{l|}{Ticket Number} & \multicolumn{1}{c|}{} & \multicolumn{1}{c|}{\cmark} & \begin{tabular}[c]{@{}l@{}}The ticket number generated\\ for the request.\end{tabular}
\end{tabular}}  
  \caption{Slot values for parcel shipping and topping up a prepaid SIM card.}
  \label{tab:parcel_choice_recharge_phone}
\end{table}
In modern post offices, service robots or other virtual agents are more commonly used to provide such services in a self-service manner. However, if something goes wrong, e.g., the shipping boxes are empty or the credit card was rejected, customers must have the option of requesting assistance from a human employee. In this case, the customer is asked to tell the agent the type of service they need assistance with. In turn, the agent creates a ticket for a human employee and returns the ticket number. We consider this as a kind of subtask to the other tasks (Request Ticket in Table~\ref{tab:parcel_choice_recharge_phone}) and do not evaluate it separately. 

\paragraph{Receptionist Services} For receptionist services, \ourdata only includes one task: access control. Table~\ref{tab:building_access} shows the slots for this task. 
\begin{table}[htb]
  \centering
  \resizebox*{\linewidth}{!}{
\begin{tabular}{llll}
\multicolumn{1}{c}{\textbf{Slot Name}} & \multicolumn{1}{c}{\textbf{Informable}} & \multicolumn{1}{c}{\textbf{Requestable}} & \multicolumn{1}{c}{\textbf{Description}} \\ \hhline{====}
\multicolumn{4}{c}{\textbf{Access Control}} \\ \hhline{====}
\multicolumn{1}{l|}{Guest Name} & \multicolumn{1}{c|}{\cmark} & \multicolumn{1}{c|}{} & \begin{tabular}[c]{@{}l@{}}The name of the person \\ who wants to access the \\ building.\end{tabular} \\ \cline{1-4} 
\multicolumn{1}{l|}{Host Name} & \multicolumn{1}{c|}{\cmark} & \multicolumn{1}{c|}{} & \begin{tabular}[c]{@{}l@{}}The name of the person\\ the guest wants to visit.\end{tabular} \\ \cline{1-4} 
\multicolumn{1}{l|}{Host E-Mail} & \multicolumn{1}{c|}{\cmark} & \multicolumn{1}{c|}{} & \begin{tabular}[c]{@{}l@{}}The E-Mail address of \\ the host.\end{tabular} \\ \cline{1-4} 
\multicolumn{1}{l|}{\begin{tabular}[c]{@{}l@{}}Alternative\\ Host Name\end{tabular}} & \multicolumn{1}{c|}{\cmark} & \multicolumn{1}{c|}{} & \begin{tabular}[c]{@{}l@{}}An alternative host, e.g.,\\ in case the host is not\\ available.\end{tabular} \\ \cline{1-4} 
\multicolumn{1}{l|}{\begin{tabular}[c]{@{}l@{}}Alternative\\ Host E-Mail\end{tabular}} & \multicolumn{1}{c|}{\cmark} & \multicolumn{1}{c|}{} & \begin{tabular}[c]{@{}l@{}}E-Mail address of the\\ alternative host.\end{tabular} \\ \cline{1-4} 
\multicolumn{1}{l|}{\begin{tabular}[c]{@{}l@{}}Meeting Date\\ and Time\end{tabular}} & \multicolumn{1}{c|}{\cmark} & \multicolumn{1}{c|}{} & \begin{tabular}[c]{@{}l@{}}Date and time of the \\ appointment.\end{tabular} \\ \cline{1-4} 
\multicolumn{1}{l|}{\begin{tabular}[c]{@{}l@{}}Meeting Room\\ Identifier\end{tabular}} & \multicolumn{1}{c|}{\cmark} & \multicolumn{1}{c|}{} & \begin{tabular}[c]{@{}l@{}}Unique identifier of the\\ room where the meeting\\ will take place.\end{tabular} \\ \hline
\multicolumn{1}{l|}{\begin{tabular}[c]{@{}l@{}}Verification\\ Call\end{tabular}} & \multicolumn{1}{c|}{} & \multicolumn{1}{c|}{\cmark} & \begin{tabular}[c]{@{}l@{}}The system can set up a\\ verification call to let the\\ host visually inspect the\\ guest and authorize \\ access.\end{tabular} \\ \cline{1-4} 
\multicolumn{1}{l|}{\begin{tabular}[c]{@{}l@{}}Confirmation\\ to Open Turn-\\ stile\end{tabular}} & \multicolumn{1}{c|}{} & \multicolumn{1}{c|}{\cmark} & \begin{tabular}[c]{@{}l@{}}This is a signal to the\\ system that controls the\\ turnstile to let the guest\\ enter.\end{tabular} \\ \cline{1-4} 
\multicolumn{1}{l|}{\begin{tabular}[c]{@{}l@{}}Add. Safety\\ Information\end{tabular}} & \multicolumn{1}{c|}{} & \multicolumn{1}{c|}{\cmark} & \begin{tabular}[c]{@{}l@{}}Any additional safety\\ information, e.g., related\\ to COVID-19.\end{tabular}
\end{tabular}}  
  \caption{Slot values for access control.}
  \label{tab:building_access}
\end{table}
It is an essential task in hotels, office buildings, or other facilities with restricted access. Visitors usually need to register at the reception desk before being allowed to enter. As of today, electronic access controls (EAC) are more common than reception desks, especially in the case of office buildings, and they are becoming increasingly intelligent. In our case, we focus on a scenario in which a visitor has an appointment with an employee in an office building. To access the building, the visitor needs to provide the EAC with information about the appointment, e.g., the name of the host, date and time, and the room number. The EAC can then decide to grant access or to call the host for confirming the visitor's identity. If necessary, the EAC can also provide additional safety information, e.g., hygiene guidelines. 

\paragraph{Customer Service in the Insurance Domain} 
For customer service in the insurance domain, we focus on question answering in the context of pet, health and heritage insurance, as well as bank transactions and account conditions. As a source, we use the insurance policies from POSTE Italiane, which are also available in English language\footnote{\href{https://www.posteitaliane.it/en/insurance-services.html}{POSTE Italiane Insurance Policies}, last accessed 13 January 2024.}. Table~\ref{tab:question_answering} lists the slots.
\begin{table}[htb]
  \centering
  \resizebox*{\linewidth}{!}{
\begin{tabular}{llll}
\multicolumn{1}{c}{\textbf{Slot Name}} & \multicolumn{1}{c}{\textbf{Informable}} & \multicolumn{1}{c}{\textbf{Requestable}} & \multicolumn{1}{c}{\textbf{Description}} \\ \hline
\multicolumn{4}{c}{\textbf{Question Answering}} \\ \hline
\multicolumn{1}{l|}{Question} & \multicolumn{1}{c|}{\cmark} & \multicolumn{1}{c|}{} & \begin{tabular}[c]{@{}l@{}}A question related to\\ one of the topics.\end{tabular} \\ \cline{1-4} 
\multicolumn{1}{l|}{Type of Bills} & \multicolumn{1}{c|}{\cmark} & \multicolumn{1}{c|}{} & \begin{tabular}[c]{@{}l@{}}If the user asks a question\\ regarding a specific pay-\\ ment slip, they need to\\ provide the type.\end{tabular} \\ \hline
\multicolumn{1}{l|}{Evidence} & \multicolumn{1}{c|}{} & \multicolumn{1}{c|}{\cmark} & \begin{tabular}[c]{@{}l@{}}The answer to the user's\\ question.\end{tabular} \\ \cline{1-4} 
\multicolumn{1}{l|}{\begin{tabular}[c]{@{}l@{}}Bill Form\\ Description\end{tabular}} & \multicolumn{1}{c|}{} & \multicolumn{1}{c|}{\cmark} & \begin{tabular}[c]{@{}l@{}}Description of the \\ specific payment form \\ (if the question was about\\ a payment form).\end{tabular} \\ \cline{1-4} 
\multicolumn{1}{l|}{\begin{tabular}[c]{@{}l@{}}Bill Form\\ Name\end{tabular}} & \multicolumn{1}{c|}{} & \multicolumn{1}{c|}{\cmark} & \begin{tabular}[c]{@{}l@{}}Name of the payment\\ form (if the question was\\ about a payment form).\end{tabular} \\ \cline{1-4} 
\multicolumn{1}{l|}{\begin{tabular}[c]{@{}l@{}}Bill Form\\ Payment\\ Procedure\end{tabular}} & \multicolumn{1}{c|}{} & \multicolumn{1}{c|}{\cmark} & \begin{tabular}[c]{@{}l@{}}Information on how to fill\\ the payment form (if the\\ question was about a pay-\\ ment form).\end{tabular}
\end{tabular}}  
  \caption{Slot values for question answering.}
  \label{tab:question_answering}
\end{table}
In the past, customers called their insurance agent or visited their local bank branch for all questions related to such topics. Today, it is more common to talk to chatbots or other service agents first and only in exceptional cases to human employees. Overall, we extracted 313 question-document pairs, i.e., questions paired with a paragraph that contains the answer, 19 for bank transactions, 93 for account conditions, 78 for health, 84 for heritage, and 39 for pet insurance, from the POSTE documents. 

\paragraph{Greeting}
In the prompts for dialogue generation (see Appendix~\ref{appendix:prompts}), we instruct GPT-3.5 to have a separate turn at the beginning and ending of a dialogue in which both roles greet each other by also considering the generated background story. However, we do not consider this as a separate task in the sense of this work and do not evaluate it separately.


\section{Dataset Features}\label{appendix:dataset_features}
In this section, we provide additional details on the demographic information and the error and user feedback types used to create \ourdata.

\paragraph{Demographic Information}
We distinguish 12 different language styles, including \emph{Their Age and Job}, \emph{Standard}, \emph{Colloquial}, \emph{Formal}, \emph{Gutter}, \emph{Polite}, \emph{Informal}, \emph{Regional Dialect}, \emph{Social Dialect}, \emph{Jargon}, \emph{Slang}, and \emph{Age}. For age ranges, we consider five demographic cohorts, including \emph{Boomers} (born between 1952 and 1962), \emph{Generation X} (born between 1962 and 1977), \emph{Millenials} (born between 1977 and 1992), \emph{Generation Z} (born between 1992 and 2007), and \emph{Generation Alpha} (born between 2007 and 2016). For occupations, we use a list of 1,155 job titles sampled from The Gazette\footnote{Available in \href{https://github.com/TheGazette/Transformations/blob/master/EnrichmentService/gazetteer/des_occupation.lst}{GitHub} (last accessed on 16 January 2024).}, including among others jobs from the fields of science and technology, education, arts and entertainment, healthcare, or manufacturing. As a source for the names, we use the list of the 2,000 most popular American baby names in 2010\footnote{Published by \href{https://www.babymed.com/baby-names/popular-1000-baby-names-year-2010}{babymed.com} (last accessed 12 February 2024).}. For each dialogue, we randomly sample a new value for each characteristic and apply simple plausibility checks, e.g., a person from \emph{Generation Alpha} can only be a pupil.

\paragraph{Error and User Feedback Types}
To generate generation errors and implicit user feedback, we use the error and user feedback type taxonomies proposed by \newcite{petrak-etal-2023-learning}. For generation errors in system utterances they define the following nine error types as relevant for task-oriented and document-grounded dialogues:

\begin{itemize}
    \item \textbf{Ignore Question} --- This error occurs when the system fails to address a user's question. Instead of providing a relevant response or clarification, the system disregards their input.
    \item \textbf{Ignore Request} --- A situation in which the system fails to take action on a user's request. It can occur due to various reasons, such as misinterpretation of the request, technical limitations, or system glitches.
    \item \textbf{Ignore Expectation} --- This error happens when the system fails to fulfill the user's expectation in terms of understanding and addressing their needs within the context of the task.
    \item \textbf{Attribute Error} --- If the system fails to correctly extract or understand the necessary slots or attributes from a user's utterance, this is called an attribute error.
    \item \textbf{Factually Incorrect} --- System responses that are factually wrong or inaccurate.
    \item \textbf{Topic Transition Error} --- A situation in which the system's response abruptly shifts to a different or previously discussed topic without a logical connection or adequate context.
    \item \textbf{Conversationality} --- Bad conversationality occurs when the system fails to maintain a coherent and natural conversation flow, e.g., it repeats previous responses or contradicts itself without recognizing or asking for new or missing information.
    \item \textbf{Unclear Intention} --- This error is characterized by the system's failure to accurately address a user's intended objective. 
    \item \textbf{Lack of Sociality} --- If a system's response doesn't adhere to social conventions, fails to include basic greetings, or exhibit toxic and disrespectful behavior or language, this is referred to as a lack of sociality.
\end{itemize}

They also define an error type for common sense errors, but found them rare in task-oriented and  document-grounded dialogues. For this reason, we do not consider this error type in our work. 

For user feedback in response to generation errors, they propose the following taxonomy:

\begin{itemize}
    \item \textbf{Ignore and Continue} --- The user ignores the error and continues the conversation, e.g., "Okay. Let's leave it like that.".
    \item \textbf{Repeat or Rephrase} --- Instead of ignoring the error in the system utterance, the user repeats or rephrases their original concern, e.g., "Actually, I wanted you to ...".
    \item \textbf{Make Aware With Correction} --- The user makes the system aware of its error and provides a correction or response alternative, e.g., "Partly. This doesn't take into account that ...".
    \item \textbf{Make Aware Without Correction} --- Instead of providing a correction or response alternative, the user just makes the system aware of its error, e.g., "You're wrong.".
    \item \textbf{Ask for Clarification} --- In case of error, the user asks the system for clarification, e.g., "I'm not sure what you mean. Is it about ...".
\end{itemize}

\section{Prompts for Dialogue Generation and Annotation}\label{appendix:prompts}
Prompt engineering played a major role in this work. The instructions used to generate the dialogues and annotations were continuously improved in an iterative process to generate valid data within the given parameters. This section only focuses on the final instructions used in this work. Additionally added source data is highlighted in \textcolor[HTML]{0000FF}{blue} in the figures below.

\begin{figure}[htb]
\centering
  \includegraphics[width=0.5\linewidth]{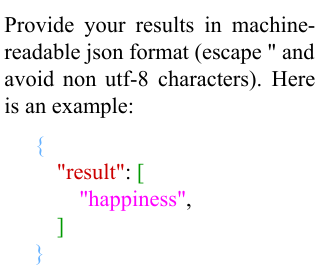}  
  \caption{Instruction to return the results in json for emotion annotation.}
  \label{fig:json_example}
\end{figure}

\paragraph{JSON Schemes} As described in Section~\ref{sec:dialog_generation}, we require GPT-3.5 to return all results in a predefined JSON scheme, which depends on the prompt, i.e., dialogue generation or annotation, and ensures that the returned values contain all required fields and is processable without human intervention. If the values returned do not adhere to the required scheme, we drop the whole dialogue. Figure~\ref{fig:json_example} shows an example for the annotation of emotions.

We append these json schemes at the end of the prompts. We basically provide the required fields and example values, and instruct the model to return only utf-8 encoded characters and escape quotation marks (so that we can treat it as a string in Python). Please refer to our GitHub repository for all prompts and their json schemes\textsuperscript{\ref{github}}.

\begin{figure}[htb]
\centering
  \includegraphics[width=1.0\linewidth]{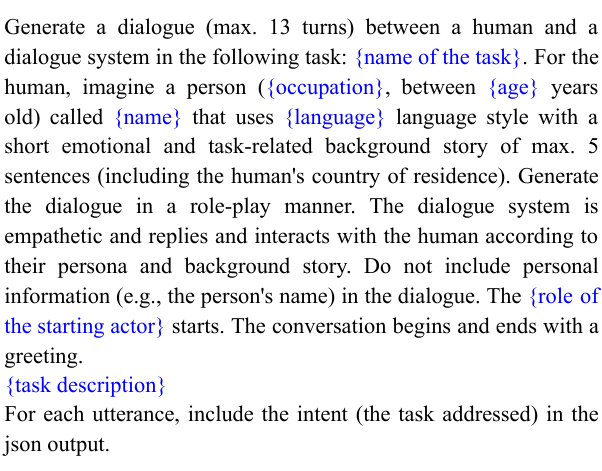}  
  \caption{Instruction for generating feedback-free dialogues.}
  \label{fig:dialog_generation}
\end{figure}

\paragraph{Feedback-Free Dialogues} For dialogue generation, we distinguish feedback-free and feedback dialogues. Figure~\ref{fig:dialog_generation} shows the instruction used to generate feedback-free dialogues.

We provide GPT-3.5 with the demographic information, the role of the starting actor, and the task description. We require the model to use this information to generate a background story and to use this as an additional source for dialogue generation. We also instruct the model to return the utterance-level annotations for intents in this step.

\begin{figure}[htb]
\centering
  \includegraphics[width=1.0\linewidth]{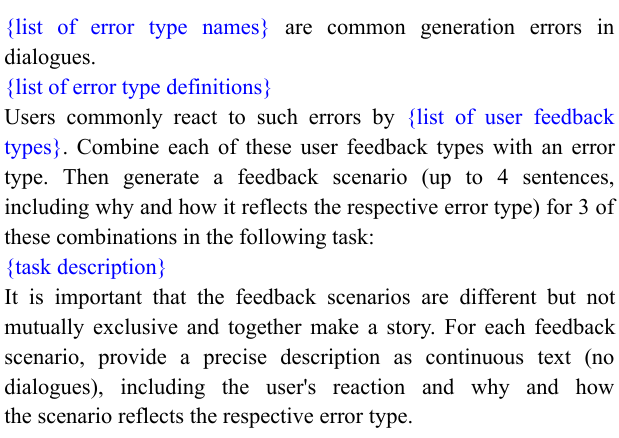}  
  \caption{Instruction for generating the feedback scenarios.}
  \label{fig:error_scenarios}
\end{figure}

\paragraph{Feedback Dialogues} Figure~\ref{fig:error_scenarios} shows the instruction for the generation of feedback scenarios, which are required as an additional source for feedback dialogues. 

\begin{figure}[htb]
  \centering
    \includegraphics[width=1.0\linewidth]{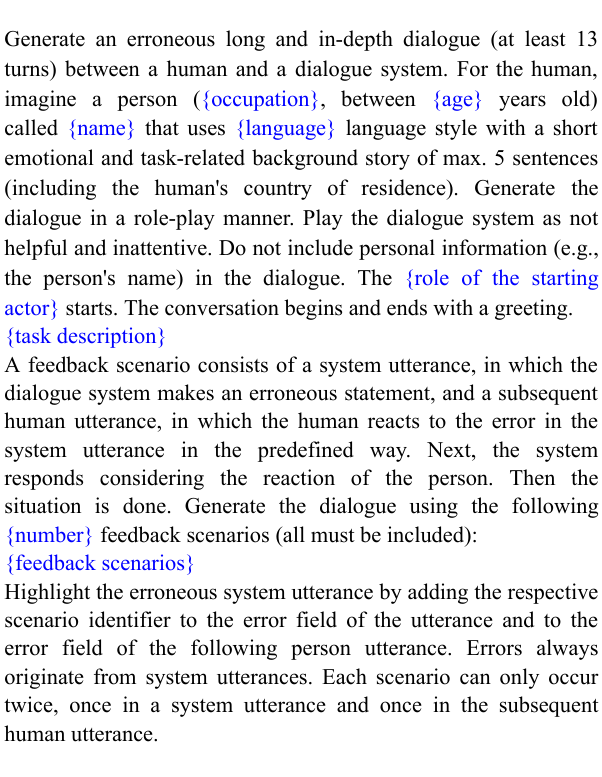}  
    \caption{Instruction for generating feedback dialogues.}
    \label{fig:erroneous_dialogs_prompt}
  \end{figure}

For each feedback dialogue, we generate three feedback scenarios using the same prompt in a separate step before dialogue generation. Figure~\ref{fig:erroneous_dialogs_prompt} shows the instruction for the generation of feedback dialogues.

The instruction is longer and more detailed than the one used for generating the feedback-free dialogues (Figure~\ref{fig:dialog_generation}). For example, it explicitely describes how to process feedback scenarios. Another difference is the length limitation. While feedback-free dialogues are restricted to 13 turns, we require feedback dialogues to have at least 13 turns. In practice, the length of the feedback dialogues is similar to the length of the feedback-free dialogues, but we observed that feedback dialogues are likely to be cut off without this requirement. We consider the generated dialogue as Version 1.

\paragraph{Resolving Feedback Scenarios} For each feedback dialogue (Version 1), we generate three additional versions of the same dialogue, each resolving one of the feedback scenarios. For this, we experimented with different ideas:
\begin{itemize}
    \item Using the implicit user feedback and the task description and instruct GPT-3.5 to rewrite the whole dialogue.
    \item Providing GPT-3.5 with the whole dialogue and only instruct it to rewrite the affected turn.
    \item Using the respective feedback scenario as additional input to regenerate the affected system utterance.
\end{itemize}

\begin{figure}[htb]
  \centering
    \includegraphics[width=0.9\linewidth]{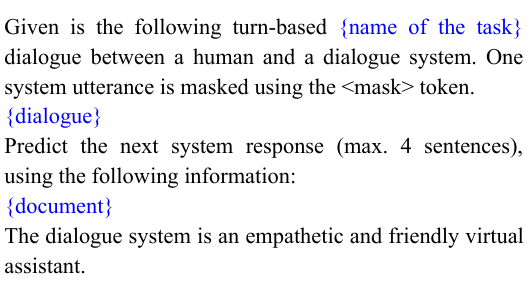}  
    \caption{Instruction for regenerating the system utterance to replace the one with the generation error.}
    \label{fig:fix_prompt}
  \end{figure}

They all resulted in inconsistent dialogues and off-topic or unnatural system utterances. We found that using the dialogue context up to the affected system utterance, masking and regenerating this utterance (in a friendly and polite manner), leads to the best matching and most coherent replacements. Figure~\ref{fig:fix_prompt} shows the instruction. 

It includes the dialogue context, the name of the task and the document if the task is question answering. Although GPT-3.5 has a long context length, we found that including the full task descriptions was distracting rather than improving the replacements. This means that the model can only use internal knowledge and information from the dialogue context for generating the replacements, which sometimes had a negative impact on the completeness of the slot annotations, e.g., for parcel shipping and topping up a prepaid SIM card (see Section~\ref{sec:analysis}). 

\begin{figure*}[ht]
\centering
  \includegraphics[width=1.0\linewidth]{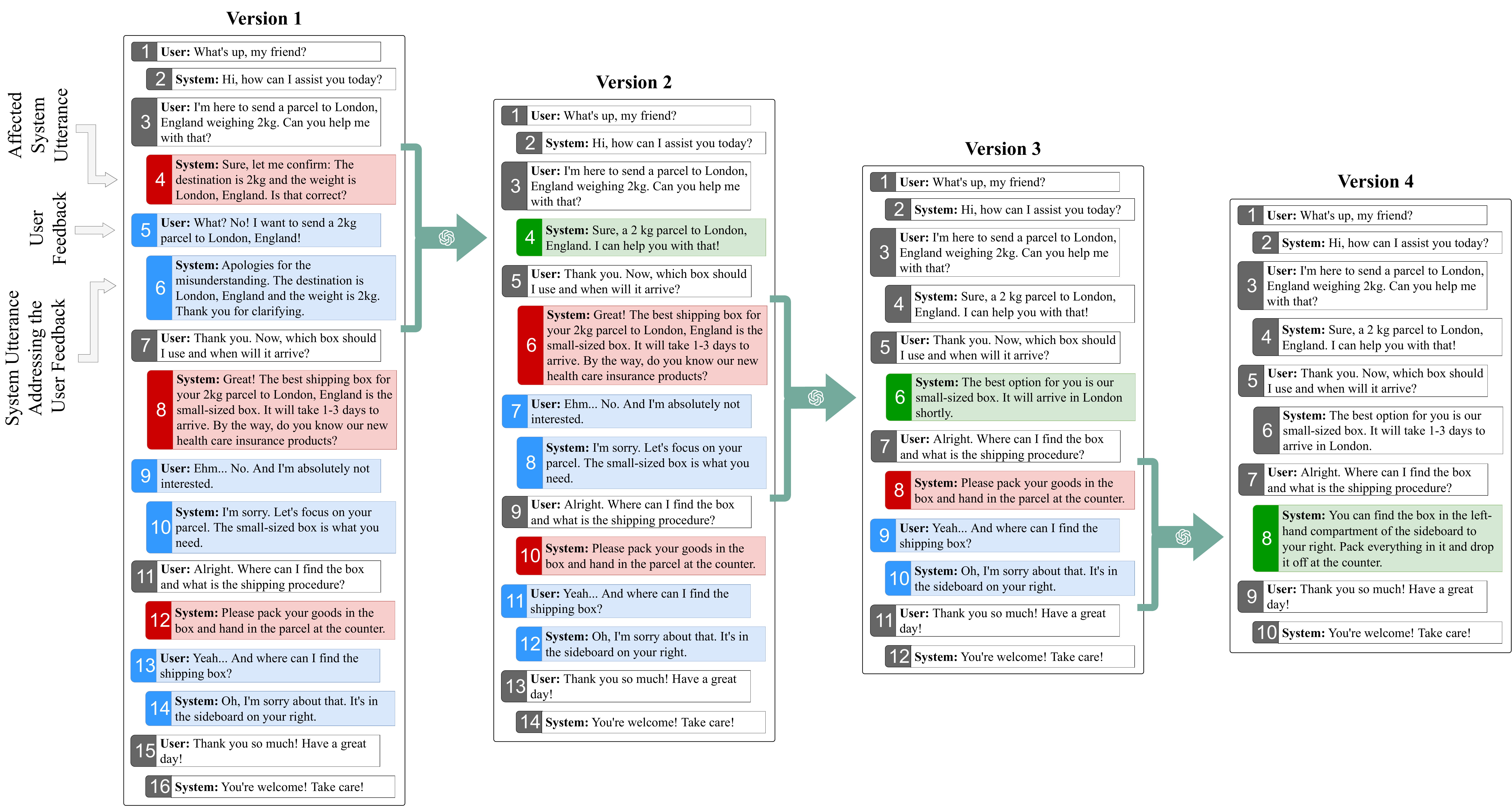}
  \caption{Example dialogue from \ourdata for illustrating our approach for resolving feedback scenarios. In each version, we keep the previous part of the dialogue, regenerate the affected system utterance and drop the following two utterances (the user feedback and the system utterance which addresses the user feedback), since they are directly related to the generation error.} 
  \label{fig:feedback_example}
\end{figure*}

After replacing the affected system utterance, we regenerate its slot values. We remove the following two utterances to ensure the dialogue flow is not corrupted (since they directly refer to the generation error). The conversation then continues with the next regular user utterance. Figure~\ref{fig:feedback_example} shows an example dialogue from \ourdata to illustrate this procedure.

\begin{figure}[htb]
\centering
  \includegraphics[width=1.0\linewidth]{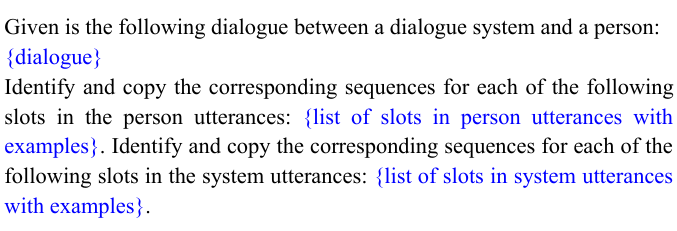}  
  \caption{Instruction for slot annotation in a generated dialogue.}
  \label{fig:slot_annotations}
\end{figure}

\paragraph{Slot Annotations}
Figure~\ref{fig:slot_annotations} shows our instruction for generating the slot annotations. 

For this, we provide GPT-3.5 with the complete dialogue and distinguish between slots for each role (person and system). The slots to be annotated are provided in lists (including example values). We also instruct the model to just use sequences from the dialogue as slot values (to avoid hallucinated slot values).

\begin{figure}[htb]
  \centering
    \includegraphics[width=1.0\linewidth]{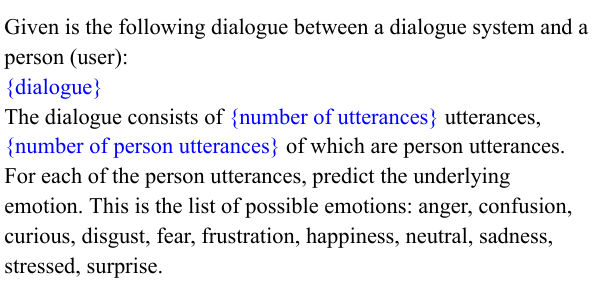}  
    \caption{Instruction for generating emotions.}
    \label{fig:prompt_emotions}
  \end{figure}

\paragraph{Emotion Annotations}
Figure~\ref{fig:prompt_emotions} shows the instruction for emotion generation. 

We generate emotions just based on the dialogue context. We do not provide additional information, such as examples. However, we additionally provide the number of utterances in the dialogue and those related to the user.

\section{Test Data Collection and Curation Study}\label{appendix:human_annotators}
We hired student assistants for our test data collection and curation study. In this section, we want to provide more insights into the application criteria, hiring procedure, and data collection.

\subsection{Application Criteria and Hiring Procedure}\label{appendix:application_criteria}
To participate, we required a formal application. Our criteria were as follows:

\begin{itemize}
    \item Enrollment in computational linguistics, linguistics, data and discourse studies, computer science, business informatics or comparable.
    \item Fluent in reading, speaking and writing English.
    \item Good communication and organization skills.
\end{itemize}

We considered a background in NLP, interest in conversational AI and experience in data annotation as a plus. We did not restrict the job advertisement to our university. Also, we did not consider gender. We asked all applicants who fulfilled those criteria to participate in a recruitment test, in which we asked them to collect and annotate dialogues in a self-chat manner, given a task description from our work. We then assessed and ranked their results based on (1) time needed for one dialogue, (2) annotation completeness, (3) number of turns per dialogue, (4) avg. utterance length. 

Overall, we received 11 applications that fulfilled our criteria. Eight passed the recruitment test and were hired for an hourly salary of 12,95\$. While all participated in the test data collection only two were involved in the data curation study.

\subsection{Test Data Collection}\label{appendix:test_data_collection}
The test data for \ourdata was collected by eight computer science students in overall 136 paid working hours. We randomly assigned participants to groups of two to collect the dialogues in one hour sessions dedicated to one task. For each task, we provided the task description, including slots with examples and four persona profiles (combinations of demographic information) and background stories as inspiration. However, we encouraged them to think about own persona profiles and background stories. For user emotions, we provided them with a list of available options. For question answering, we provided them with the question-document pairs extracted from the POSTE Italiane data (Section~\ref{appendix:task_descriptions}).

\begin{figure}[htb]
  \centering
    \includegraphics[width=0.8\linewidth]{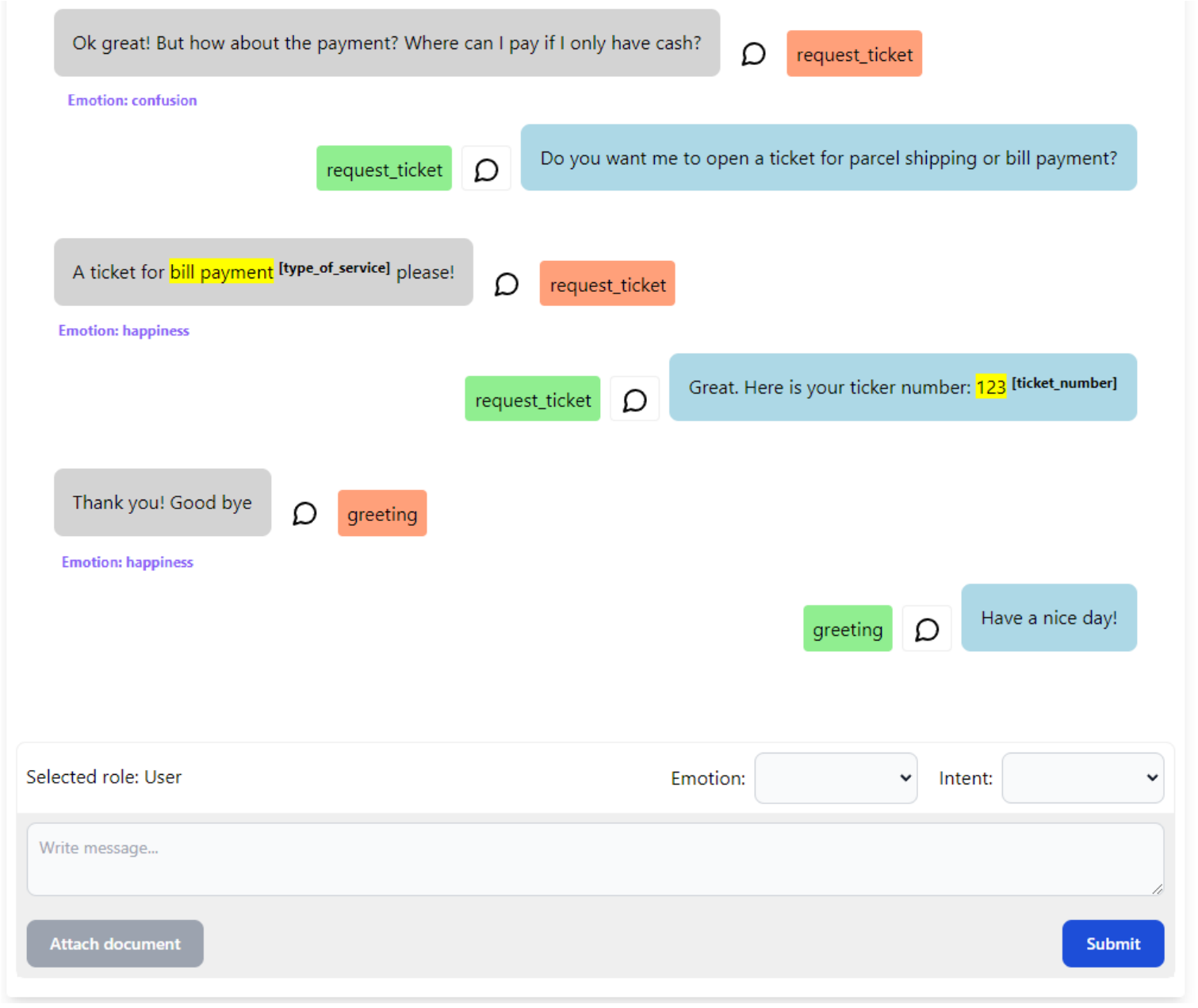}  
    \caption{The user interface of the data collection platform used to collect the test data.}
    \label{fig:chat_pane}
  \end{figure}

For data collection, we used a self-developed web-based platform that allows to collect and annotate dialogues between two humans. Figure~\ref{fig:chat_pane} shows the user interface. 

Each message is annotated with the respective intent (orange or green, depending on the role). Slot annotations are highlighted in yellow, with the slot type as superscript. User emotion annotations are colored purple. For Question Answering, the chatpane also allows attaching a document to a message (a text file).

\subsection{Data Curation Study}\label{appendix:procedure_data_curation}
\begin{figure}[htb]
  \centering
    \includegraphics[width=1.0\linewidth]{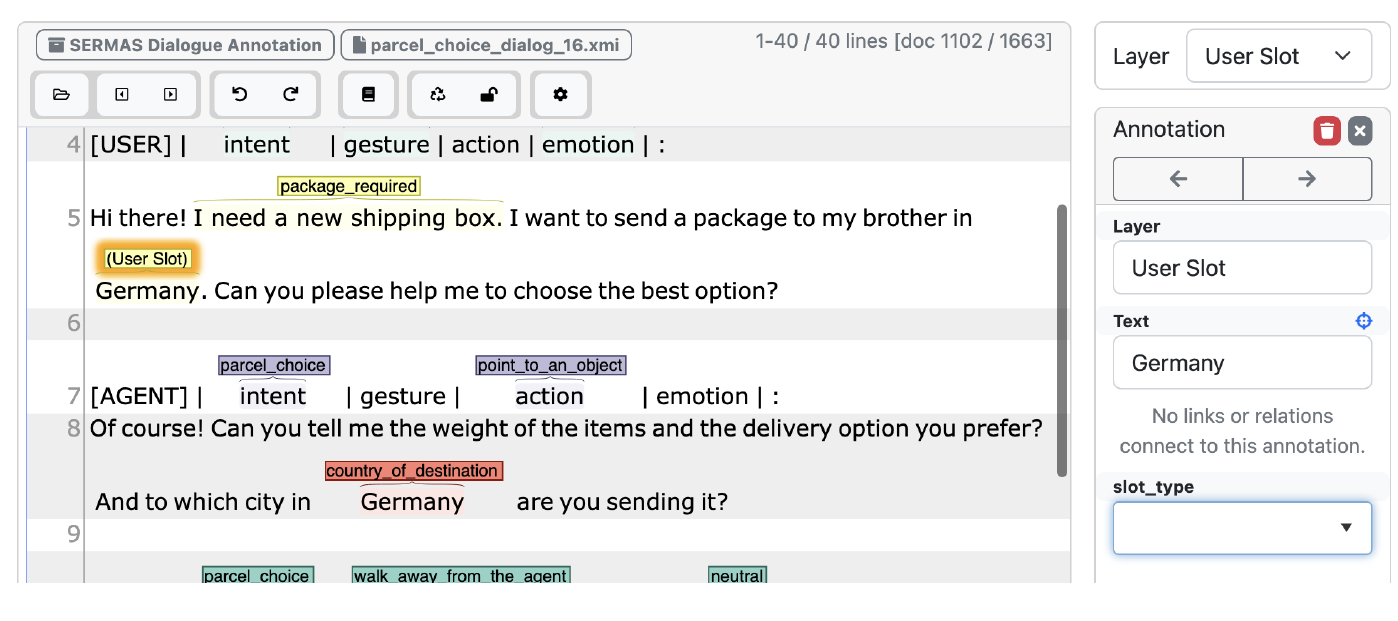}  
    \caption{User interface for intent and slot annotation curation in INCEpTION. It's a parcel shipping dialogue and the annotation for country of destination (\textit{Germany} in line eight) is misplaced, because this slot should be provided by the user, who has already mentioned it in line five.} 
    \label{fig:inception_feedback_free}
  \end{figure}

For the curation of the generated data, the procedure was different for feedback-free and feedback dialogues. For feedback-free dialogues, we asked the annotators to assess and correct (add/modify/delete) the generated slot and intent annotations per utterance, and their completeness on dialogue level (with respect to the task description). We assigned the annotators to the tasks and asked them to work through the corresponding dialogues provided in INCEpTION~\cite{klie-etal-2018-inception}. Figure~\ref{fig:inception_feedback_free} shows the user interface for intent and slot annotation curation.

For feedback dialogues, we asked the annotators to assess and correct the annotations for implemented feedback scenarios, i.e., the annotation for error type in the affected system utterance and the user feedback type in the subsequent user utterance. 
In addition to the information available in the user interface, we provided the annotators with the task descriptions (Appendix~\ref{appendix:task_descriptions}). For feedback dialogues, we also provided them with the definitions of error and user feedback types (Appendix~\ref{appendix:dataset_features}). 

\subsection{Dialogue Collection: Human vs. LLM}\label{appendix:final_thoughts}
In our human test data collection, eight students collected 326 test dialogues in 136 paid working hours. With an hourly salary of 12.95\$, this adds up to a cost of 1,761.20\$ (not including additional costs, such as for supervision). Generating and annotating 8,526 dialogues using GPT-3.5 cost 75.73\$, including API calls for prompt engineering and debugging. On average, collecting and annotating a human-human dialogue cost 5.40\$. Using GPT-3.5, it is 0.009\$. Based on this, collecting and annotating dialogues with human participants is rather uneconomic and inefficient. However, with 175B parameters, GPT-3.5 is an extremely large model. Without access to such a model, this might be different. In a preliminary study, we used Llama-30B~\cite{touvron2023llama} for dialogue generation and annotation. We asked a student assistant from our lab to assess the results. They constantly rated the Llama-30B dialogues lower in terms of naturalness, coherence, engagement, task coverage, i.e., how close is the generated dialogue to the task description, and (turn) length (see Table~\ref{tab:doan}). 

\begin{table}[htb]
  \centering
  \resizebox*{\linewidth}{!}{
\begin{tabular}{lrrrrr}
\multicolumn{1}{c}{\textbf{Model}} & \multicolumn{1}{c}{\textbf{Naturalness}} & \multicolumn{1}{c}{\textbf{Coherence}} & \multicolumn{1}{c}{\textbf{Engagement}} & \multicolumn{1}{c}{\textbf{Task Coverage}} & \multicolumn{1}{c}{\textbf{Length}} \\ \hhline{======}
\multicolumn{1}{l|}{GPT-3.5-Turbo} & \multicolumn{1}{r|}{4.40} & \multicolumn{1}{r|}{4.92} & \multicolumn{1}{r|}{1.0} & \multicolumn{1}{r|}{4.68} & 7,12 \\ \hline
\multicolumn{1}{l|}{LLaMA-30B} & \multicolumn{1}{r|}{3.12} & \multicolumn{1}{r|}{3.52} & \multicolumn{1}{r|}{0.8} & \multicolumn{1}{r|}{3.52} & 3,24
\end{tabular}}  
  \caption{Result of our analysis comparing dialogues generated by GPT-3.5 and Llama-30B. Except for Engagement and Length, all measurments are based on a Likert scale from 1 (lowest rating) to 5 (highest rating).}
  \label{tab:doan}
\end{table}

We suspect that this is rather due to the differences in model size and context window. While GPT-3.5 has a context window of 4k tokens, Llama-30B has a context window of only 2k tokens. However, regardless of the model used, LLM-generated data oftentimes suffers from various kinds of hallucinations~\cite{zhang2023sirens, Ji_2023}, which makes data curation with humans inevitable. In our data curation study (Section~\ref{sub_sec:human_curation_study}), we learned that this is not only much easier for humans, they are also much more efficient in curating annotated dialogues than collecting and annotating them from scratch. For example, collecting and annotating one dialogue takes on average ten minutes and requires two humans. For GPT-3.5 it is only 90 seconds. Curating an annotated dialogue took on average four minutes and did not require a partner.

\section{\ourdata -- Additional Analysis}\label{appendix:additional_analysis}
In this section, we provide additional analysis about the composition of \ourdata. Overall, \ourdata consists of 8,852 dialogues, 1,988 feedback-free and 6,864 feedback dialogues. Table~\ref{tab:dataset_and_split_sizes} shows the distribution of dialogues in the dataset. Test refers to the human-human collected test data. 

\begin{table}[ht!]
  \centering
  \resizebox*{\linewidth}{!}{
\begin{tabular}{lrrrrrrrr}
\multicolumn{1}{c}{\textbf{Task}} & \multicolumn{3}{c}{\textbf{\begin{tabular}[c]{@{}c@{}}Feedback-Free\\ Dialogues\end{tabular}}} & \multicolumn{5}{c}{\textbf{Feedback Dialogues}} \\ \hline
\multicolumn{1}{l|}{} & \multicolumn{1}{c}{\textbf{Train}} & \multicolumn{1}{c}{\textbf{Dev}} & \multicolumn{1}{c|}{\textbf{Test}} & \multicolumn{1}{c}{\textbf{Version 1}} & \multicolumn{1}{c}{\textbf{Version 2}} & \multicolumn{1}{c}{\textbf{Version 3}} & \multicolumn{1}{c|}{\textbf{Version 4}} & \multicolumn{1}{c}{\textbf{Dev}} \\ \hhline{=========}
\multicolumn{1}{l|}{\begin{tabular}[c]{@{}l@{}}Parcel\\ Shipping\end{tabular}} & 186 & 20 & \multicolumn{1}{r|}{38} & 193 & 193 & 193 & \multicolumn{1}{r|}{193} & 84 \\ \hline
\multicolumn{1}{l|}{\begin{tabular}[c]{@{}l@{}}Top Up\\ SIM Card\end{tabular}} & 187 & 20 & \multicolumn{1}{r|}{39} & 193 & 193 & 193 & \multicolumn{1}{r|}{193} & 84 \\ \hline
\multicolumn{1}{l|}{\begin{tabular}[c]{@{}l@{}}Access\\ Control\end{tabular}} & 183 & 20 & \multicolumn{1}{r|}{42} & 215 & 215 & 215 & \multicolumn{1}{r|}{215} & 92 \\ \hline
\multicolumn{1}{l|}{\begin{tabular}[c]{@{}l@{}}Question\\ Answering\end{tabular}} & 943 & 103 & \multicolumn{1}{r|}{207} & 945 & 945 & 945 & \multicolumn{1}{r|}{945} & 420 \\ \hhline{=========}
\multicolumn{1}{l|}{\textbf{Per Split}} & \textbf{1,499} & \textbf{163} & \multicolumn{1}{r|}{\textbf{326}} & \textbf{1,546} & \textbf{1,546} & \textbf{1,546} & \multicolumn{1}{r|}{\textbf{1,546}} & \textbf{680} \\ \hline
\multicolumn{1}{l|}{\textbf{Total}} & \multicolumn{3}{r|}{\textbf{1,988}} & \multicolumn{5}{r}{\textbf{6,864}}
\end{tabular}}  
  \caption{Data splits included in \ourdata and their sizes.}
  \label{tab:dataset_and_split_sizes}
\end{table}

\begin{figure}[htb]
\centering
  \includegraphics[width=1.0\linewidth]{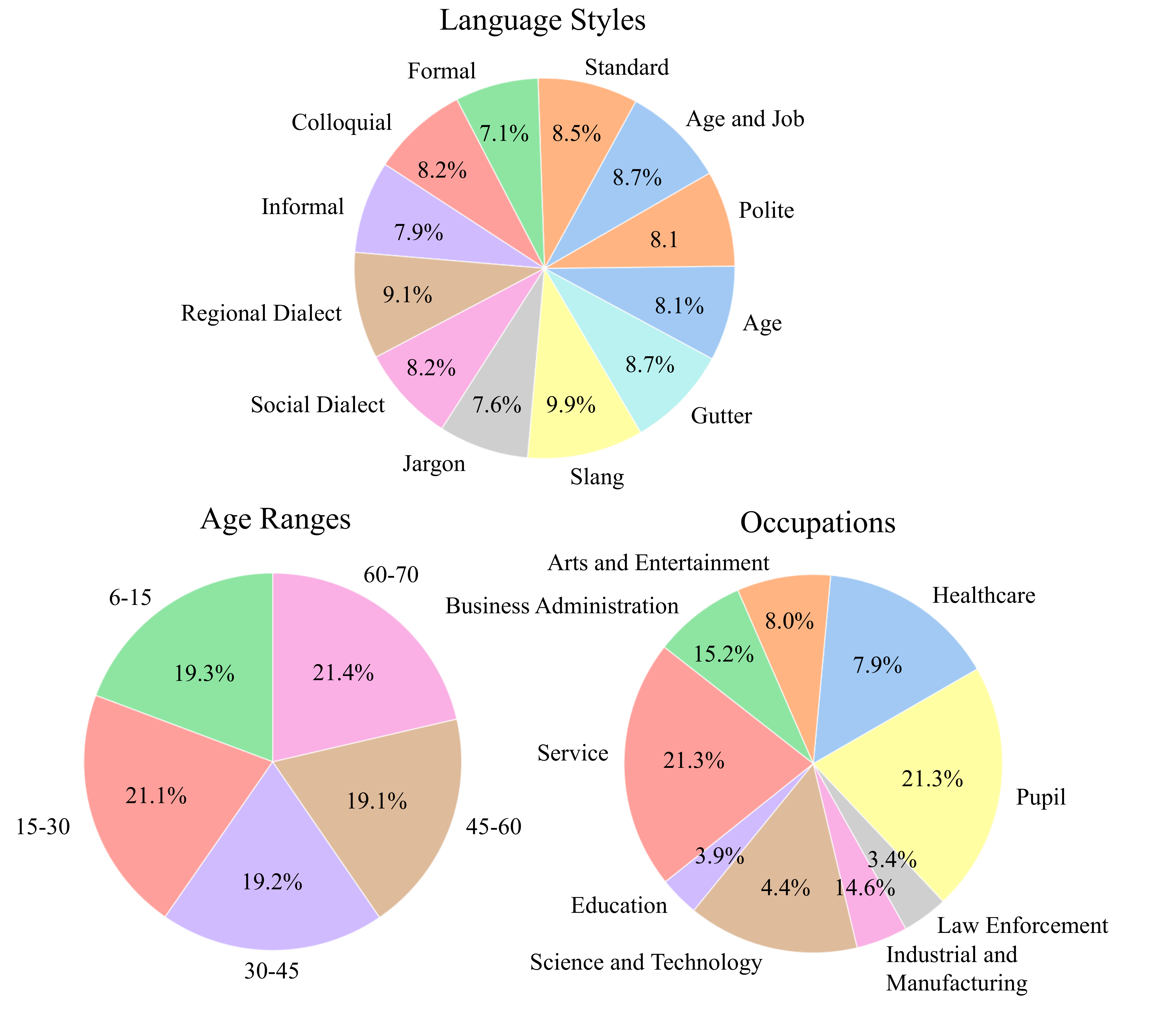}
  \caption{The distribution of persona attributes represented in the background stories (excluding human-human test dialogues).}
  \label{fig:persona}
\end{figure}

\paragraph{Demographic Information} 
Figure~\ref{fig:persona} shows the distribution of language styles, age ranges and occupations randomly sampled for background story generation.

Language styles are almost equally weighted. For occupations, the figure shows that jobs from the categories of business administration, service, industrial and manufacturing, and pupil largely outweigh the other categories, which makes sense in the context of the tasks and topics represented in \ourdata\footnote{The original list did not provide categories. We generated them using GPT-3.5.}. Overall, we observe 693 unique job titles in \ourdata. The figures do not show the distribution of names. We found 1,496 different names in the dialogues. 638 (42\%) are unique, and 712 (47.59\%) occur two to three times. The remaining 146 names occur four or more times in the entire dataset.

\begin{figure}[htb]
\centering
  \includegraphics[width=0.8\linewidth]{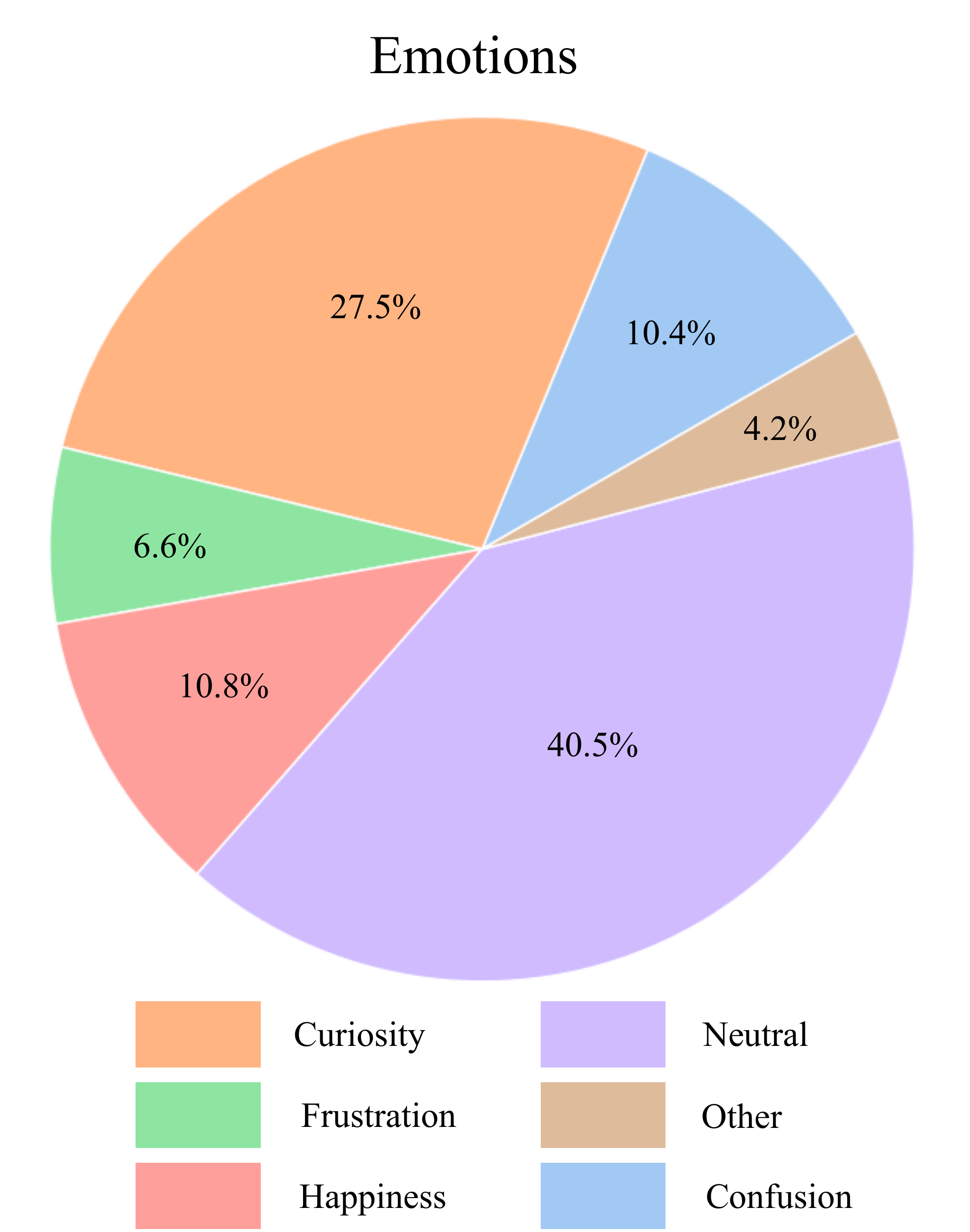}
  \caption{Illustration of the distribution of emotions in \ourdata.}
  \label{fig:emotions}
\end{figure}

\paragraph{Emotions} 
The chart in Figure~\ref{fig:emotions} shows the distribution of emotions in the dialogues of \ourdata.

With 40.5\%, Neutral is the most common emotion, followed by \emph{Curiosity} (27.5\%). \emph{Frustration} and \emph{Confusion} are relatively rare. We observe them mostly in the feedback dialogues. Other refers to emotions that are represented $\leq 5\%$, including \emph{Anger}, \emph{Disgust}, \emph{Fear}, \emph{Surprise}, and \emph{Stress}.

\begin{figure}[htbp]
    \centering
  \includegraphics[width=1.0\linewidth]{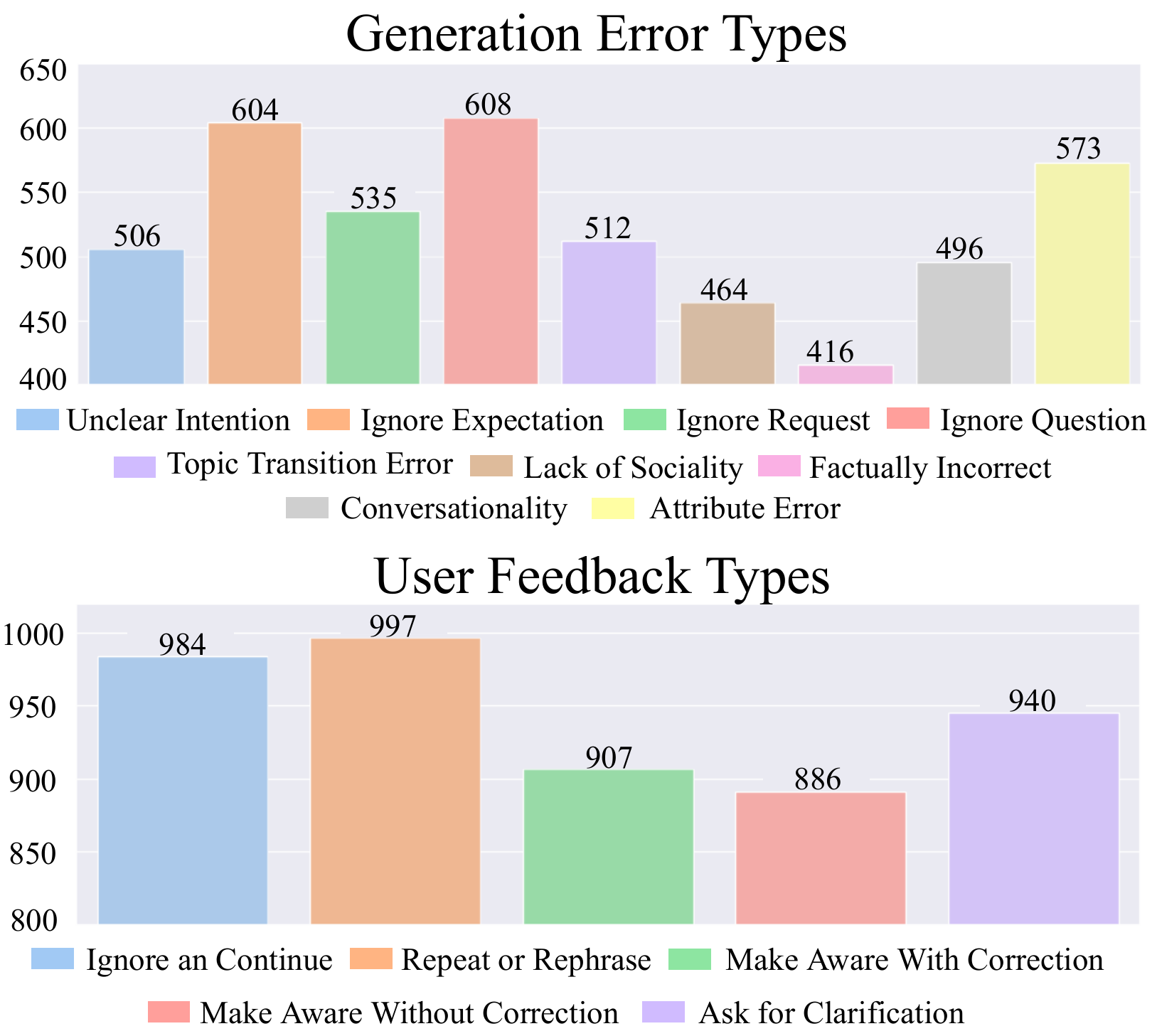}
  \caption{Distribution of generation error and user feedback types in the feedback dialogues of \ourdata.}
  \label{fig:error_user_reaction_types}
\end{figure}

\paragraph{Feedback Scenarios}
Overall, we generated 4,714 feedback scenarios that are included in the feedback dialogues of Version 1. Figure~\ref{fig:error_user_reaction_types} shows the distribution of generation error and user feedback types.

Given that most of the dialogues are about question answering (Table~\ref{tab:dataset_and_split_sizes}), it is not surprising that \emph{Ignore Question} is the most frequent error type. Table~\ref{tab:most_common_error_types} shows the ten most commonly observed error and user feedback type combinations.  

\begin{table}[htb]
  \centering
  \resizebox*{\linewidth}{!}{
\begin{tabular}{rllr}
\multicolumn{1}{l}{} &
  \multicolumn{1}{c}{\textbf{Error Type}} &
  \multicolumn{1}{c}{\textbf{Feedback Type}} &
  \multicolumn{1}{c}{\textbf{Frequency}} \\ \hhline{====}
\multicolumn{1}{r|}{\textbf{1}} & \multicolumn{1}{l|}{Ignore Question}    & \multicolumn{1}{l|}{Ignore and Continue}   & 273 \\ \hline
\multicolumn{1}{r|}{\textbf{2}} & \multicolumn{1}{l|}{Ignore Request}     & \multicolumn{1}{l|}{Ignore and Continue}   & 208 \\ \hline
\multicolumn{1}{r|}{\textbf{3}} & \multicolumn{1}{l|}{Ignore Expectation} & \multicolumn{1}{l|}{Ignore and Continue}   & 199 \\ \hline
\multicolumn{1}{r|}{\textbf{4}} & \multicolumn{1}{l|}{Unclear Intention}  & \multicolumn{1}{l|}{Ask for Clarification} & 191 \\ \hline
\multicolumn{1}{r|}{\textbf{5}} & \multicolumn{1}{l|}{Ignore Question}    & \multicolumn{1}{l|}{Repeat or Rephrase}    & 187 \\ \hline
\multicolumn{1}{r|}{\textbf{6}} &
  \multicolumn{1}{l|}{Factually Incorrect} &
  \multicolumn{1}{l|}{\begin{tabular}[c]{@{}l@{}}Make Aware With\\ Correction\end{tabular}} &
  166 \\ \hline
\multicolumn{1}{r|}{\textbf{7}} &
  \multicolumn{1}{l|}{\begin{tabular}[c]{@{}l@{}}Topic Transition\\ Error\end{tabular}} &
  \multicolumn{1}{l|}{Ask for Clarification} &
  158 \\ \hline
\multicolumn{1}{r|}{\textbf{8}} & \multicolumn{1}{l|}{Attribute Error}    & \multicolumn{1}{l|}{Repeat or Rephrase}    & 156 \\ \hline
\multicolumn{1}{r|}{\textbf{9}} & \multicolumn{1}{l|}{Ignore Expectation} & \multicolumn{1}{l|}{Repeat or Rephrase}    & 151 \\ \hline
\multicolumn{1}{r|}{\textbf{10}} &
  \multicolumn{1}{l|}{Lack of Sociality} &
  \multicolumn{1}{l|}{\begin{tabular}[c]{@{}l@{}}Make Aware Without\\ Correction\end{tabular}} &
  141
\end{tabular}}
  \caption{The table shows the most common error and user feedback type combinations included in \ourdata.}
  \label{tab:most_common_error_types}
\end{table}

\emph{Ignore Question} and \emph{Ignore Request} are two of the most frequent error types. While we observe the first one more common in question answering dialogues, the second one is more common in the other tasks. For both we observe that \emph{Ignore and Continue} is the most frequent user feedback type, followed by \emph{Repeat or Rephrase}. \emph{Unclear Intention} is an error type mostly observed in parcel shipping, topping up a prepaid SIM card, and access control. The most frequently observed user feedback to this is \emph{Ask for Clarification}. Based on absolute numbers, \emph{Factually Incorrect} is the rarest error type. It is mostly observed in question answering and in combination with \emph{Make Aware With Correction}.

\subsection{Curation Study -- Results Analysis} \label{appendix:curation_further_analysis}

To further investigate the quality of the generated annotations, we provide the human curation results in this section. Table~\ref{tab:curation_detailed_analysis} investigates the intent and slot annotations of the feedback-free dialogues before and after curation.

\begin{table}[htb]
  \centering
  \resizebox*{\linewidth}{!}{
\begin{tabular}{lrrrr}
\multicolumn{1}{c}{\textbf{Task}} & \multicolumn{2}{c}{\textbf{Intents}} & \multicolumn{2}{c}{\textbf{Slots}} \\ \hhline{=====}
\multicolumn{1}{c|}{} & \multicolumn{1}{c}{\textbf{Non-Curated}} & \multicolumn{1}{c|}{\textbf{Curated}} & \multicolumn{1}{c}{\textbf{Non-Curated}} & \multicolumn{1}{c}{\textbf{Curated}} \\ \hline
\multicolumn{1}{l|}{\begin{tabular}[c]{@{}l@{}}Parcel\\ Shipping\end{tabular}} & 0.63 & \multicolumn{1}{r|}{1.0} & 0.48 & 0.62 \\ \hline
\multicolumn{1}{l|}{\begin{tabular}[c]{@{}l@{}}Top Up\\ SIM Card\end{tabular}} & 0.71 & \multicolumn{1}{r|}{1.0} & 0.61 & 0.91 \\ \hline
\multicolumn{1}{l|}{\begin{tabular}[c]{@{}l@{}}Access \\ Control\end{tabular}} & 0.95 & \multicolumn{1}{r|}{1.0} & 0.38 & 0.60 \\ \hline
\multicolumn{1}{l|}{\begin{tabular}[c]{@{}l@{}}Question\\ Answering\end{tabular}} & 0.96 & \multicolumn{1}{r|}{1.0} & 0.60 & 1.0
\end{tabular}}
  \caption{The table compares the completeness with respect to intent and slot values before and after human curation.}
  \label{tab:curation_detailed_analysis}
\end{table}

As said in Section~\ref{sub_sec:human_curation_study}, the slot and intent annotations in the parcel shipping dialogues were most affected by human curation. The intent annotations for these dialogues are now complete. The completeness of the slot values was increased by 0.14 to 0.62. For access control, the situation is similar. The ratio of slot annotations in the \emph{Top Up SIM Card} dialogues was increased by 0.29, and the question answering dialogues are now fully annotated. 


\begin{figure}[htb]
\centering
  \includegraphics[width=\linewidth]{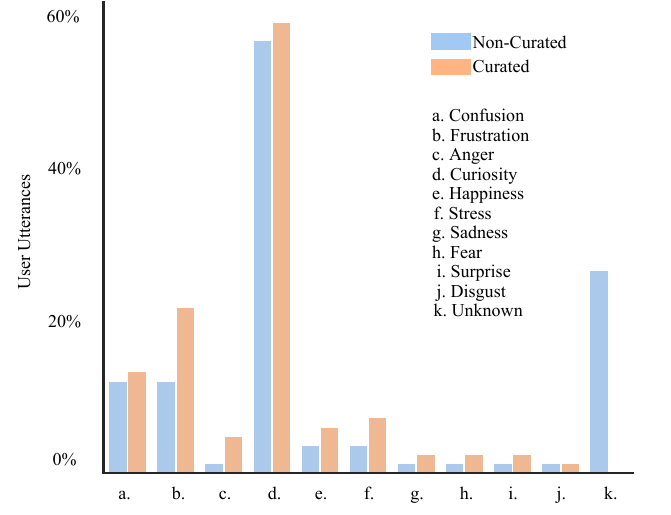}
  \caption{Illustration of the distribution of emotions before and after human curation.}
  \label{fig:emotions_curated}
\end{figure}

Figure~\ref{fig:emotions_curated} shows the changes in emotion annotations in the curated feedback-free dialogues. The ratio of each emotion has increased slightly, primarily due to the correction of unknown emotion annotations by the human curators, i.e. emotions that were not included in our taxonomy. The figure does not include the \emph{Neutral} emotion which is still the most dominant emotion in the dataset (approx. 46\% in the curated data).

\begin{figure}[htb]
\centering
  \includegraphics[width=\linewidth]{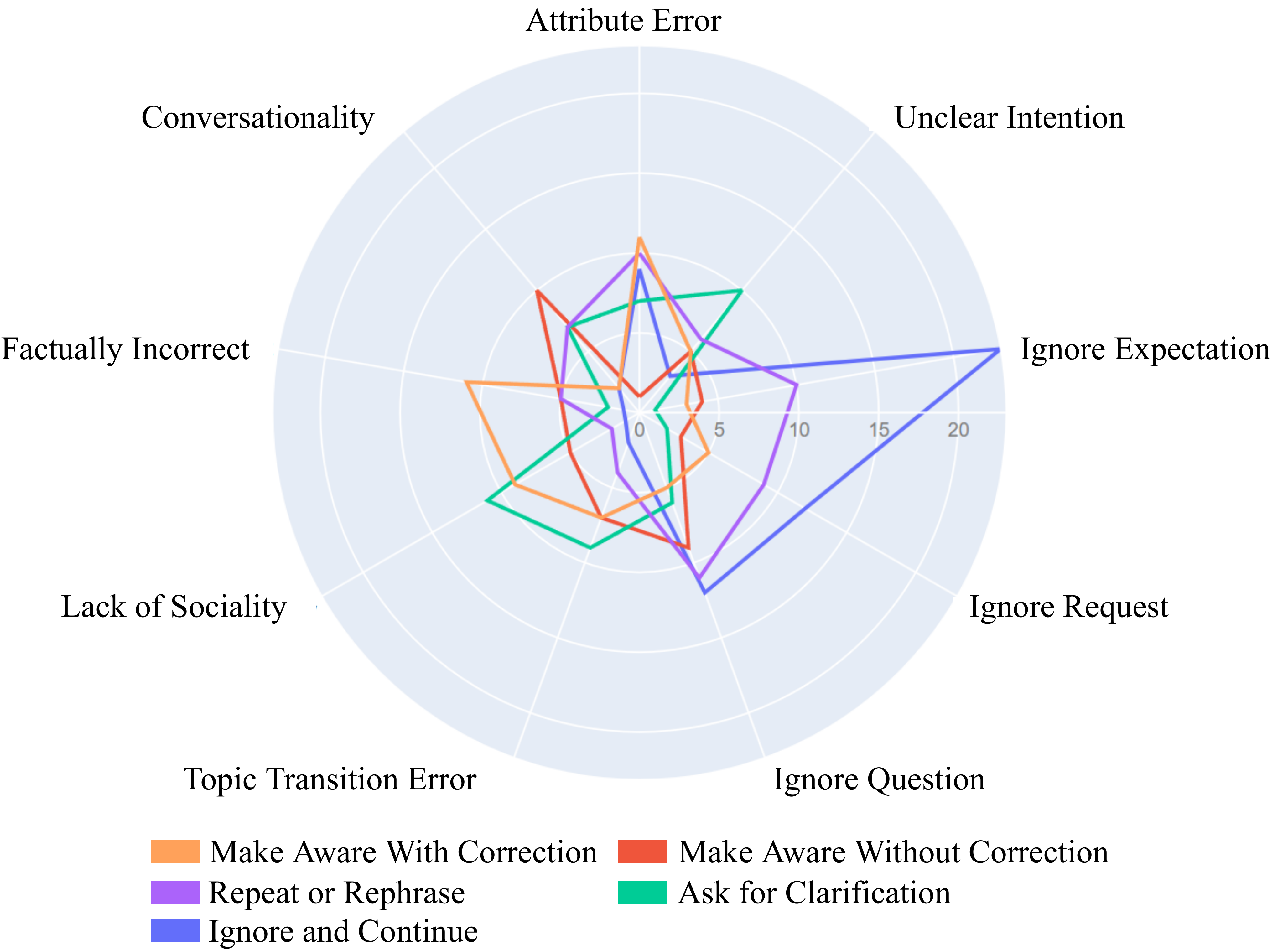}
  \caption{Illustration of the distribution of user feedback types to generation error types before human curation.}
  \label{fig:errors_non_curated}
\end{figure}

\begin{figure}[htb]
\centering
  \includegraphics[width=\linewidth]{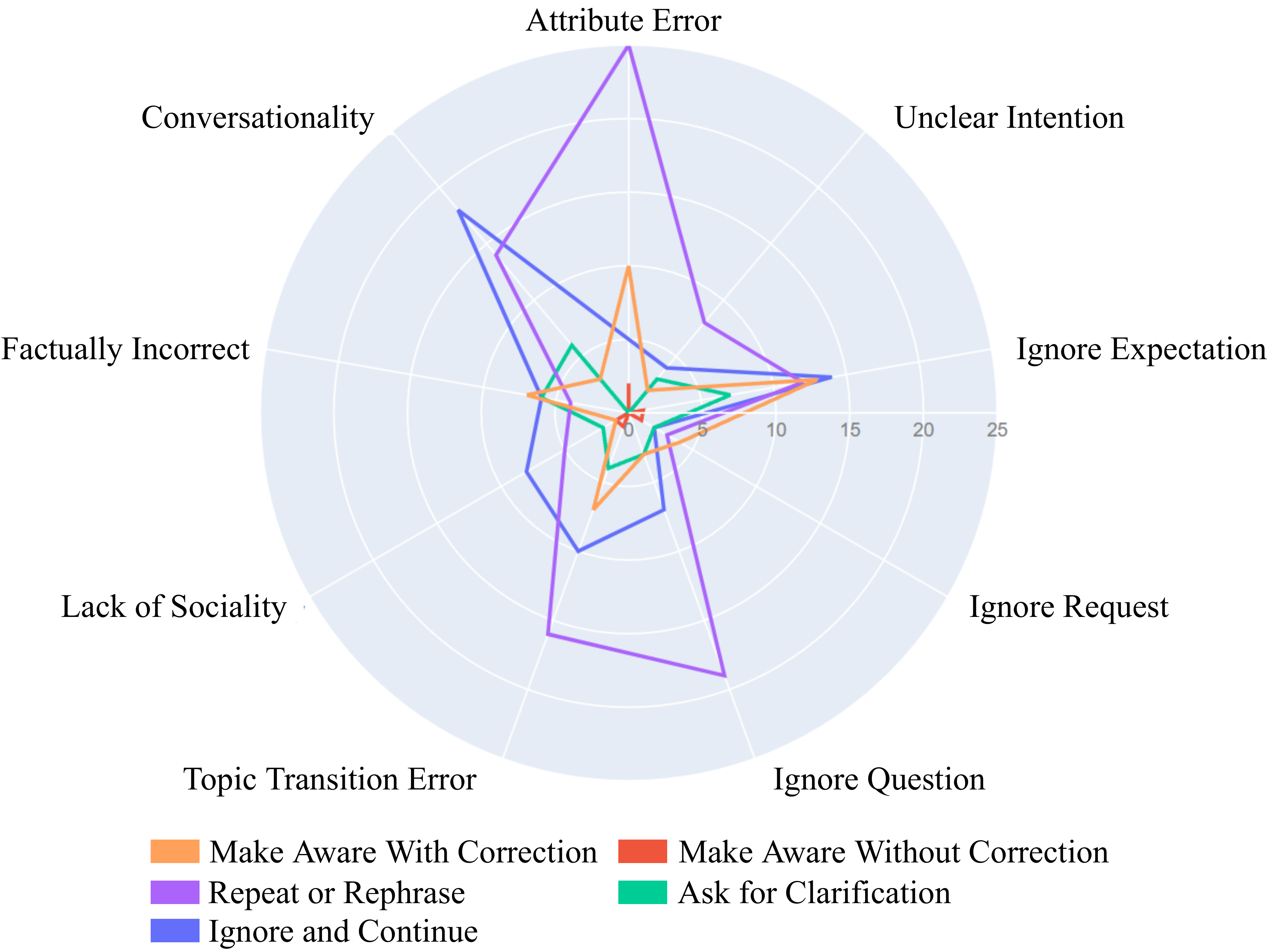}
  \caption{Illustration of the distribution of user feedback types to generation error types after human curation.}
  \label{fig:errors_curated}
\end{figure}

Figure~\ref{fig:errors_non_curated} shows the distribution of user feedback to generation error types before human curation. It shows a strong tendency towards \emph{Ignore and Continue} in the case of \emph{Ignore Expectation}, \emph{Ignore Request} and \emph{Ignore Question} errors. Figure~\ref{fig:errors_curated} shows that many of these annotations were changed by the human curators to \emph{Make Aware With Correction} and \emph{Repeat or Rephrase}. The significant changes concerning \emph{Attribute Error}, \emph{Ignore Question}, \emph{Conversationality} and \emph{Topic Transition Error} show that the error type annotations were corrected particularly frequently in these cases. 

The total number of error annotations itself has only changed slightly. We assume that the examples used for annotation were not distinct enough to support GPT-3.5 in interpreting the definitions for user feedback and generation errors (see Figure~\ref{fig:overview}; for feedback scenario generation, we provided the model with definitions for error types and user feedback and examples). Especially in the case of e.g. \emph{Ignore Question} and \emph{Ignore Request} (see Appendix~\ref{appendix:dataset_features} for the definitions), it depends very much on specific details (e.g., punctuation marks).


\section{Experimental Details and Additional Results}\label{appendix:expeirments}
In this section, we provide additional information on our experiments, including hyperparameters, input sequences, and additional results.

\subsection{Training Configuration}\label{appendix:hyperparameters}
\paragraph{Hyperparameter}
For the experiments with feedback-free dialogues, we trained all models for five epochs, except for Llama 2~\cite{touvron2023llama2}, which was trained for ten epochs, since it took already five epochs to adapt the pretrained model to our prompting mechanism (we used the plain pretrained model in our experiments, not the one finetuned on dialogue data). For the experiments with feedback dialogues, we subsequently trained the best performing feedback-free models for ten epochs using the feedback data (ten epochs since we have seen further improvements after the fifth epoch). 

For all experiments, we used a batch size of 32 and a learning rate of $5e-5$ with no warmup steps. As optimizer, we used the implementation of AdamW~\cite{loshchilov2019decoupled} in Pytorch\footnote{\href{https://pytorch.org/docs/stable/generated/torch.optim.AdamW.html}{AdamW} in the Pytorch documentation (last accessed 30 January 2024).}. Except for Llama 2, we fully-finetuned all models. For Llama 2, we only finetuned the LoRA~\cite{hu2022lora} weights, using a rank of 8, an alpha of 16, and a dropout rate of 0.05.

\paragraph{Data Configuration for Feedback Training}
The feedback experiments investigate the impact of learning from generation errors and feedback in the input sequence on intent prediction, slot extraction, and response generation. We only use the dialogues from Version 2 to Version 4 (see Table~\ref{tab:dataset_and_split_sizes}) for these experiments and include them as additional information in the input sequences (see Appendix~\ref{appendix:input_sequences}). We note that dialogues from Version 4 are corrected, i.e., dialogues without generation errors or feedback (see Figure~\ref{fig:feedback_example}). We include them to avoid training too much on generation errors and feedback~\cite{xu-etal-2023-learning, ung-etal-2022-saferdialogues}. We did not use the dialogues from Version 1, as these only include generation errors.

\subsection{Input Sequences}\label{appendix:input_sequences}
\begin{figure}[htb]
  \centering
    \includegraphics[width=1.0\linewidth]{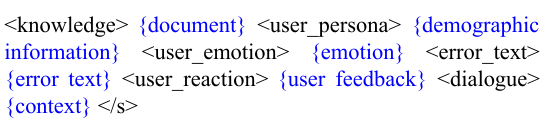}  
    \caption{Input sequence for Flan-T5.}
    \label{fig:input_seq_flan_t5}
  \end{figure}

Each model used in this work requires a different input sequence. In general, the components of the input sequence depend on the features used (e.g., user emotions or demographic information). Figure~\ref{fig:input_seq_flan_t5} shows the input sequence used for training and inference using Flan-T5~\cite{flan_t5}. Additionally added source data is highlighted in \textcolor[HTML]{0000FF}{blue} in the figures below.

\begin{figure}[htb]
  \centering
    \includegraphics[width=1.0\linewidth]{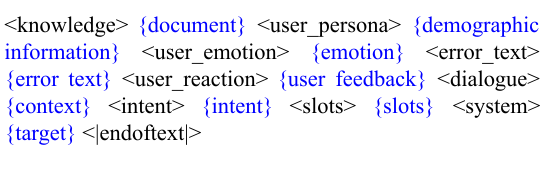}  
    \caption{Input sequence for GPT-2.}
    \label{fig:input_seq_gpt_2}
  \end{figure}

The target sequence includes the intent, slot values, and system response. It is basically the same as the last part of the input sequence for GPT-2~\cite{gpt2}, which is shown in Figure~\ref{fig:input_seq_gpt_2} (starting from <intent>).

\begin{figure}[htb]
  \centering
    \includegraphics[width=1.0\linewidth]{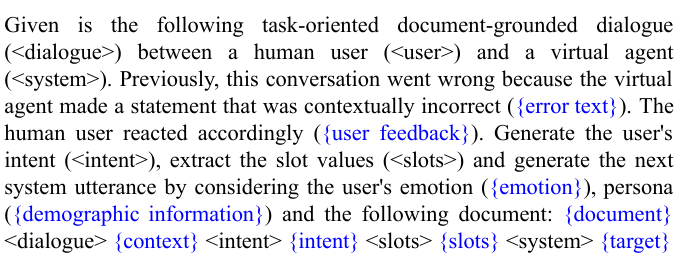}  
    \caption{Input sequence for Llama 2.}
    \label{fig:input_seq_llama_2}
  \end{figure}

For inference with GPT-2, we used the same sequence as for Flan-T5. For Llama 2~\cite{touvron2023llama2}, Figure~\ref{fig:input_seq_llama_2} shows the sequence.

\begin{figure}[htb]
  \centering
    \includegraphics[width=1.0\linewidth]{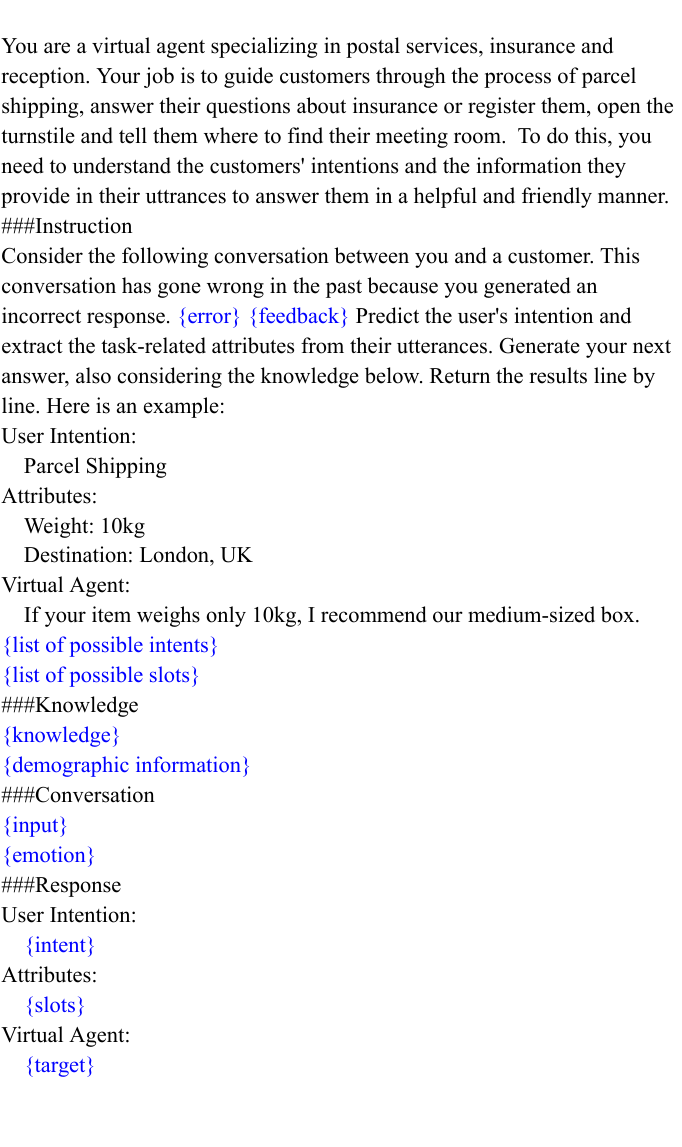}  
    \caption{Input sequence for Llama 3.}
    \label{fig:input_seq_llama_3}
  \end{figure}

For inference, we only use the sequence up to the dialogue context (similar to GPT-2). Figure~\ref{fig:input_seq_llama_3} shows the instruction used for Llama 3~\cite{llama3}.

It is much more detailed than the Llama 2 input sequence and provides behavioral instructions (at the beginning of the sequence), and a list of possible intent and slot values (but without further description or examples). In fact, it is the sequence we originally designed for Llama 2. In Llama 2, it did not lead to reliable results, and the behavioral instructions and the list of possible intents and slots appeared to be rather distracting and a source of potential hallucinations, which is why we removed them.

For inference, we only use the sequence up to the response tag.


\subsection{Feedback Data as Negative Samples}\label{appendix:negative_example}
 We attribute the performance improvements in the feedback experiments to the additional context provided by the generation error and user feedback. We assume they serve as a negative example during training and help the models to learn to generate more accurate intents, slots and responses that better reflect the knowledge documents. This section provides an example from our experiments to support this intuition. 

 The dialogue is a question answering dialogue from the financial domain, and the subject of the conversation is using the Postepay Evolution Card. This is the respective knowledge document: \textit{The Postepay Evolution Card allows the Cardholder, within the limits of the amount available, to make cash withdrawals and payment transactions (with the exception of purchases by mail or telephone and all transactions that are not authorized online, i.e., at the same time as the payment) in Italy and abroad. In addition, the card allows for the payment of highway tolls on the enabled sections.} In the dialogue, the user wants to know if they can use the card for online purchases. The target intent is \textit{question answering} (the respective task) and the target slot is the user's question (\textit{I also wanted to inquire about the usability of the Postepay Evolution Card for online purchases}). The feedback scenario is the following: The system provided the user with a factually incorrect response (\textit{Yes, you can also use the card for purchases by mail or telephone.}), and the user responded with a correction (\textit{Sorry, but that's not true. The Postepay Evolution Card cannot be used for purchases by mail or telephone.}). We generated the next system utterance using the configuration and model from the feedback-free and feedback experiments highlighted in Table~\ref{tab:main_experiments_automatic_evaluation}. 

\begin{figure}[htb]
 \centering
   \includegraphics[width=1.0\linewidth]{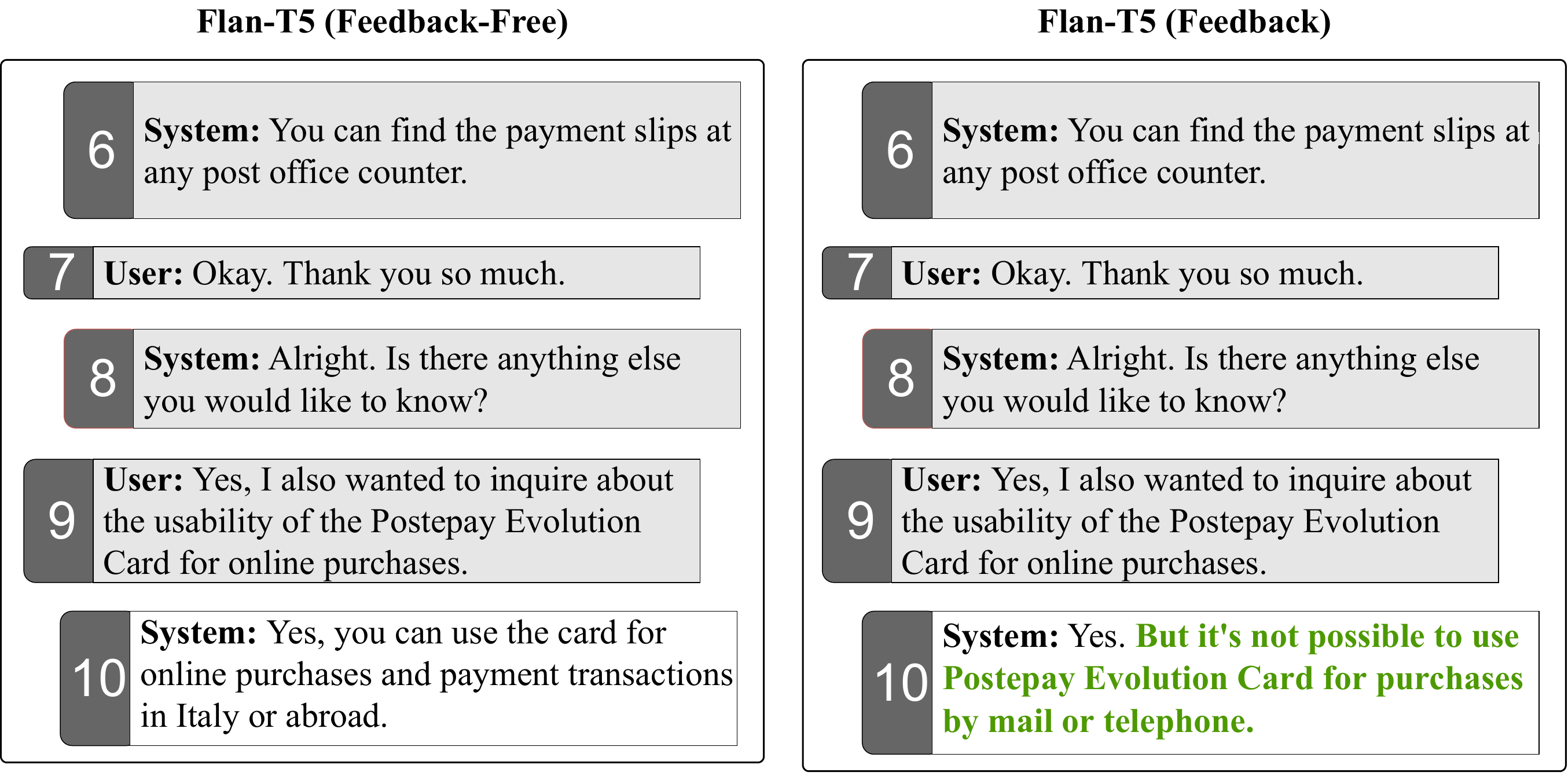}  
   \caption{Example for Flan-T5. The feedback-free model was trained using user emotions as an additional input signal. The feedback model was then additionally trained using generation errors and user feedback.}
   \label{fig:use_case_flan_t5}
 \end{figure}

 Figure~\ref{fig:use_case_flan_t5} shows the results for Flan-T5~\cite{flan_t5}. The responses are focused on the question and do not provide any additional details, but the feedback model focuses more on the information from the knowledge document. The feedback-free model predicted Bill Form Payment Procedure as the slot, which is incorrect, and set the complete user utterance as the value. Question would have been the correct slot type and the target value does not include the complete user utterance, but only the part after the comma. The feedback model predicted both correctly.

\begin{figure}[htb]
 \centering
   \includegraphics[width=1.0\linewidth]{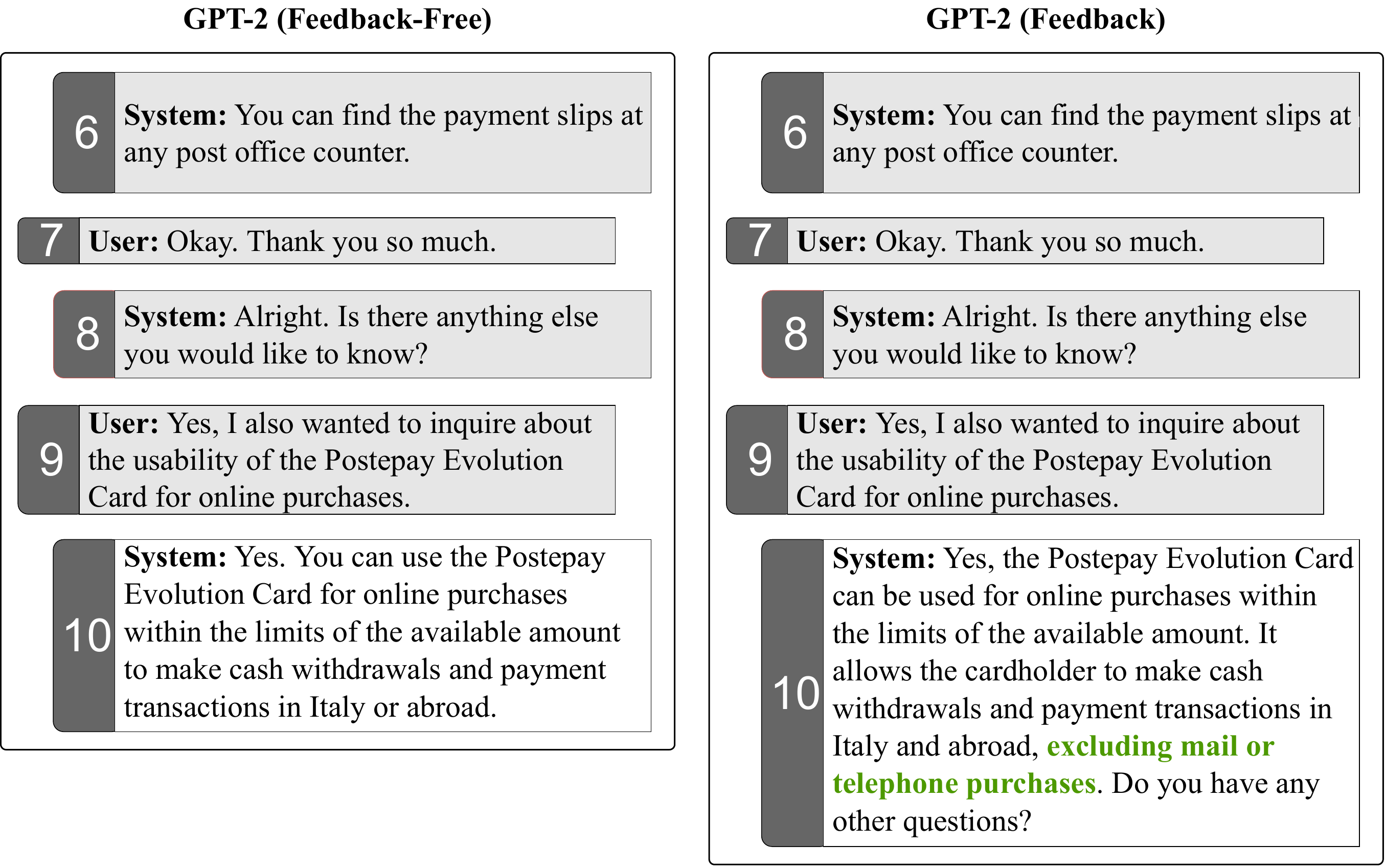}  
   \caption{Example for GPT-2. The feedback-free model was trained using user emotions and demographic information as additional input signals. The feedback model was then additionally trained using generation errors and user feedback (like Flan-T5).}
   \label{fig:use_case_gpt_2}
 \end{figure}

Figure~\ref{fig:use_case_gpt_2} shows the results for GPT-2~\cite{gpt2}. The responses provide more details and read more naturally. However, the response from the feedback model is closer to the knowledge document and ends with a call for interaction. Both the feedback-free and feedback models correctly predicted the intent, slot values, and types.

\begin{figure}[htb]
 \centering
   \includegraphics[width=1.0\linewidth]{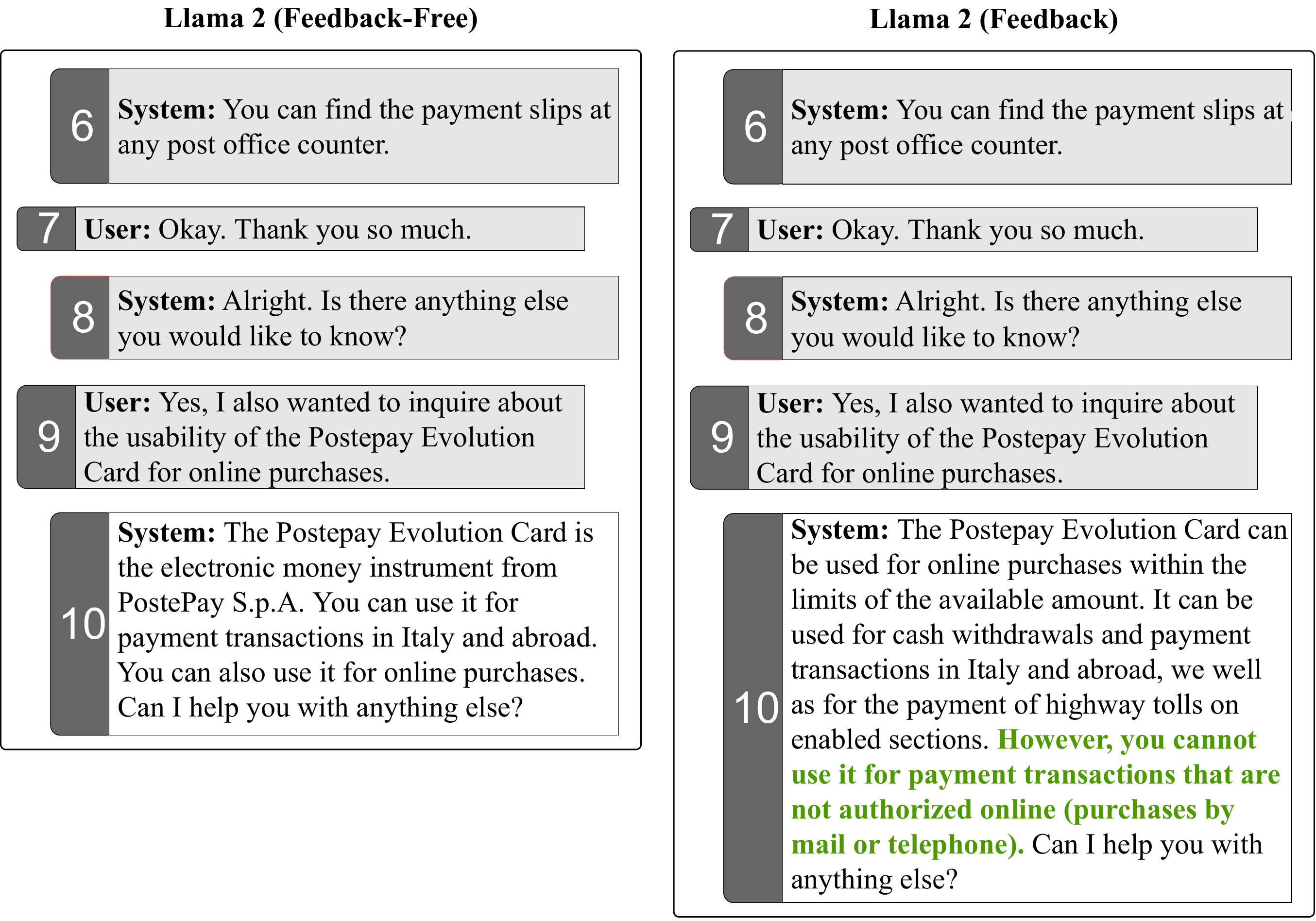}  
   \caption{Example for Llama 2. The feedback-free model was trained using user emotions as additional input signals. The feedback model was then additionally trained using just user feedback.}
   \label{fig:use_case_llama_2}
 \end{figure}

 Figure~\ref{fig:use_case_llama_2} shows the responses from Llama 2~\cite{touvron2023llama2}. In terms of content, they hardly differ from the system utterances generated by GPT-2, they are just not as concise. However, in contrast to Flan-T5 and GPT-2, Llama 2 mispredicted the intent and slot values in both cases. The feedback model predicted information retrieval as value for intent. For the slot value, it did not extract the user's question from their utterance (although this instruction is included in the input sequence, see Appendix~\ref{appendix:input_sequences}). Instead, the model returned a reformulation: \textit{Can I use the Postepay Evolution Card to make purchases by mail or telephone?}

\subsection{Generation Accuracy Performance Gap}\label{appendix:performance_gap}
Table~\ref{tab:tod_qa_comparison_flan_t5}, Table~\ref{tab:tod_qa_comparison_gpt_2} and Table~\ref{tab:tod_qa_comparison_llama_2} show the results from Table~\ref{tab:main_experiments_automatic_evaluation} divided into question answering (QA) and the other tasks (Others), including parcel shipping, topping up a prepaid SIM card, and access control. For feedback, we only consider the best configuration for each model. As in Table~\ref{tab:main_experiments_automatic_evaluation}, we use the respective base models as deltas and highlight the best performing configurations. Since the \ourdata test split contains only 119 ToD dialogs and 207 QA dialogs (see Table~\ref{tab:dataset_and_split_sizes}), we randomly selected 119 samples from question answering to ensure comparability.

\begin{table}[ht]
  \centering
  \resizebox*{\linewidth}{!}{
\begin{tabular}{llrrr}
 & \multicolumn{1}{l}{\textbf{Experiment}} & \multicolumn{1}{l}{\textbf{F1}} & \multicolumn{1}{l}{\textbf{BLEU}} & \multicolumn{1}{l}{\textbf{BertScore}} \\ \hhline{=====}
 
 \multicolumn{1}{l|}{\multirow{5}{*}{QA}} & \multicolumn{1}{l|}{Flan-T5} & 47.6 & 25.9 & 88.1 \\ \hhline{~|====}
 
 \multicolumn{1}{l|}{} & \multicolumn{1}{l|}{\textbf{+ Emotions}} & \textbf{\textcolor{positive_color}{53.5}} & \textbf{\textcolor{positive_color}{31.3}} & \textbf{\textcolor{positive_color}{89.7}} \\ \cline{2-5} 
 
 \multicolumn{1}{l|}{} & \multicolumn{1}{l|}{+ Demographics} & \textcolor{positive_color}{52.2} & \textcolor{positive_color}{30.2} & 88.9 \\ \cline{2-5} 
 
 \multicolumn{1}{l|}{} & \multicolumn{1}{l|}{\begin{tabular}[c]{@{}l@{}}+ Emotions\\ + Demographics\end{tabular}} & \textcolor{positive_color}{50.0} & \textcolor{positive_color}{28.2} & 88.7 \\ \cline{2-5} 
 
 \multicolumn{1}{l|}{} & \multicolumn{1}{l|}{\begin{tabular}[c]{@{}l@{}}+ Emotions\\ + Generation Error\\ + User Feedback\end{tabular}} & \textcolor{positive_color}{48.8} & \textcolor{positive_color}{32.0} & \textcolor{positive_color}{89.2} \\ \cline{1-5} 
 
 \multicolumn{1}{l|}{\multirow{5}{*}{Others}} & \multicolumn{1}{l|}{Flan-T5} & 33.6 & 5.2 & 87.2 \\ \hhline{~|====}
 
 \multicolumn{1}{l|}{} & \multicolumn{1}{l|}{\textbf{+ Emotions}} & \textbf{\textcolor{positive_color}{36.7}} & \textbf{\textcolor{positive_color}{6.8}} & \textbf{\textcolor{positive_color}{88.4}} \\ \cline{2-5} 

 \multicolumn{1}{l|}{} & \multicolumn{1}{l|}{+ Demographics} & 32.9 & 5.6 & 86.5 \\ \cline{2-5} 
 
 \multicolumn{1}{l|}{} & \multicolumn{1}{l|}{\begin{tabular}[c]{@{}l@{}}+ Emotions\\ + Demographics\end{tabular}} & \textcolor{negative_color}{32.3} & 5.8 & 87.6 \\ \cline{2-5} 
 
 \multicolumn{1}{l|}{} & \multicolumn{1}{l|}{\begin{tabular}[c]{@{}l@{}}+ Emotions\\ + Generation Error\\ + User Feedback\end{tabular}} & \textcolor{negative_color}{30.9} & 5.4 & \textcolor{negative_color}{85.5}
 
\end{tabular}}
  \caption{Generation accuracy in the question answering and task-oriented dialogues for Flan-T5. The best-performing models are printed in \textbf{bold}. Differences from the baselines that are greater than $\pm1.0$ are colored \textcolor{positive_color}{green} and \textcolor{negative_color}{red}.}
  \label{tab:tod_qa_comparison_flan_t5}
\end{table}

For Flan-T5~\cite{flan_t5} and GPT-2~\cite{gpt2}, the results in question answering are usually much higher than for the other tasks. A manual analysis revealed that the responses generated for question answering are primarily summaries of the corresponding knowledge documents, like the target sequences for this task. Therefore, we assume that this is the reason for the comparatively good generation accuracy for this dialogue type. We also assume that these knowledge documents serve as a regulating mechanism when learning from feedback, similar to those used in related work~\cite{xu-etal-2023-learning, xu2023improving, ung-etal-2022-saferdialogues} (see also the examples in Appendix~\ref{appendix:negative_example}). We found that the responses generated for the other tasks in these experiments still fit the context well, but often deviate from the target sequences. This is also expressed in the behavior of the scores. While the F1-Score measures word overlap and is therefore more affected, the other metrics, which focus more on contextual similarity, are less affected. We assume that the results could be different if we could find a similar guiding mechanism (or guiding source) for the other tasks. The task descriptions from dialogue generation (Appendix~\ref{appendix:task_descriptions}) could be an interesting starting point for such experiments, as they provide a pattern for the expected dialogue flow and information about the required slot values.

\begin{table}[ht]
  \centering
  \resizebox*{\linewidth}{!}{
\begin{tabular}{llrrr}
  & \multicolumn{1}{l}{\textbf{Experiment}} & \multicolumn{1}{l}{\textbf{F1}} & \multicolumn{1}{l}{\textbf{BLEU}} & \multicolumn{1}{l}{\textbf{BertScore}} \\ \hhline{=====}
  
 \multicolumn{1}{l|}{\multirow{5}{*}{QA}} & \multicolumn{1}{l|}{GPT-2} & 35.3 & 11.3 & 86.4\\ \hhline{~|====}
 
 \multicolumn{1}{l|}{} & \multicolumn{1}{l|}{\textbf{+ Emotions}} & \textbf{\textcolor{positive_color}{40.6}} & \textbf{\textcolor{positive_color}{16.2}} & \textbf{\textcolor{positive_color}{89.6}} \\ \cline{2-5} 
 
 \multicolumn{1}{l|}{} & \multicolumn{1}{l|}{+ Demographics} & \textcolor{positive_color}{38.6} & \textcolor{negative_color}{10.1} & \textcolor{positive_color}{89.6}\\ \cline{2-5} 
 
 \multicolumn{1}{l|}{} & \multicolumn{1}{l|}{\begin{tabular}[c]{@{}l@{}}+ Emotions\\ + Demographics\end{tabular}} & \textcolor{positive_color}{40.9} & 11.2 & \textcolor{positive_color}{89.2}\\ \cline{2-5}
 
 \multicolumn{1}{l|}{} & \multicolumn{1}{l|}{\begin{tabular}[c]{@{}l@{}}+ Emotions\\ + Demographics\\ + Generation Error\\ + User Feedback\end{tabular}} & \textcolor{positive_color}{37.4} & 12.1 & \textcolor{positive_color}{89.0} \\ \cline{1-5}
 
 \multicolumn{1}{l|}{\multirow{5}{*}{Others}} & \multicolumn{1}{l|}{GPT-2} & 34.4 & 11.1 & 86.3 \\ \hhline{~|====}
 
 \multicolumn{1}{l|}{} & \multicolumn{1}{l|}{+ Emotions} & 34.1 & 10.7 & 86.1 \\ \cline{2-5}
 
 \multicolumn{1}{l|}{} & \multicolumn{1}{l|}{+ Demographics} & \textcolor{negative_color}{31.7} & \textcolor{negative_color}{9.8} & 85.9 \\ \cline{2-5}
 
 \multicolumn{1}{l|}{} & \multicolumn{1}{l|}{\textbf{\begin{tabular}[c]{@{}l@{}}+ Emotions\\ + Demographics\end{tabular}}} & \textbf{34.5} & \textbf{11.6} & \textbf{86.4} \\ \cline{2-5}
 
 \multicolumn{1}{l|}{} & \multicolumn{1}{l|}{\begin{tabular}[c]{@{}l@{}}+ Emotions\\ + Demographics\\ + Generation Error\\ + User Feedback\end{tabular}} & \textcolor{negative_color}{26.2} & \textcolor{negative_color}{7.9} & 85.6
 
\end{tabular}}
  \caption{Generation accuracy in the question answering and task-oriented dialogues for GPT-2. The best-performing models are printed in \textbf{bold}. Differences from the baselines that are greater than $\pm1.0$ are colored \textcolor{positive_color}{green} and \textcolor{negative_color}{red}.}
  \label{tab:tod_qa_comparison_gpt_2}
\end{table}

For Llama 2, we do not observe any significant change in performance for either question answering or the other tasks. We attribute this to the observation made in Section~\ref{sec:experiments} that the system utterances generated by Llama 2 usually significantly deviate in length from the target sequence (although we used the same number of new tokens in all our experiments), resulting in lower word-overlapping scores.

\begin{table}[ht]
  \centering
  \resizebox*{\linewidth}{!}{
\begin{tabular}{llrrr}
  & \multicolumn{1}{l}{\textbf{Experiment}} & \multicolumn{1}{l}{\textbf{F1}} & \multicolumn{1}{l}{\textbf{BLEU}} & \multicolumn{1}{l}{\textbf{BertScore}} \\ \hhline{=====}
  
 \multicolumn{1}{l|}{\multirow{5}{*}{QA}} & \multicolumn{1}{l|}{\textbf{Llama 2}} & \textbf{34.0} & \textbf{12.7} & \textbf{84.8} \\ \hhline{~|====}
 \multicolumn{1}{l|}{} & \multicolumn{1}{l|}{+ Emotions} & \textcolor{negative_color}{32.1} & \textcolor{negative_color}{10.6} & 84.9 \\ \cline{2-5} 
 
 \multicolumn{1}{l|}{} & \multicolumn{1}{l|}{+ Demographics} & \textcolor{negative_color}{31.5} & \textcolor{negative_color}{6.2} & \textcolor{positive_color}{86.1} \\ \cline{2-5} 
 
 \multicolumn{1}{l|}{} & \multicolumn{1}{l|}{\begin{tabular}[c]{@{}l@{}}+ Emotions\\ + Demographics\end{tabular}} & \textcolor{negative_color}{29.1} & \textcolor{negative_color}{6.1} & \textcolor{positive_color}{85.9} \\ \cline{2-5} 
 
 \multicolumn{1}{l|}{} & \multicolumn{1}{l|}{\begin{tabular}[c]{@{}l@{}}+ Emotions\\ + User Feedback\end{tabular}} & \textcolor{negative_color}{24.4} & \textcolor{negative_color}{7.9} & \textcolor{negative_color}{76.0} \\ \cline{1-5} 
 
 \multicolumn{1}{l|}{\multirow{5}{*}{Others}} & \multicolumn{1}{l|}{Llama 2} & 26.5 & 8.0 & 86.6 \\ \hhline{~|====}
 
 \multicolumn{1}{l|}{} & \multicolumn{1}{l|}{+ Emotions} & \textcolor{positive_color}{28.3} & \textcolor{negative_color}{5.9} & \textcolor{negative_color}{85.4} \\ \cline{2-5} 
 
 \multicolumn{1}{l|}{} & \multicolumn{1}{l|}{+ Demographics} & \textcolor{positive_color}{28.9} & \textcolor{negative_color}{5.6} & 85.7 \\ \cline{2-5} 
 
 \multicolumn{1}{l|}{} & \multicolumn{1}{l|}{\textbf{\begin{tabular}[c]{@{}l@{}}+ Emotions\\ + Demographics\end{tabular}}} & \textbf{27.4} & \textbf{8.3} & \textbf{86.3} \\ \cline{2-5} 
 
 \multicolumn{1}{l|}{} & \multicolumn{1}{l|}{\begin{tabular}[c]{@{}l@{}}+ Emotions\\ + User Feedback\end{tabular}} & \textcolor{negative_color}{22.9} & \textcolor{negative_color}{4.7} & 87.4
 
\end{tabular}}
  \caption{Generation accuracy in the question answering and task-oriented dialogues for Llama 2. The best-performing models are printed in \textbf{bold}. Differences from the baselines that are greater than $\pm1.0$ are colored \textcolor{positive_color}{green} and \textcolor{negative_color}{red}.}
  \label{tab:tod_qa_comparison_llama_2}
\end{table}

\begin{table*}[ht]
  \centering
  \resizebox*{0.9\linewidth}{!}{\begin{tabular}{llrrrrrrrrr}
    \multicolumn{1}{c}{\textbf{}} &  \multicolumn{1}{c}{\textbf{Experiment}} & \multicolumn{4}{c}{\textbf{Task Completion}} & \multicolumn{2}{c}{\textbf{Quality}} & \multicolumn{3}{c}{\textbf{Generation Accuracy}} \\ \hline
     & \multicolumn{1}{l}{} & \multicolumn{1}{c}{\textbf{Inform}} & \multicolumn{1}{c}{\textbf{Success}} & \multicolumn{1}{c}{\textbf{Intent Acc.}} & \multicolumn{1}{c|}{\textbf{Slot Acc.}} & \multicolumn{1}{c}{\textbf{Q²}} & \multicolumn{1}{c|}{\textbf{Toxicity}} & \multicolumn{1}{c}{\textbf{F1}} & \multicolumn{1}{c}{\textbf{BLEU}} & \multicolumn{1}{c|}{\textbf{BertScore}} \\ \hhline{===========}

    \multicolumn{1}{l|}{\multirow{4}{*}{\begin{tabular}[c]{@{}l@{}}Llama 3 \\ Feedback-Free\end{tabular}}} & \multicolumn{1}{l|}{Llama 3} & 74.5 & 71.2 & 79.6 & \multicolumn{1}{r|}{82.7} & 24.5 & \multicolumn{1}{r|}{0.02} & 31.4 & 12.3 & \multicolumn{1}{r|}{87.5} \\ \hhline{~|==========|}
        
    \multicolumn{1}{c|}{}  & \multicolumn{1}{l|}{\textbf{\quad+User Emotions}} & \textbf{\textcolor{positive_color}{88.1}} & \textbf{\textcolor{positive_color}{85.4}} & \textbf{\textcolor{positive_color}{85.1}} & \multicolumn{1}{r|}{\textbf{\textcolor{positive_color}{94.5}}} & \textbf{\textcolor{positive_color}{31.4}} & \multicolumn{1}{r|}{\textbf{0.01}} &\textbf{\textcolor{positive_color}{39.6}} & \textbf{\textcolor{positive_color}{14.1}} & \multicolumn{1}{r|}{\textbf{87.2}} \\ \cline{2-11}
    
    \multicolumn{1}{c|}{}  & \multicolumn{1}{l|}{\quad+Demographic Info.} & \textcolor{positive_color}{87.2} & \textcolor{positive_color}{85.4} & \textcolor{positive_color}{83.5} & \multicolumn{1}{r|}{\textcolor{positive_color}{89.4}} & \textcolor{positive_color}{28.3} & \multicolumn{1}{r|}{0.02} & \textcolor{positive_color}{35.2} & \textcolor{positive_color}{17.2} & \multicolumn{1}{r|}{87.2} \\ \cline{2-11}
    
    \multicolumn{1}{c|}{}  & \multicolumn{1}{l|}{\begin{tabular}[c]{@{}l@{}}\quad+User Emotions\\ \quad+Demographic Info.\end{tabular}} & \textcolor{positive_color}{84.6} & \textcolor{positive_color}{83.1} & \textcolor{positive_color}{85.7} & \multicolumn{1}{r|}{\textcolor{positive_color}{93.1}} & \textcolor{positive_color}{26.1} & \multicolumn{1}{r|}{0.02} & 31.7 & 12.4 & \multicolumn{1}{r|}{87.1}\\ \hline
    
    \multicolumn{1}{l|}{\multirow{3}{*}{\begin{tabular}[c]{@{}l@{}}Feedback\end{tabular}}}  & \multicolumn{1}{l|}{\quad+Generation Error} & \multicolumn{1}{r}{\textcolor{positive_color}{93.4}} & \multicolumn{1}{r}{\textcolor{positive_color}{91.5}} & \multicolumn{1}{r}{\textcolor{positive_color}{74.6}} & \multicolumn{1}{r|}{\textcolor{positive_color}{98.1}} & \textcolor{positive_color}{29.5} & \multicolumn{1}{r|}{0.02} & \multicolumn{1}{r}{\textcolor{negative_color}{28.8}} & \multicolumn{1}{r}{\textcolor{negative_color}{9.7}} & \multicolumn{1}{r|}{86.5} \\ \cline{2-11}
    
    \multicolumn{1}{c|}{}  & \multicolumn{1}{l|}{\textbf{\quad+User Feedback}} & \multicolumn{1}{r}{\textbf{\textcolor{positive_color}{96.1}}} & \multicolumn{1}{r}{\textbf{\textcolor{positive_color}{95.7}}} & \multicolumn{1}{r}{\textbf{\textcolor{positive_color}{82.5}}} & \multicolumn{1}{r|}{\textbf{\textcolor{positive_color}{98.3}}} &  \textbf{\textcolor{positive_color}{39.8}} & \multicolumn{1}{r|}{\textbf{0.02}} & \multicolumn{1}{r}{\textbf{\textcolor{positive_color}{33.1}}} & \multicolumn{1}{r}{\textbf{11.6}} & \multicolumn{1}{r|}{\textbf{86.4}} \\ \cline{2-11}
    
    \multicolumn{1}{c|}{}  & \multicolumn{1}{l|}{\begin{tabular}[c]{@{}l@{}}\quad+Generation Error\\     \quad+User Feedback\end{tabular}} & \multicolumn{1}{r}{\textcolor{positive_color}{92.4}} & \multicolumn{1}{r}{\textcolor{positive_color}{90.9}} & \multicolumn{1}{r}{\textcolor{negative_color}{75.1}} & \multicolumn{1}{r|}{\textcolor{negative_color}{79.6}} & \textcolor{positive_color}{39.4} & \multicolumn{1}{r|}{0.02} & \multicolumn{1}{r}{\textcolor{positive_color}{33.5}} & \multicolumn{1}{r}{11.7} & \multicolumn{1}{r|}{86.8} \\ \hhline{===========}

    \multicolumn{1}{l|}{Llama 3} & \multicolumn{1}{l|}{In-Context} & 16.4 & 19.3 & 18.7 & \multicolumn{1}{r|}{22.7} & 17.3 & \multicolumn{1}{r|}{0.02} & 12.5 & 4.4 & \multicolumn{1}{r|}{82.3}
    
    \end{tabular}}

  \caption{Results of our experiments with Llama 3~\cite{llama3}. In general, we observe a huge performance improvement compared to our experiments with Llama 2~\cite{touvron2023llama2} (see Table~\ref{tab:main_experiments_automatic_evaluation}). As in Table~\ref{tab:main_experiments_automatic_evaluation}, we use the pretrained models finetuned on the feedback-free dialogues as deltas. The best-performing models are printed in \textbf{bold}. Differences greater than $\pm1.0$ are colored \textcolor{positive_color}{green} and \textcolor{negative_color}{red}.}  \label{tab:llama3_results}
\end{table*}

\subsection{Experiments with Llama 3}\label{appendix:llama3}
During our work on this project, Meta AI released the Llama 3 model series~\cite{llama3} as the successor to Llama 2~\cite{touvron2023llama2}. In our experiments with Llama 2 (see Section~\ref{sec:experiments}), the model showed a low capacity for intent and slot prediction, and we found that the generated responses often suffered from hallucinations. Our human annotators reported the responses generated by the feedback-free model as frequently unrelated to the dialogue context and factually incorrect. For this reason, we repeated our experiments from Llama 2 with Llama 3 (8B) and applied our metrics for automatic evaluation on the results. The instruction used for training and evaluation can be found in Appendix~\ref{appendix:input_sequences}. 

The results in Table~\ref{tab:llama3_results} shows that in comparison to the Llama 2 results (Table~\ref{tab:main_experiments_automatic_evaluation}), Llama 3 performs significantly better. The finetuned models also mostly show improved performance compared to Flan-T5~\cite{flan_t5} and GPT-2~\cite{gpt2}, especially in the task completion metrics. We manually inspected the quality of some randomly selected dialogues (both feedback-free and feedback) for generated intents, slots and responses. We could not reproduce the observations from the Llama 2 human evlauation (Section~\ref{sub_sec:second_set_of_experiments}). The generated responses were predominantly relevant in the context of the dialogue (hallucinations were very rarely observed) and the generated slots and intents were mostly complete and correct. We partly attribute this to the improved prompt we used in these experiments (see Appendix~\ref{appendix:input_sequences}). 

The results of the in-context learning experiment (we included the task description along with examples in the instruction) are significantly worse in all aspects, emphasizing the importance of finetuning in task-oriented and document-grounded dialogues. However, they show an overall advantage of Llama 3 over Llama 2 (see Table~\ref{tab:main_experiments_automatic_evaluation}). Therefore, we assume this is primarily due to the improved capabilities of Llama 3 in language generation and reasoning tasks~\cite{llama3}.

\subsection{Impact of Learning From Generation Errors on Toxicity}\label{appendix:impact_toxicity}
Although their share is small (Lack of Sociality in Figure~\ref{fig:error_user_reaction_types}), the generation errors in \ourdata contain potentially toxic and disrespectful language. Table~\ref{tab:main_experiments_automatic_evaluation} shows that the toxicity of generated responses is generally negligible (values are $\leq 0.03$). However, we observe some outliers in the Flan-T5~\cite{flan_t5} and Llama 2~\cite{touvron2023llama2} feedback models which score $\geq 0.1$. For example, \emph{Flan-T5 + User Emotions + Generation Error + User Feedback} once generated \textit{Alright, that's a start. What else? And don't forget, I need it in simple terms. None of that fancy shit.} to request for missing information in the case of parcel shipping. For Llama 2 + Emotions + Generation Error, toxicity scores $\geq 0.1$ are sometimes observed in the case of question answering, e.g., \textit{The Legal Protection does not apply to events resulting from popular riots, acts of terrorism, vandalism, earthquakes, strikes and lock-outs, possession or use of radioactive substances, disputes concerning family, inheritance and gift law, tax and administrative disputes, events resulting from popular riots, insurrections, military operations, acts of terrorism, vandalism, earthquakes, strikes and lock-outs}. However, we consider these false positives, as they may contain critical terms but do not offend the user personally. Overall, generated system utterances with a toxicity score $\geq 0.1$ are extremely rare ($\leq 0.1\%$ of the responses generated with these models in the test data). 

In the feedback-free experiments, we did not observe any generated system utterance with a toxicity score $\geq 0.1$.

\section{Crowdsourcing Study}\label{appendix:crowdsourcing}
We did a crowdsourcing study to investigate how humans perceive the impact of feedback training. For this, we hired 42 crowdworkers on Prolific for an hourly salary of 9,00\$ (the hourly salary recommended by the platform). Our requirement for participation was as follows:

\begin{itemize}
    \item Fluent in English.
    \item At least 10 previous submissions to other studies on Prolific.
    \item Approval rate of at least 90\%.
\end{itemize}

We did not restrict participation to US citizens. We also did not consider gender, age or other educational background. We had no further influence on the allocation of participants. To manage this (and the payment) is the purpose of Prolific. The participants were forwarded to Google Forms, which we used to implement our study (see appendix~\ref{appendix:prolific}).

Overall, from the 42 people who decided to participate, 23 were from South Africa and 19 from european countries. 24 of the participants were female. 18 were male. The average age was 28.54 years. The youngest person was 21 years old. The oldest person 62. We did not conduct any recruitment test in advance. Instead, we provided the participants with three test samples in the live study so that they could become familiar with the task and our rating scheme. We reviewed all submissions in detail and decided to exclude the results of two participants, as they contained predominantly incomprehensible ratings (we paid them nevertheless).

\begin{figure}[htb]
  \centering
    \includegraphics[width=1.0\linewidth]{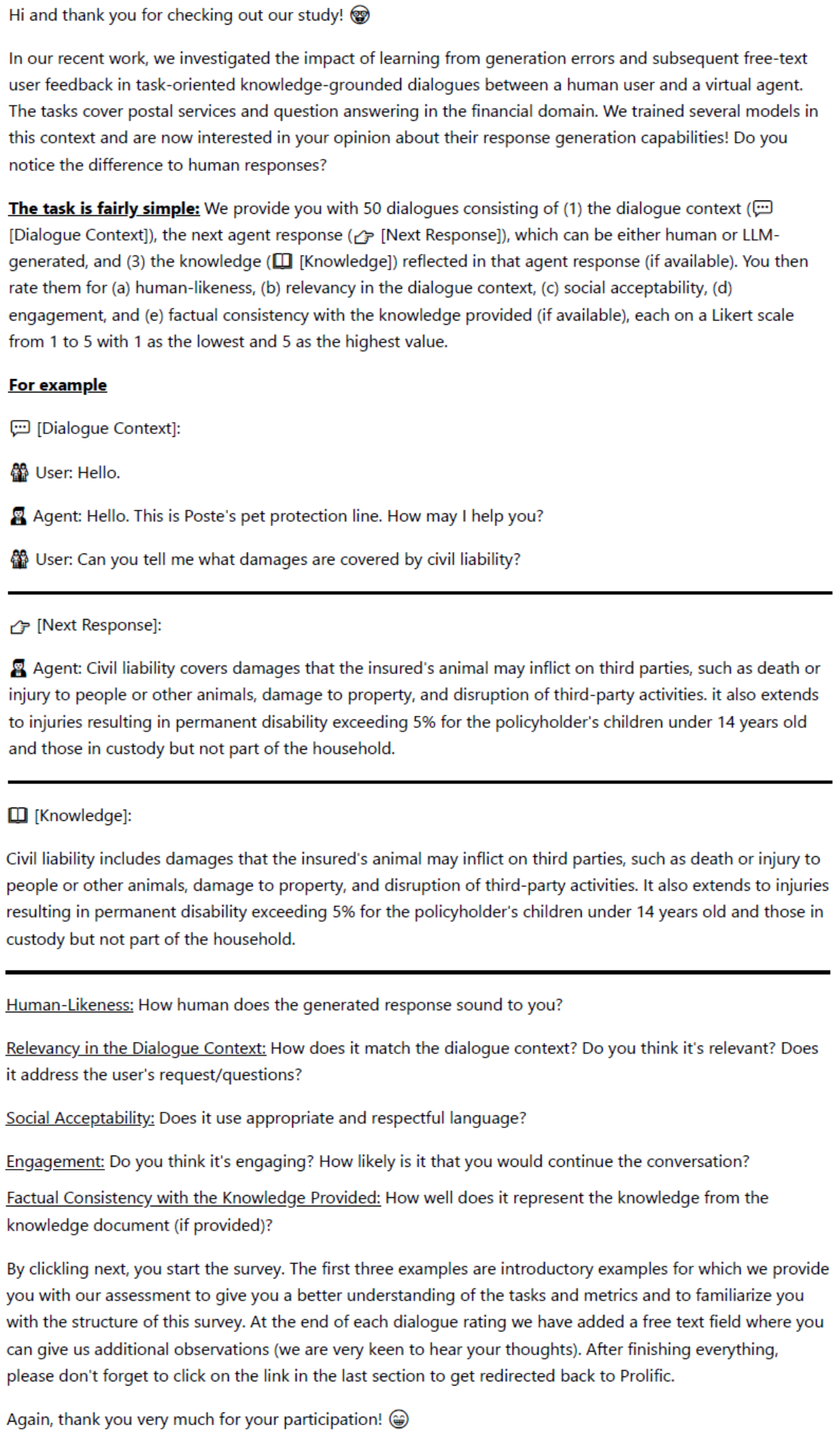}  
    \caption{Task Description for our crowdsourcing study.} 
    \label{fig:task_description_crowdsourcing}
  \end{figure}

\subsection{Implementation and Procedure}\label{appendix:prolific}
\begin{figure}[htb]
  \centering
    \includegraphics[width=1.0\linewidth]{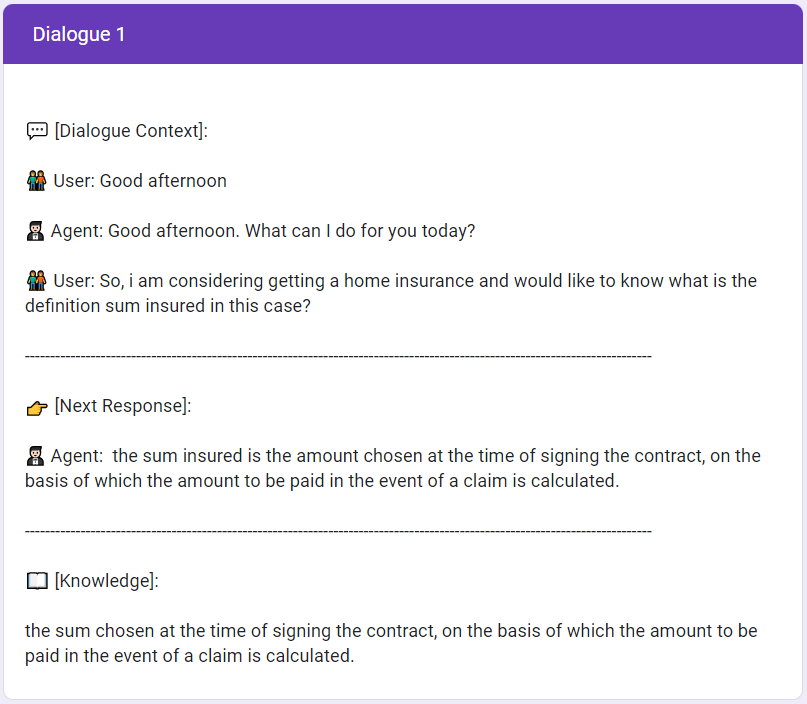}  
    \caption{Example dialogue from our crowdsourcing study. Each dialogue was represented as a separate section in a Google Forms survey. } 
    \label{fig:google_forms}
  \end{figure}

We implemented the crowdsourcing study using Google Forms\footnote{\href{https://www.google.de/intl/de/forms/about/}{Google Forms} is a survey management software that is part of the free, web-based Google Docs Editor Suite from Google (last accessed 09 May 2024).}, using one section per dialogue. At the beginning of the survey, we provided the participants with extensive instructions describing the task and the rating scheme (see Figure~\ref{fig:task_description_crowdsourcing}). Figure~\ref{fig:google_forms} shows an example dialogue from our study.

\begin{figure}[htb]
  \centering
    \includegraphics[width=1.0\linewidth]{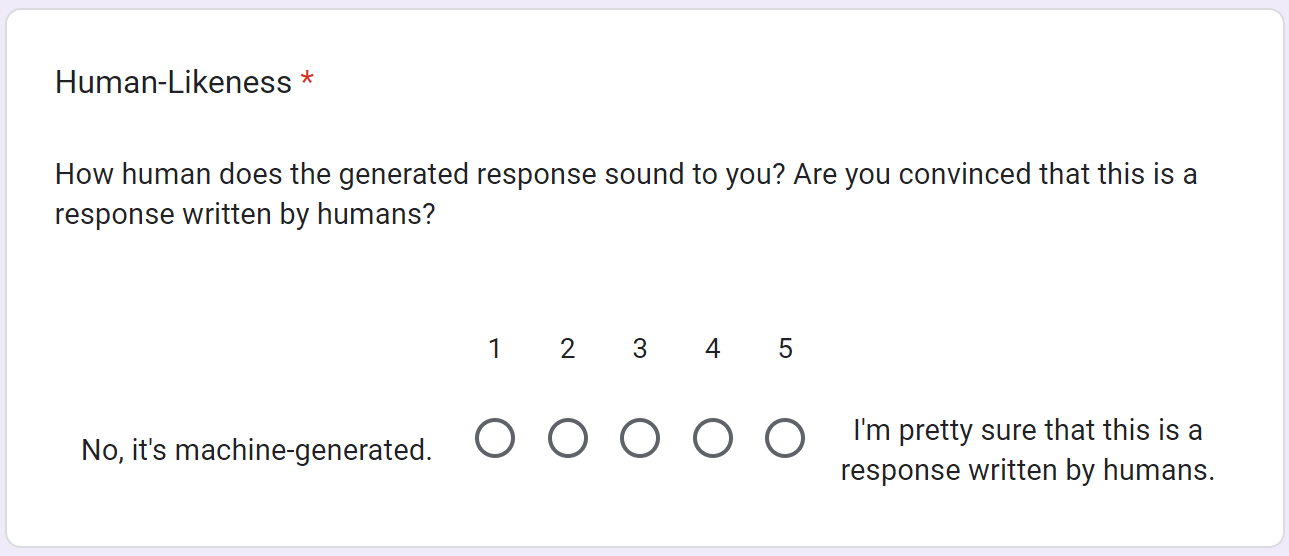}  
    \caption{Linear scale for human-likeness from our crowdsourcing study.} 
    \label{fig:rating_example}
  \end{figure}

For each dialogue, we presented the annotators the dialogue context, generated response and knowledge document (in the case of question answering), but did not indicate whether the response was generated by a human or language model. We used Python scripts and the Google Forms API to automatically create and fill surveys with 50 dialogues randomly sampled from the 300 pre-selected test dialogues. Below the dialogues, we added the rating forms using linear scales from one to five. Figure~\ref{fig:rating_example} shows an example.

We asked the annotators to rate the generated responses ([Next Response]) for the following attributes: human likeness (how human does the generated response sound?), relevancy in the dialogue context (does it match the dialogue context? does it address the user's concern?), sociality (does it use appropriate and respectful language?), factual consistency (how well does it represent the knowledge from the document?), and engagement (do you think it is engaging? would you like to continue the conversation?). Filling the rating forms was mandatory. At the end of each section we added a free text field in which they were asked to provide us with additional observations (if any). 

\subsection{Examples}\label{appendix:crowdsourcing_add_insights}
\begin{figure}[htb]
  \centering
    \includegraphics[width=0.8\linewidth]{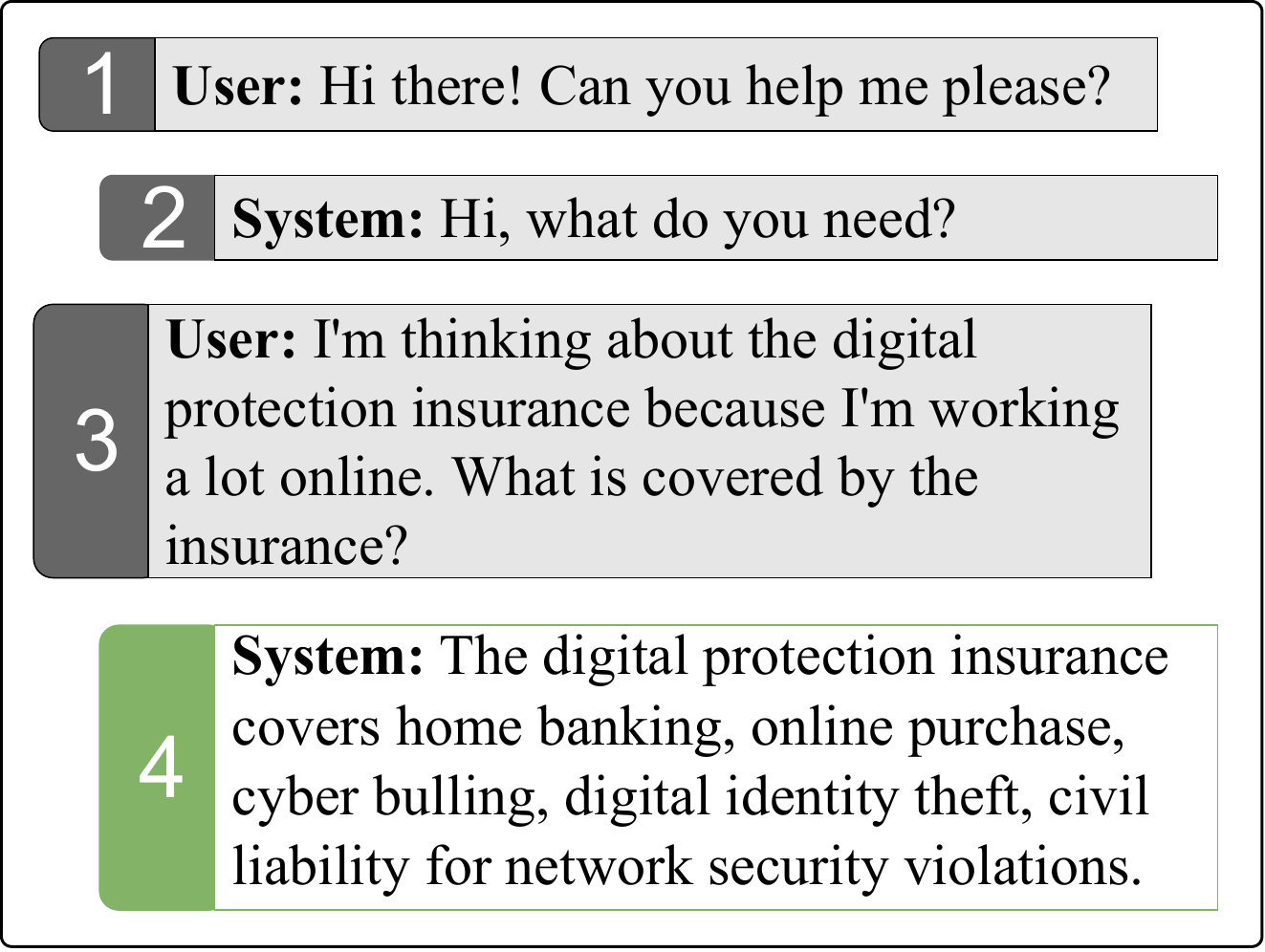}  
    \caption{A response generated by the Flan-T5 feedback model presented to the annotators in our crowdsourcing study.} 
    \label{fig:flan_t5_crowdsourcing}
  \end{figure}

In this section, we provide examples to illustrate the observations reported by the annotators in our crowdsourcing study. The responses generated by the models used are highlighted in green in the figures. Figure~\ref{fig:flan_t5_crowdsourcing} shows the context of a dialogue and the response that was generated by the Flan-T5~\cite{flan_t5} feedback model. While the annotators agreed that the information presented in the response is correct, they reported in their comments that they felt it was not inviting to continue the conversation.

\begin{figure}[htb]
  \centering
    \includegraphics[width=0.8\linewidth]{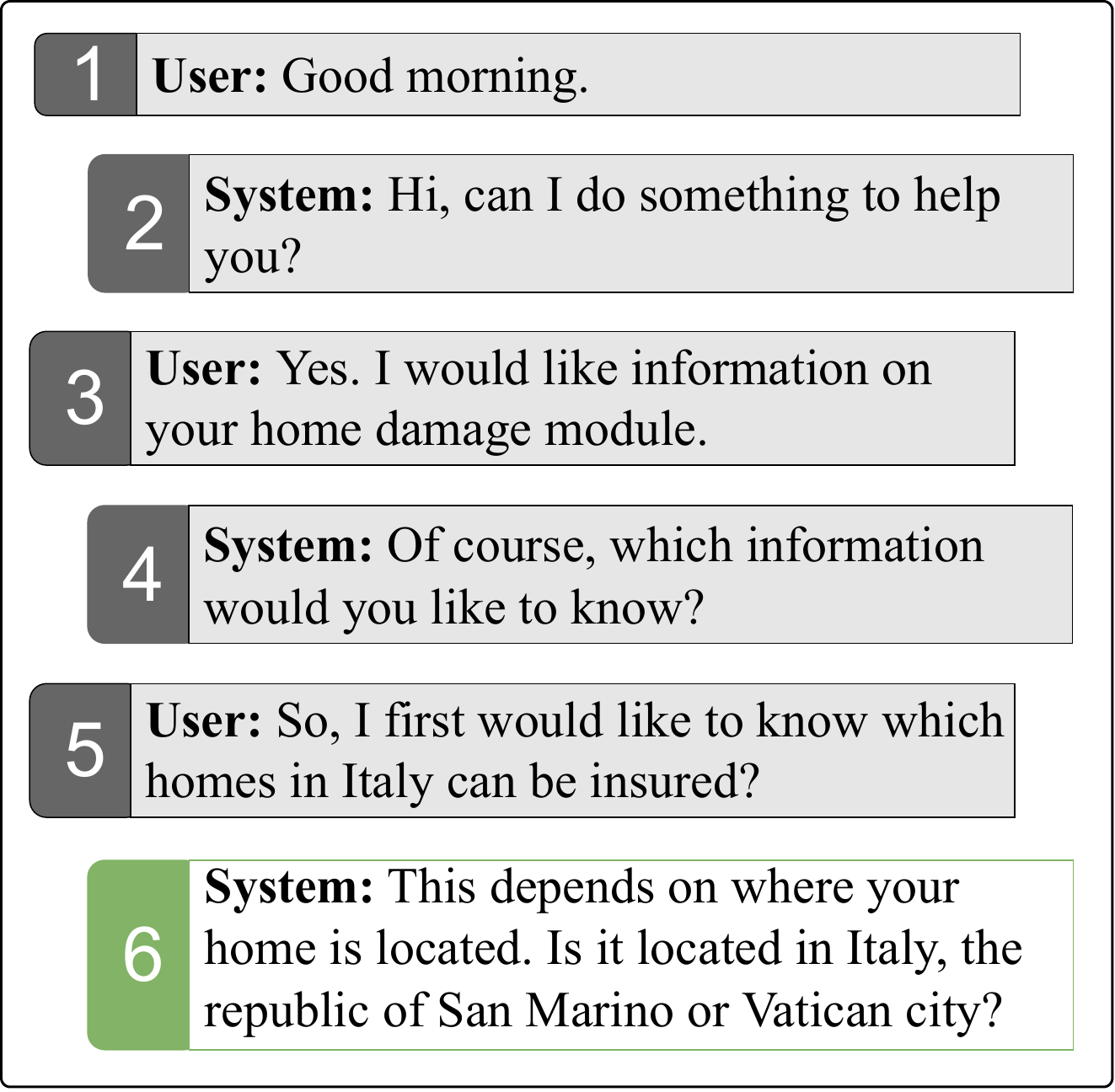}  
    \caption{A response generated by the GPT-2 feedback model presented to the annotators in our crowdsourcing study.} 
    \label{fig:gpt2_crowdsourcing}
  \end{figure}

It answers the question, but does not contain any further request for interaction. Figure~\ref{fig:gpt2_crowdsourcing} shows a response generated by the GPT-2~\cite{gpt2} feedback model. This is one of the responses reported as less attentive. The user asks for information about insurance for home damages and focuses on houses in Italy in utterance five. The model does not pick up this information and returns a counter-question asking the user whether the house is in Italy, the Republic of San Marino or the Vatican City.

\begin{figure}[htb]
  \centering
    \includegraphics[width=0.8\linewidth]{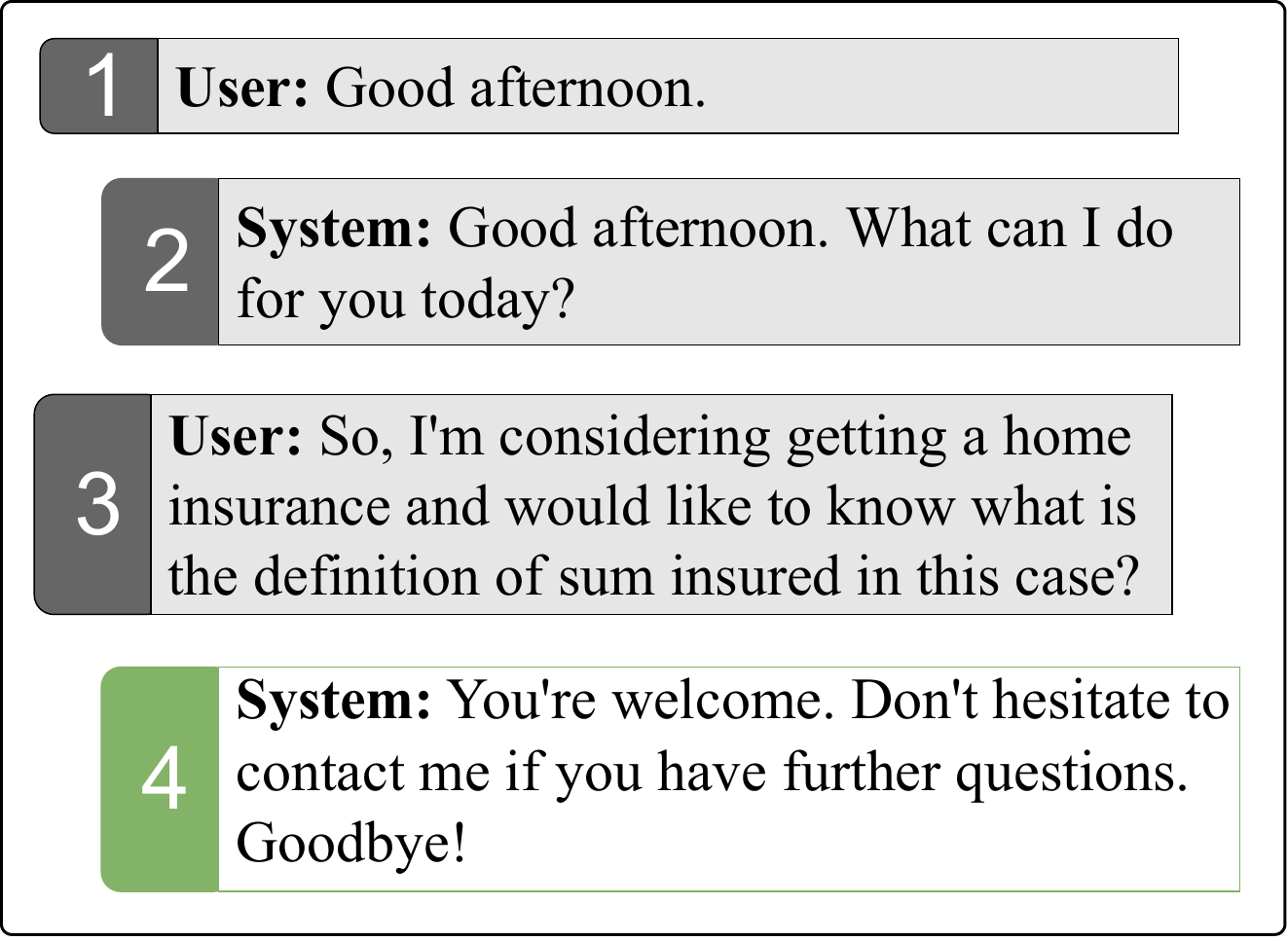}  
    \caption{A response generated by the Llama 2 feedback-free model presented to the annotators in our crowdsourcing study.} 
    \label{fig:llama2_crowdsourcing}
  \end{figure}

Figure~\ref{fig:llama2_crowdsourcing} shows a sample from the Llama 2~\cite{touvron2023llama2} feedback-free model, which illustrates why annotators reported many of them as illogical or unrelated to the dialogue context. The user is asking about the definition of sum insured in the case of home insurance. Instead of responding to this, the model says goodbye to the user.

We selected these samples because they are exemplary for the observations made by the annotators. The same phenomena were also observed in responses generated to longer dialogue contexts.

\section{Continual Learning From Feedback Data}\label{appendix:continual_learning}
Table~\ref{tab:continual_learning_test_data} shows the results of our continual learning experiments using the most promising configurations from Section~\ref{sec:experiments} and the human-human test dialogues. For each model, we use the best performing feedback-free model from Section~\ref{sec:experiments} (Table~\ref{tab:main_experiments_automatic_evaluation}) as a starting point. We train the models sequentially with each version of the feedback dialogues, starting with Version 2 and once with annotations for implicit user feedback (Feedback) and once without (No Feedback). 
The rest of the training procedure and hyperparameter configuration corresponds to what is described in Appendix~\ref{appendix:hyperparameters}. Due to the large number of experiments, we only present single run results here (the results in Section~\ref{sec:experiments} were averaged over three runs).

Interestingly, the results are rather mixed. We observe a tendency for the task completion metrics to improve with each version of the dialogues, especially when using the annotations for implicit user feedback. The same applies to factual consistency ($Q^2$~\cite{honovich-etal-2021-q2}). 

 \begin{table*}[htb]
  \centering
  \resizebox*{\linewidth}{!}{\begin{tabular}{llrrrrrrrrr}
\multicolumn{1}{c}{\textbf{Model}} & \multicolumn{1}{c}{\textbf{Experiment}} & \multicolumn{4}{c}{\textbf{Task Completion}} & \multicolumn{2}{c}{\textbf{Quality}} & \multicolumn{3}{c}{\textbf{Generation Accuracy}} \\ \hline
 & \multicolumn{1}{c|}{\textbf{}} & \multicolumn{1}{c}{\textbf{Inform}} & \multicolumn{1}{c}{\textbf{Success}} & \multicolumn{1}{c}{\textbf{Intent Acc.}} & \multicolumn{1}{c|}{\textbf{Slot Acc.}} & \textbf{Toxicity} & \multicolumn{1}{c|}{\textbf{Q²}} & \multicolumn{1}{c}{\textbf{F1}} & \multicolumn{1}{c}{\textbf{BLEU}} & \multicolumn{1}{c|}{\textbf{BertScore}}  \\ \hhline{===========}
 
\multicolumn{11}{c}{\textbf{Version 2}} \\ \hhline{===========}
\multicolumn{1}{l|}{\multirow{3}{*}{\begin{tabular}[c]{@{}l@{}}No Feedback\end{tabular}}} & \multicolumn{1}{l|}{\begin{tabular}[c]{@{}l@{}}Flan-T5 \\\quad+User Emotions\end{tabular}} & 86.5 & 83.2 & 86.8 & \multicolumn{1}{r|}{85.0} & 0.02 & \multicolumn{1}{r|}{55.6} & 52.8 & 29.4 & 89.4 \\ \cline{2-11}
\multicolumn{1}{c|}{}  & \multicolumn{1}{l|}{\begin{tabular}[c]{@{}l@{}}GPT-2 \\\quad+User Emotions \\\quad+Demographic Info.\end{tabular}} & 86.4 & 83.9 & 89.0 & \multicolumn{1}{r|}{81.6} & 0.02 & \multicolumn{1}{r|}{31.7} & 35.4 & 9.9 & 85.0 \\ \cline{2-11}
\multicolumn{1}{c|}{}  & \multicolumn{1}{l|}{\begin{tabular}[c]{@{}l@{}}Llama 2 \\\quad+User Emotions\end{tabular}} & 88.4 & 86.1 & 40.6 & \multicolumn{1}{r|}{39.8} & 0.02 & \multicolumn{1}{r|}{29.5} & 45.7 & 25.1 & 85.4 \\ \hline

\multicolumn{1}{l|}{\multirow{3}{*}{\begin{tabular}[c]{@{}l@{}}Feedback\end{tabular}}}  & \multicolumn{1}{l|}{\begin{tabular}[c]{@{}l@{}}Flan-T5 \\\quad+User Emotions \\\quad+Generation Error \\\quad+User Feedback\end{tabular}} & \multicolumn{1}{r}{95.6} & \multicolumn{1}{r}{93.2} & \multicolumn{1}{r}{87.5} & \multicolumn{1}{r|}{85.3} & 0.02 & \multicolumn{1}{r|}{59.8} & \multicolumn{1}{r}{54.9} & \multicolumn{1}{r}{33.0} & 89.7 \\ \cline{2-11}
\multicolumn{1}{c|}{}  & \multicolumn{1}{l|}{\begin{tabular}[c]{@{}l@{}}GPT-2 \\\quad+User Emotions \\\quad+Demographic Info. \\\quad+Generation Error \\\quad+User Feedback\end{tabular}} & \multicolumn{1}{r}{84.7} & \multicolumn{1}{r}{83.3} & \multicolumn{1}{r}{93.0} & \multicolumn{1}{r|}{85.0} & 0.02 & \multicolumn{1}{r|}{28.9} & \multicolumn{1}{r}{35.4} & \multicolumn{1}{r}{10.3} & \multicolumn{1}{r}{85.3} \\ \cline{2-11}
\multicolumn{1}{c|}{}  & \multicolumn{1}{l|}{\begin{tabular}[c]{@{}l@{}}Llama 2 \\\quad+User Emotions \\\quad+User Feedback\end{tabular}} & \multicolumn{1}{r}{91.1} & \multicolumn{1}{r}{94.9} & \multicolumn{1}{r}{51.2} & \multicolumn{1}{r|}{52.6} & 0.01 & \multicolumn{1}{r|}{30.3} & \multicolumn{1}{r}{40.8} & \multicolumn{1}{r}{19.6} & \multicolumn{1}{r}{84.9} \\ \hhline{===========}

\multicolumn{11}{c}{\textbf{Version 3}} \\ \hhline{===========}
\multicolumn{1}{l|}{\multirow{3}{*}{\begin{tabular}[c]{@{}l@{}}No Feedback\end{tabular}}} & \multicolumn{1}{l|}{\begin{tabular}[c]{@{}l@{}}Flan-T5 \\\quad+User Emotions\end{tabular}} & 86.9 & \textcolor{positive_color}{85.4} & \textcolor{negative_color}{80.8} & \multicolumn{1}{r|}{85.0} & 0.02 & 55.3 & \multicolumn{1}{|r}{52.5} & \textcolor{positive_color}{31.5} & \multicolumn{1}{r}{88.8}  \\ \cline{2-11}
\multicolumn{1}{c|}{}  & \multicolumn{1}{l|}{\begin{tabular}[c]{@{}l@{}}GPT-2 \\\quad+User Emotions \\\quad+Demographic Info.\end{tabular}} & 86.5 & 83.3 & 89.0 & \multicolumn{1}{r|}{\textcolor{positive_color}{83.4}} & 0.02 & \textcolor{negative_color}{29.2} & \multicolumn{1}{|r}{\textcolor{negative_color}{33.7}} & 9.6 & 84.3 \\ \cline{2-11}
\multicolumn{1}{c|}{}  & \multicolumn{1}{l|}{\begin{tabular}[c]{@{}l@{}}Llama 2 \\\quad+User Emotions\end{tabular}} & \textcolor{negative_color}{87.4} & 85.2 & \textcolor{negative_color}{38.5} & \multicolumn{1}{r|}{\textcolor{negative_color}{37.6}} & 0.02 & 30.4 & \multicolumn{1}{|r}{\textcolor{negative_color}{30.0}} & \textcolor{negative_color}{15.3} & \textcolor{negative_color}{83.0} \\ \hline

\multicolumn{1}{l|}{\multirow{3}{*}{\begin{tabular}[c]{@{}l@{}}Feedback\end{tabular}}}  & \multicolumn{1}{l|}{\begin{tabular}[c]{@{}l@{}}Flan-T5 \\\quad+User Emotions \\\quad+Generation Error \\\quad+User Feedback\end{tabular}} & \multicolumn{1}{r}{96.1} & \multicolumn{1}{r}{\textcolor{positive_color}{95.1}} & \multicolumn{1}{r}{\textcolor{negative_color}{82.3}} & \multicolumn{1}{r|}{84.6} & 0.02 & \multicolumn{1}{r}{\textcolor{negative_color}{58.8}} & \multicolumn{1}{|r}{\textcolor{negative_color}{49.2}} & \multicolumn{1}{r}{\textcolor{negative_color}{29.8}} & \textcolor{negative_color}{88.3} \\ \cline{2-11}
\multicolumn{1}{c|}{}  & \multicolumn{1}{l|}{\begin{tabular}[c]{@{}l@{}}GPT-2 \\\quad+User Emotions \\\quad+Demographic Info. \\\quad+Generation Error \\\quad+User Feedback\end{tabular}} & \multicolumn{1}{r}{\textcolor{positive_color}{94.7}} & \multicolumn{1}{r}{\textcolor{positive_color}{89.1}} & \multicolumn{1}{r}{93.0} & \multicolumn{1}{r|}{85.0} & 0.02 & \multicolumn{1}{r}{\textcolor{positive_color}{33.2}} & \multicolumn{1}{|r}{36.1} & \multicolumn{1}{r}{\textcolor{positive_color}{12.0}} & \multicolumn{1}{r}{85.1} \\ \cline{2-11}
\multicolumn{1}{c|}{}  & \multicolumn{1}{l|}{\begin{tabular}[c]{@{}l@{}}Llama 2 \\\quad+User Emotions \\\quad+User Feedback\end{tabular}} & \multicolumn{1}{r}{92.0} & \multicolumn{1}{r}{\textcolor{negative_color}{90.6}} & \multicolumn{1}{r}{\textcolor{positive_color}{55.1}} & \multicolumn{1}{r|}{\textcolor{positive_color}{58.6}} & 0.01 & \multicolumn{1}{r}{\textcolor{positive_color}{32.4}} & \multicolumn{1}{|r}{\textcolor{negative_color}{39.4}} & \multicolumn{1}{r}{\textcolor{positive_color}{21.2}} & \multicolumn{1}{r}{\textcolor{negative_color}{74.9}}\\ \hhline{===========}

\multicolumn{11}{c}{\textbf{Version 4}} \\ \hhline{===========}
\multicolumn{1}{l|}{\multirow{3}{*}{\begin{tabular}[c]{@{}l@{}}No Feedback\end{tabular}}} & \multicolumn{1}{l|}{\begin{tabular}[c]{@{}l@{}}Flan-T5 \\\quad+User Emotions\end{tabular}} & 85.9 & 83.2 & \textcolor{negative_color}{81.0} & \multicolumn{1}{r|}{\textcolor{negative_color}{82.9}} & 0.02 & \multicolumn{1}{r}{\textcolor{positive_color}{57.3}} & \multicolumn{1}{|r}{\textcolor{negative_color}{49.6}} & 28.7 & \multicolumn{1}{r}{\textcolor{negative_color}{88.3}} \\ \cline{2-11}
\multicolumn{1}{c|}{}  & \multicolumn{1}{l|}{\begin{tabular}[c]{@{}l@{}}GPT-2 \\\quad+User Emotions \\\quad+Demographic Info.\end{tabular}} & 87.1 & 83.6 & \textcolor{negative_color}{86.0} & \multicolumn{1}{r|}{\textcolor{positive_color}{84.6}} & 0.02 & 31.4 & \multicolumn{1}{|r}{\textcolor{negative_color}{33.4}} & 10.2 & \multicolumn{1}{r}{84.8} \\ \cline{2-11}
\multicolumn{1}{c|}{}  & \multicolumn{1}{l|}{\begin{tabular}[c]{@{}l@{}}Llama 2 \\\quad+User Emotions\end{tabular}} & \textcolor{positive_color}{90.1} & 86.7 & 41.0 & \multicolumn{1}{r|}{\textcolor{positive_color}{42.3}} & 0.02 & \textcolor{negative_color}{31.6} & \multicolumn{1}{|r}{\textcolor{negative_color}{28.7}} & \textcolor{negative_color}{14.5} & \multicolumn{1}{r}{85.4} \\ \hline

\multicolumn{1}{l|}{\multirow{3}{*}{\begin{tabular}[c]{@{}l@{}}Feedback\end{tabular}}}  & \multicolumn{1}{l|}{\begin{tabular}[c]{@{}l@{}}Flan-T5 \\\quad+User Emotions \\\quad+Generation Error \\\quad+User Feedback\end{tabular}} & \multicolumn{1}{r}{\textcolor{positive_color}{98.1}} & \multicolumn{1}{r}{\textcolor{positive_color}{96.2}} & \multicolumn{1}{r}{\textcolor{negative_color}{81.3}} & \multicolumn{1}{r|}{85.0} & 0.02 & \multicolumn{1}{r}{60.5} & \multicolumn{1}{|r}{\textcolor{negative_color}{50.6}} & \multicolumn{1}{r}{32.7} & \multicolumn{1}{r}{\textcolor{negative_color}{88.6}} \\ \cline{2-11}
\multicolumn{1}{c|}{}  & \multicolumn{1}{l|}{\begin{tabular}[c]{@{}l@{}}GPT-2 \\\quad+User Emotions \\\quad+Demographic Info. \\\quad+Generation Error \\\quad+User Feedback\end{tabular}} & \multicolumn{1}{r}{\textcolor{positive_color}{99.3}} & \multicolumn{1}{r}{\textcolor{positive_color}{97.5}} & \multicolumn{1}{r}{\textcolor{negative_color}{91.0}} & \multicolumn{1}{r|}{85.5} & 0.02 & \multicolumn{1}{r}{\textcolor{positive_color}{34.9}} & \multicolumn{1}{|r}{34.9} & \multicolumn{1}{r}{\textcolor{positive_color}{11.7}} & \multicolumn{1}{r}{\textcolor{positive_color}{87.5}} \\ \cline{2-11}
\multicolumn{1}{c|}{}  & \multicolumn{1}{l|}{\begin{tabular}[c]{@{}l@{}}Llama 2 \\\quad+User Emotions \\\quad+User Feedback\end{tabular}} &  \multicolumn{1}{r}{\textcolor{positive_color}{94.5}} & \multicolumn{1}{r}{\textcolor{positive_color}{96.1}} & \multicolumn{1}{r}{\textcolor{positive_color}{54.4}} & \multicolumn{1}{r|}{\textcolor{positive_color}{60.2}} & 0.01 & \multicolumn{1}{r}{\textcolor{positive_color}{33.9}} & \multicolumn{1}{|r}{40.1} & \multicolumn{1}{r}{\textcolor{negative_color}{15.4}} & \multicolumn{1}{r}{\textcolor{negative_color}{82.1}}
\end{tabular}}

  \caption{Results achieved on the test data for each stage. We use the respective models from Version 2 as deltas for calculating the difference in Version 3 and 4.}
  \label{tab:continual_learning_test_data}
\end{table*}

\end{document}